\PassOptionsToPackage{table,dvipsnames}{xcolor}
\documentclass[]{selfevolagent}
\usepackage{microtype}
\usepackage{amsfonts}
\usepackage{xcolor}
\definecolor{lavender}{RGB}{230,230,250} 
\definecolor{softlavender}{RGB}{238, 223, 255} 
\definecolor{selfevolagent}{RGB}{220,211,237} 
\usepackage{graphicx}
\usepackage{booktabs}
\usepackage{wrapfig}
\usepackage{float}
\usepackage{amsmath}
\usepackage{amsthm}
\usepackage{subcaption}
\usepackage{makecell}
\usepackage{algpseudocode}
\usepackage[linesnumbered,lined,boxed,commentsnumbered,ruled,longend]{algorithm2e}
\theoremstyle{plain}

\theoremstyle{definition}

\theoremstyle{remark}

\usepackage{enumitem}
\usepackage{pifont}
\usepackage[edges]{forest}
\usetikzlibrary{arrows.meta}
\usepackage{tikz}
\usepackage{graphicx}     
\usepackage{booktabs}     
\usepackage{enumitem}     
\usepackage[most]{tcolorbox} 
\usepackage{fontawesome5}
\usepackage{booktabs} 
\usepackage{url}
\usepackage{wrapfig}
\usepackage{longtable}
\usepackage{tabularx}
\usetikzlibrary{
  positioning,
  calc,
  shapes.symbols,
  shapes.geometric,
  shapes.misc
}
\usepackage[T1]{fontenc}
\definecolor{selfevolagent_dark}{HTML}{37D2A6} 
\definecolor{selfevolagent_light}{HTML}{9BE9D3}
\definecolor{selfevolagent_lighter}{HTML}{CDF4E9}

\newcommand\blfootnote[1]{%
  \begingroup
  \renewcommand\thefootnote{}\footnote{#1}%
  \addtocounter{footnote}{-1}%
  \endgroup
}

\usepackage{svg}

\usepackage{xltabular}
\usepackage{booktabs}
\usepackage{caption}

\usepackage{fontawesome5}

\newcommand{\ghlink}[1]{\faIcon{github}\,\href{#1}{GitHub}}
\newcommand{\weblink}[1]{\faIcon{globe}\,\href{#1}{Website}}

\newcommand{\cmark}{\textcolor{green}{\ding{52}}}
\newcommand{\xmark}{\textcolor{red}{\ding{56}}}

\definecolor{rootcolor}{RGB}{101, 45, 144}   
\definecolor{catcolor}{RGB}{255, 192, 0}     
\definecolor{subcatcolor}{RGB}{237, 125, 49} 
\definecolor{papercolor}{RGB}{68, 114, 196}  

\title{Memory in the Age of AI Agents: A Survey\\
\Large Forms, Functions and Dynamics
}
\author{Yuyang Hu$^{\dagger}$}
\author{Shichun Liu$^{\dagger}$}
\author{Yanwei Yue$^{\dagger}$}
\author{Guibin Zhang$^{\dagger\text{\faCube}}$}
\author{Boyang Liu}
\author{Fangyi Zhu}
\author{Jiahang Lin} 
\author{\mbox{Honglin Guo}}
\author{Shihan Dou}
\author{Zhiheng Xi}
\author{Senjie Jin}
\author{Jiejun Tan}
\author{Yanbin Yin}
\author{Jiongnan Liu}
\author{Zeyu Zhang}
\author{Zhongxiang Sun}
\author{Yutao Zhu}
\author{Hao Sun}
\author{Boci Peng}
\author{Zhenrong Cheng}
\author{Xuanbo Fan}
\author{Jiaxin Guo}
\author{Xinlei Yu}
\author{Zhenhong Zhou}
\author{Zewen Hu}
\author{Jiahao Huo}
\author{Junhao Wang}
\author{Yuwei Niu}
\author{Yu Wang}

\author{Zhenfei Yin}
\author{Xiaobin Hu}
\author{\mbox{Yue Liao}}
\author{Qiankun Li}
\author{Kun Wang}
\author{Wangchunshu Zhou}
\author{Yixin Liu}

\author{Dawei Cheng}
\author{Qi Zhang}
\author{Tao Gui$^\ddagger$}

\author{\mbox{Shirui Pan}}
\author{Yan Zhang$^\ddagger$}

\author{Philip Torr}
\author{Zhicheng Dou$^\ddagger$}
\author{Ji-Rong Wen}

\author{Xuanjing Huang$^\ddagger$}
\author{Yu-Gang Jiang}
\author{Shuicheng Yan$^\ddagger$}

\affiliation{
\vspace{0.5em}
$^\dagger$Core Contributors with Names Listed Alphabetically.\;\;\faCube \;Project Organizer.\;\;$^\ddagger$Core Supervisors.

\vspace{0.5em}

\texttt{\textbf{Affiliations}: National University of Singapore, Renmin University of China, Fudan University, Peking University, Nanyang Technological University, Tongji University, University of California San Diego, Hong Kong University of Science and Technology (Guangzhou), Griffith University, Georgia Institute of Technology, OPPO, Oxford University}}

\abstract{
\begin{abstract}

Memory has emerged, and will continue to remain, a core capability of foundation model-based agents. It underpins long-horizon reasoning, continual adaptation, and effective interaction with complex environments. As research on agent memory rapidly expands and attracts unprecedented attention, the field has also become increasingly fragmented. Existing works that fall under the umbrella of agent memory often differ substantially in their motivations, implementations, assumptions, and evaluation protocols, while the proliferation of loosely defined memory terminologies has further obscured conceptual clarity. Traditional taxonomies such as long/short-term memory have proven insufficient to capture the diversity and dynamics of contemporary agent memory systems.
This survey aims to provide an up-to-date and comprehensive landscape of current agent memory research. We begin by clearly delineating the scope of agent memory and distinguishing it from related concepts such as LLM memory, retrieval augmented generation (RAG), and context engineering. We then examine agent memory through the unified lenses of \textbf{forms}, \textbf{functions}, and \textbf{dynamics}. From the perspective of forms, we identify three dominant realizations of agent memory, namely \textit{token-level}, \textit{parametric}, and \textit{latent memory}. From the perspective of functions, we move beyond coarse temporal categorizations and propose a finer-grained taxonomy that distinguishes \textit{factual}, \textit{experiential}, and \textit{working memory}. From the perspective of dynamics, we analyze how memory is formed, evolved, and retrieved over time as agents interact with their environments.
To support empirical research and practical development, we compile a comprehensive summary of representative benchmarks and open source memory frameworks. Beyond consolidation, we articulate a forward-looking perspective on emerging research frontiers, including automation-oriented memory design, the deep integration of reinforcement learning with memory systems, multimodal memory, shared memory for multi-agent systems, and trustworthiness issues. We hope this survey serves not only as a reference for existing work, but also as a conceptual foundation for rethinking memory as a first-class primitive in the design of future agentic intelligence.

\end{abstract}

}
\metadata[\faEnvelope\ Main Contact]{guibinz@u.nus.edu, yuyang.hu@ruc.edu.cn, liusc24@m.fudan.edu.cn, 
ywyue25@stu.pku.edu.cn}
\metadata[{\raisebox{-0.2ex}{\includegraphics[height=1em]{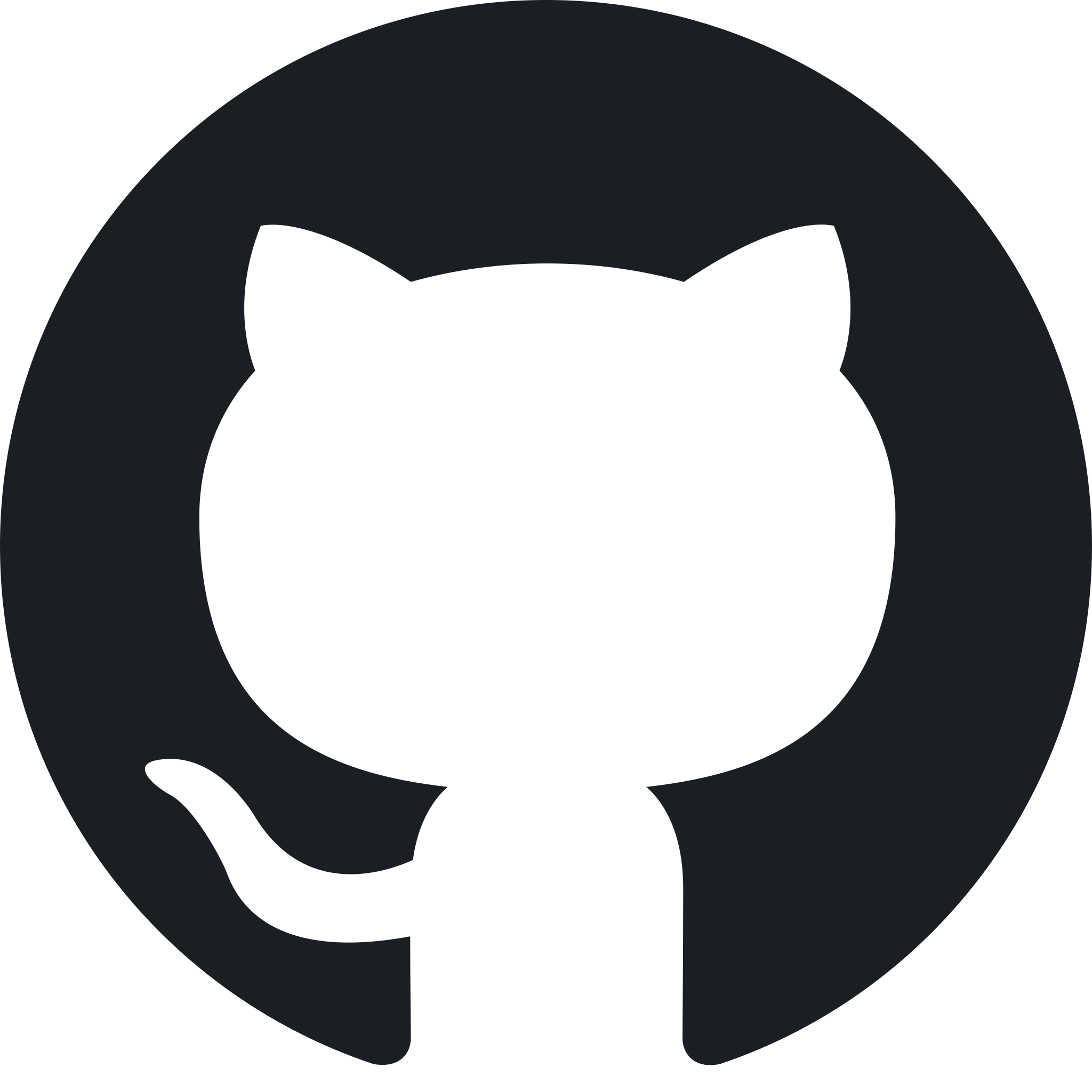}}\ Github}]{\url{https://github.com/Shichun-Liu/Agent-Memory-Paper-List}}

\begin{document}

 \maketitle

 \blfootnote{Note: If you identify your own or other papers relevant to this survey that have not been discussed (we apologize for any such omissions due to the rapidly expanding literature), please feel free to contact us via email or raise an issue on \href{https://github.com/Shichun-Liu/Agent-Memory-Paper-List}{GitHub}.}
\clearpage
\tableofcontents
\clearpage
\begin{figure}[!t]
    \centering
    \includegraphics[width=1\linewidth]{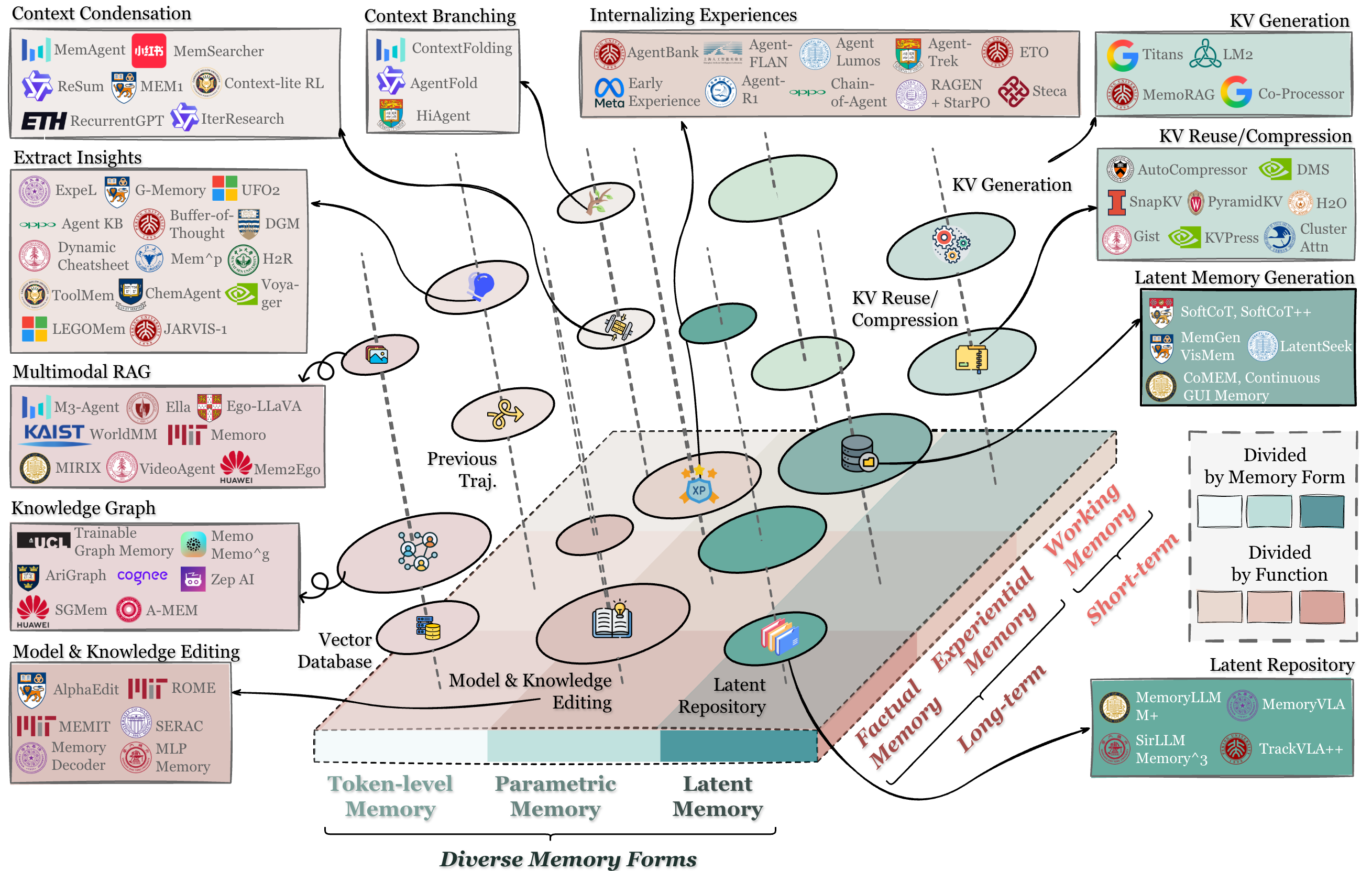}
    \caption{
    Overview of agent memory organized by the unified taxonomy of \emph{forms} (\Cref{sec:what-memory}), \emph{functions} (\Cref{sec:why-memory}), and \emph{dynamics} (\Cref{sec:how-memory}). The diagram positions memory artifacts by their dominant form and primary function. It further maps representative systems into this taxonomy to provide a consolidated landscape.
    }
    \label{fig:intro-main}
\end{figure}

\section{Introduction}

The past two years have witnessed the overwhelming evolution of increasingly capable large language models (LLMs) into powerful AI agents~\citep{matarazzo2025surveylargelanguagemodels,minaee2025largelanguagemodelssurvey,luo2025largelanguagemodelagent}. These foundation-model-powered agents have demonstrated remarkable progress across diverse domains such as deep research~\citep{xu2025comprehensivesurveydeepresearch,zhang2025deepresearchsurveyautonomous}, software engineering~\citep{wang2024agentssoftwareengineeringsurvey}, and scientific discovery~\citep{wei2025aiscienceagenticscience}, continuously advancing the trajectory toward artificial general interlligence (AGI)~\citep{fang2025comprehensivesurveyselfevolvingai,durante2024agentaisurveyinghorizons}. Although early conceptions of ``agents'' were highly heterogeneous, a growing consensus has since emerged within the community: beyond a pure LLM backbone, an agent is typically equipped with capabilities such as \textit{reasoning}, \textit{planning}, \textit{perception}, \textit{memory}, and \textit{tool-use}. Some of these abilities, such as reasoning and tool-use, have been largely internalized within model parameters through reinforcement learning~\citep{wang2025reinforcementlearningreasoninglarge,toollearningsurvey}, while some still depend heavily on external agentic scaffolds. Together, these components transform LLMs from static conditional generators into learnable policies that can interact with diverse external environments and adaptively evolve over time~\citep{zhang2025landscapeagenticreinforcementlearning,liu2025advances}.

Among these agentic faculties, \textit{memory} stands out as a cornerstone, explicitly enabling the transformation of static LLMs, whose parameters cannot be rapidly updated, into adaptive agents capable of continual adaptation through environmental interaction~\citep{zhang2025ruc_survey,wu2025humanmemoryaimemory}. From an application perspective, numerous domains demand agents with proactive memory management rather than ephemeral, forgetful behaviors: personalized chatbots~\citep{Chhikara2025mem0,li2025helloagainllmpoweredpersonalized}, recommender systems~\citep{liu2025agentcfmemoryenhancedllmbasedagents}, social simulations~\citep{park2023generativeagentsinteractivesimulacra,yang2025oasisopenagentsocial}, and financial investigations~\citep{zhang2024multimodalfoundationagentfinancial} all rely on the agent's ability to process, store, and manage historical information. From a developmental standpoint, one of the defining aspirations of AGI research is to endow agents with the capacity for continual evolution through environment interactions~\citep {hendrycks2025agidefinition}, a capability fundamentally grounded in agent memory. 

\paragraph{Agent Memory Needs A New Taxonomy} 
Given the growing significance and community attention surrounding agent memory systems, it has become both timely and necessary to provide an updated perspective on contemporary agent memory research. The motivation for a new taxonomy and survey is twofold:  
\ding{182} \textbf{Limitations of Existing Taxonomies:}  
While several recent surveys have provided valuable and comprehensive overviews of agent memory~\citep{zhang2025ruc_survey,wu2025humanmemoryaimemory}, their taxonomies were developed prior to a number of rapid methodological advances and therefore do not fully reflect the current breadth and complexity of the research landscape. For example, emerging directions in 2025, such as memory frameworks that distill reusable tools from past experiences~\citep{qiu2025agentdistilltrainingfreeagentdistillation,qiu2025alitageneralistagentenabling,zhao2025pyvisionagenticvisiondynamic}, or memory-augmented test-time scaling methods~\citep{zhang2025latentevolveselfevolvingtesttimescaling,suzgun2025dynamiccheatsheettesttimelearning}, remain underrepresented in earlier classification schemes.
\ding{183} \textbf{Conceptual Fragmentation:} With the explosive growth of memory-related studies, the concept itself has become increasingly expansive and fragmented. Researchers often find that papers claiming to study ``agent memory'' differ drastically in implementation, objectives, and underlying assumptions. The proliferation of diverse terminologies (declarative, episodic, semantic, parametric memory, etc.) further obscures conceptual clarity, highlighting the urgent need for a coherent taxonomy that can unify these emerging concepts.  

Therefore, this paper seeks to establish a systematic framework that reconciles existing definitions, bridges emerging trends, and elucidates the foundational principles of memory in agentic systems. Specifically, this survey aims to address the following key questions:

\vspace{0.6em}
\begin{tcolorbox}[
  colback=selfevolagent_light!20,
  colframe=selfevolagent_light!80,
  colbacktitle=selfevolagent_light!80,
  coltitle=black,
  title={\bfseries\fontfamily{ppl}\selectfont{Key Questions}},
  boxrule=2pt,
  arc=5pt,
  drop shadow,
  parbox=false,
  before skip=5pt,
  after skip=5pt,
  left=5pt,   
  right=5pt,
]
\begin{itemize}[leftmargin=*]
\item[\ding{182}] How is \textit{agent memory} defined, and how does it relate to related concepts such as LLM memory, retrieval-augmented generation (RAG), and context engineering?  
\item[\ding{183}] \textbf{Forms:} What architectural or representational forms can agent memory take?  
\item[\ding{184}] \textbf{Functions:} Why is agent memory needed, and what roles or purposes does it serve?  
\item[\ding{185}] \textbf{Dynamics:} How does agent memory operate, adapt, and evolve over time?  
\item[\ding{186}] What are the promising frontiers for advancing agent memory research?
\end{itemize}
\end{tcolorbox}

\vspace{1em}
To address question \ding{182}, we first provide formal definitions for LLM-based agents and agent memory systems in \Cref{sec:preliminary}, and present a detailed comparison between agent memory and related concepts such as LLM memory, RAG, and context engineering.  
Following the ``Forms–Functions–Dynamics'' triangle, we offer a structured overview of agent memory. Question \ding{183} examines the architectural forms of memory, which we discuss in \Cref{sec:what-memory}, highlighting three mainstream implementations: token-level, parametric, and latent memory. Question \ding{184} concerns the functional roles of memory, addressed in \Cref{sec:why-memory}, where we distinguish between \textit{factual memory}, which records knowledge from agents' interactions with users and the environment; \textit{experiential memory}, which incrementally enhances the agent's problem-solving capabilities through task execution; and \textit{working memory}, which manages workspace information during individual task instances. 
Question \ding{185} focuses on the lifecycle and operational dynamics of agent memory, which we present sequentially in terms of memory formulation, retrieval, and evolution.

After surveying existing research through the lenses of ``Forms–Functions–Dynamics,'' we further provide our perspectives and insights on agent memory research. 
To facilitate knowledge sharing and future development, we first summarize key benchmarks and framework resources in \Cref{sec:resource}. 
Building upon this foundation, we then address question \ding{186} by exploring several emerging yet underdeveloped research frontiers in \Cref{sec:frontier}, including automation-oriented memory design, the integration of reinforcement learning (RL), multimodal memory, shared memory for multi-agent systems, and trustworthy issues.

\vspace{-0.4em}
\paragraph{Contributions} 
The contributions of this survey can be summarized as follows:  
(1) We present an up-to-date and multidimensional taxonomy of agent memory from the perspective of ``forms–functions–dynamics,'' offering a structured lens through which to understand current developments in the field.  
(2) We provide an in-depth discussion on the suitability and interplay of different memory forms and functional purposes, offering insights into how various memory types can be effectively aligned with distinct agentic objectives.  
(3) We investigate emerging and promising research directions in agent memory, thereby outlining future opportunities and guiding pathways for advancement.  
(4) We compile a comprehensive collection of resources, including benchmarks and open-source frameworks, to support both researchers and practitioners in further exploration of agent memory systems.

\paragraph{Outline of the Survey}
The remainder of this survey is organized as follows. \Cref{sec:preliminary} formalizes LLM-based agents and agent memory systems, and clarifies their relationships with related concepts. \Cref{sec:what-memory}, \Cref{sec:why-memory}, and \Cref{sec:how-memory} respectively examine the forms, functions, and dynamics of agent memory. \Cref{sec:resource} summarizes representative benchmarks and framework resources. \Cref{sec:frontier} discusses emerging research frontiers and future directions. Finally, we conclude the survey with a summary of key insights in \Cref{sec:conclusion}.

\section{Preliminaries: Formalizing Agents and Memory}
\label{sec:preliminary}

LLM agents increasingly serve as the decision-making core of interactive systems that operate over time, manipulate external tools, and coordinate with humans or other agents. To study memory in such settings, we begin by formalizing LLM-based agent systems in a manner that encompasses both single-agent and multi-agent configurations. We then formalize the memory system coupled to the agent's decision process through read/write interactions, enabling a unified treatment of memory phenomena that arise both \emph{within} a task (inside-trial / short-term memory) and \emph{across} tasks (cross-trial / long-term memory).

\subsection{LLM-based Agent Systems}

\paragraph{Agents and Environment}
Let \( \mathcal{I} = \{1,\dots,N\} \) denote the index set of agents, where \(N=1\) corresponds to the single-agent case (e.g., ReAct), and \(N>1\) represents multi-agent settings such as debate~\citep{DBLP:conf/emnlp/LiDZHGLI24} or planner--executor architectures~\citep{wan2025remalearningmetathinkllms}.  
The environment is characterized by a state space \( \mathcal{S} \). At each time step \(t\), the environment evolves according to a controlled stochastic transition model
\[
s_{t+1} \sim \Psi(s_{t+1} \mid s_t, a_t),
\]
where \(a_t\) denotes the action executed at time \(t\). In multi-agent systems, this abstraction allows for either sequential decision-making (where a single agent acts at each step) or implicit coordination through environment-mediated effects. Each agent \(i \in \mathcal{I}\) receives an observation
\[
o_t^i = O_i(s_t, h_t^i, \mathcal{Q}),
\]
where \(h_t^i\) denotes the portion of the interaction history visible to agent \(i\). This history may include previous messages, intermediate tool outputs, partial reasoning traces, shared workspace states, or other agents' contributions, depending on the system design.  
\(\mathcal{Q}\) denotes the task specification, such as a user instruction, goal description, or external constraints, which is treated as fixed within a task unless otherwise specified.

\paragraph{Action Space}
A distinguishing feature of LLM-based agents is the heterogeneity of their action space. Rather than restricting actions to plain text generation, agents may operate over a multimodal and semantically structured action space, including:
\begin{itemize}[leftmargin=1.5em]
    \item \textbf{Natural-language generation}, such as producing intermediate reasoning, explanations, responses, or instructions~\citep{li2023camel,wu2024autogen,hong2023metagpt,qian2024chatdevcommunicativeagentssoftware}.
    \item \textbf{Tool invocation actions}, which call external APIs, search engines, calculators, databases, simulators, or code execution environments~\citep{qin2025flashsearcherfasteffectiveweb,DBLP:journals/corr/abs-2501-05366,zhou2023agents,zhou2024agents2}.
    \item \textbf{Planning actions}, which explicitly output task decompositions, execution plans, or subgoal specifications to guide later behavior~\citep{camel_workforce_docs,liu2025toolplanner,pan2024planningimaginationepisodicsimulation}.
    \item \textbf{Environment-control actions}, where the agent directly manipulates the external environment (e.g., navigation in embodied settings~\citep{shridhar2021alfworld,wang2022scienceworldagentsmarter5th}, editing a software repository~\citep{jimenez2023swe,aleithan2024swebenchenhancedcodingbenchmark}, or modifying a shared memory buffer).
    \item \textbf{Communication actions}, enabling collaboration or negotiation with other agents through structured messages~\citep{marro2024scalablecommunicationprotocolnetworks}.
\end{itemize}

These actions, though diverse in semantics, are unified by the fact that they are produced through an autoregressive LLM backbone conditioned on a contextual input. Formally, each agent \(i\) follows a policy
\[
a_t = \pi_i(o_t^i, m_t^i, \mathcal{Q}),
\]
where \(m_t^i\) is a memory-derived signal defined in \Cref{sec:formalize-agent-memory}.  
The policy may internally generate multi-step reasoning chains, latent deliberation, or scratchpad computations prior to emitting an executable action; such internal processes are abstracted away and not explicitly modeled.

\paragraph{Interaction Process and Trajectories}
A full execution of the system induces a trajectory
\[
\tau = (s_0, o_0, a_0, s_1, o_1, a_1, \dots, s_T),
\]
where \(T\) is determined by task termination conditions or system-specific stopping criteria. At each step, the trajectory reflects the interleaving of (i) environment observation, (ii) optional memory retrieval, (iii) LLM-based computation, and (iv) action execution that drives the next state transition.

This formulation captures a broad class of agentic systems, ranging from a single agent solving reasoning tasks with tool augmentation to teams of role-specialized agents collaboratively developing software~\citep{qian2024chatdevcommunicativeagentssoftware,wang2025evoagentx} or conducting scientific inquiry~\citep{weng2025deepscientistadvancingfrontierpushingscientific}. We next formalize the memory systems that integrate into this agent loop.

\subsection{Agent Memory Systems}
\label{sec:formalize-agent-memory}

While an LLM-based agent interacts with an environment, its instantaneous observation \(o_t^i\) is often insufficient for effective decision-making. Agents therefore rely on additional information derived from prior interactions, both within the current task and across previously completed tasks. We formalize this capability through a unified \emph{agent memory system}, represented as an evolving memory state
\[
\mathcal{M}_t \in \mathbb{M},
\]
where \(\mathbb{M}\) denotes the space of admissible memory configurations. No specific internal structure is imposed on \(\mathcal{M}_t\); it may take the form of a text buffer, key--value store, vector database, graph structure, or any hybrid representation.
At the beginning of a task, \(\mathcal{M}_t\) may already contain information distilled from prior trajectories (cross-trial memory). During task execution, new information accumulates and functions as short-term, task-specific memory. Both roles are supported within a single memory container, with temporal distinctions emerging from usage patterns rather than architectural separation.

\paragraph{Memory Lifecycle: Formation, Evolution, and Retrieval}
The dynamics of the memory system are characterized by three conceptual operators.

\textbf{Memory Formation}
At time step \(t\), the agent produces informational artifacts \(\phi_t\), which may include tool outputs, reasoning traces, partial plans, self-evaluations, or environmental feedback. A formation operator
\[
\mathcal{M}_{t+1}^{\mathrm{form}} = F(\mathcal{M}_t, \phi_t)
\]
selectively transforms these artifacts into memory candidates, extracting information with potential future utility rather than storing the entire interaction history verbatim.

\textbf{Memory Evolution}
Formed memory candidates are integrated into the existing memory base through an evolution operator
\[
\mathcal{M}_{t+1} = E(\mathcal{M}_{t+1}^{\mathrm{form}}),
\]
which may consolidate redundant entries~\citep{DBLP:conf/aaai/Zhao0XLLH24}, resolve conflicts~\citep{Rasmussen2025Zep,li2025memosoperatingmemoryaugmentedgeneration}, discard low-utility information~\citep{wangKARMAAugmentingEmbodied2025}, or restructure memory for efficient retrieval. The resulting memory state persists across subsequent decision steps and tasks.

\textbf{Memory Retrieval}
When selecting an action, agent \(i\) retrieves a context-dependent memory signal
\[
m_t^i = R(\mathcal{M}_t, o_t^i, \mathcal{Q}),
\]
where \(R\) denotes a retrieval operator that constructs a task-aware query and returns relevant memory content. The retrieved signal \(m_t^i\) is formatted for direct consumption by the LLM policy, for example as a sequence of textual snippets or a structured summary.

\paragraph{Temporal Roles Within the Agent Loop}
Although memory is represented as a unified state \( \mathcal{M}_t \), the three lifecycle operators (formation \(F\), evolution \(E\), and retrieval \(R\)) need not be invoked at every time step. Instead, different memory effects arise from distinct temporal invocation patterns.
For instance, some systems perform retrieval only once at task initialization,
\[
m_t^i =
\begin{cases}
R(\mathcal{M}_{0}, o_0^i, \mathcal{Q}), & t = 0, \\[4pt]
\bot, & t > 0,
\end{cases}
\]
where $\bot$ denotes null retrieval strategy. Others may retrieve memory intermittently or continuously based on contextual triggers.  
Similarly, memory formation may range from minimal accumulation of raw observations,
\[
\mathcal{M}_{t+1}^{\mathrm{form}} = \mathcal{M}_t \cup \{o_t^i\},
\]
to sophisticated extraction and refinement of reusable patterns or abstractions. Thus, \emph{inside a task}, short-term memory effects may arise from lightweight logging just as in~\cite{yao2023react,chen2023fireactlanguageagentfinetuning} or from more elaborate iterative refinement~\citep{huHiAgentHierarchicalWorking2025}; \emph{across tasks}, long-term memory may be updated episodically at task boundaries or continuously throughout operation. Short-term and long-term memory phenomena therefore emerge not from discrete architectural modules but from the temporal patterns with which formation, evolution, and retrieval are engaged.

\paragraph{Memory--Agent Coupling}
The interaction between memory and the agent’s decision process is similarly flexible. In general, the agent policy is written as
\[
a_t = \pi_i(o_t^i, m_t^i, \mathcal{Q}),
\]
where the retrieved memory signal \(m_t^i\) may be present or absent depending on the retrieval schedule. When retrieval is disabled at a given step, \(m_t^i\) can be treated as a distinguished null input.

Consequently, the overall agent loop consists of observing the environment, optionally retrieving memory, computing an action, receiving feedback, and optionally updating memory through formation and evolution. Different agent implementations instantiate different subsets of these operations at different temporal frequencies, giving rise to memory systems that range from passive buffers to actively evolving knowledge bases.

\subsection{Comparing Agent Memory with Other Key Concepts}

Despite the growing interest in agentic systems endowed with memory, the community’s understanding of what constitutes \emph{agent memory} remains fragmented. In practice, researchers and practitioners often conflate agent memory with related constructs such as LLM memory~\citep{wu2025humanmemoryaimemory}, retrieval-augmented generation (RAG)~\citep{gao2024retrievalaugmentedgenerationlargelanguage}, and context engineering~\citep{meiSurveyContextEngineering2025}. Although these concepts are intrinsically connected by their involvement in how information is managed and utilized in LLM-driven systems, they differ in scope, temporal characteristics, and functional roles.

\begin{figure}[!t]
    \centering
    \includegraphics[width=1\linewidth]{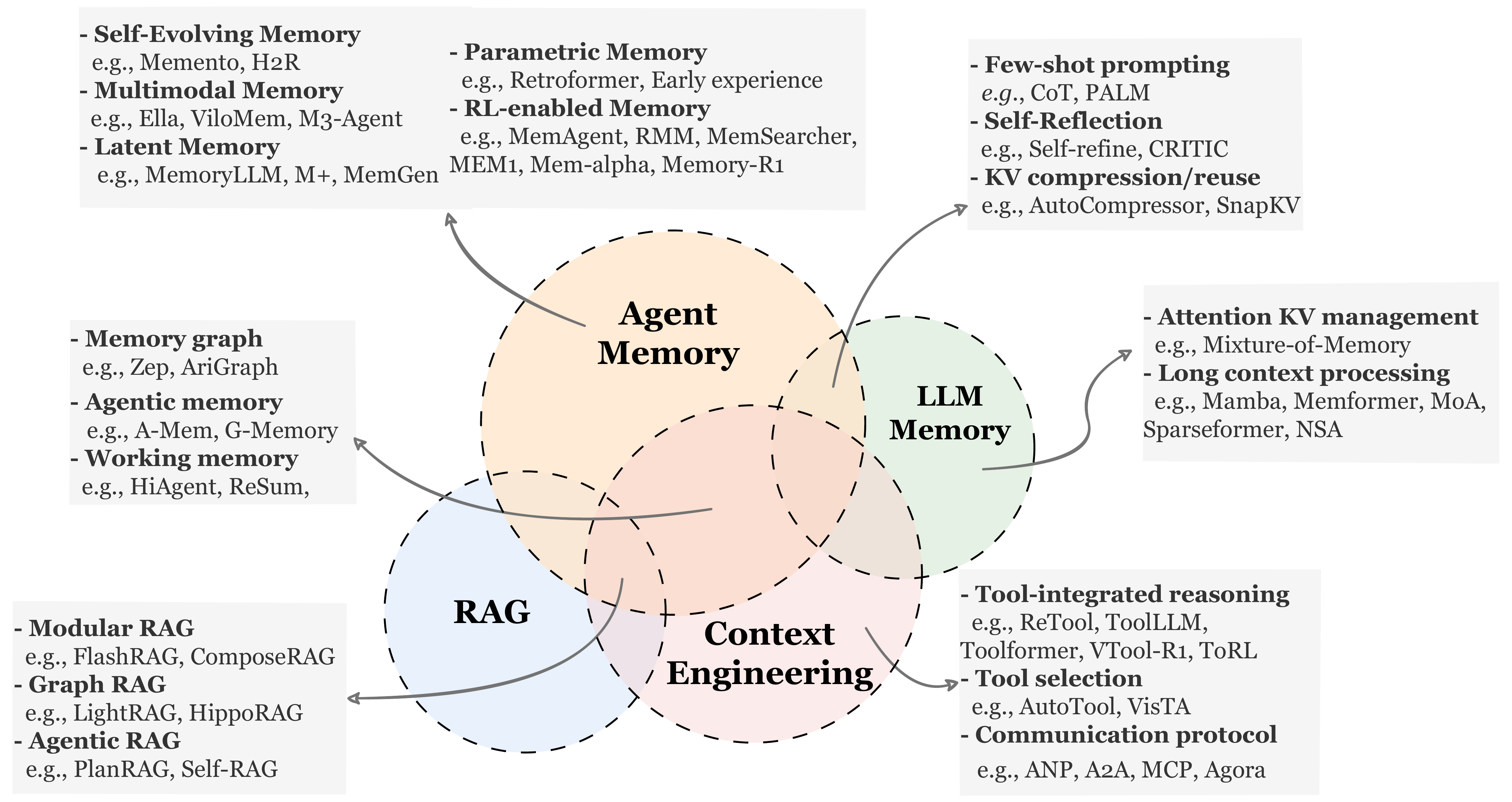}
    \caption{
    Conceptual comparison of \textbf{Agent Memory} with \textbf{LLM Memory}, \textbf{RAG}, and \textbf{Context Engineering}. The diagram illustrates shared technical implementations (e.g., KV reuse, graph retrieval) while highlighting fundamental distinctions: unlike the architectural optimizations of LLM Memory, the static knowledge access of RAG, or the transient resource management of Context Engineering, Agent Memory is uniquely characterized by its focus on maintaining a persistent and self-evolving cognitive state that integrates factual knowledge and experience. The listed categories and examples are illustrative rather than strictly parallel, serving as representative reference points to clarify conceptual relationships rather than to define a rigid taxonomy.
    }
    \label{fig:concept-comparison}
\end{figure}

These overlapping yet distinct notions have led to ambiguity in the literature and practice. To clarify these distinctions and situate agent memory within this broader landscape, we examine how agent memory \emph{relates to}, and \textit{diverges from}, LLM memory, RAG, and context engineering in the subsequent subsubsections.
\autoref{fig:concept-comparison} visually illustrates the commonalities and distinctions among these fields through a Venn diagram.

\subsubsection{Agent Memory vs. LLM Memory}

At a high level, \emph{agent memory} almost fully subsumes what has traditionally been referred to as \emph{LLM memory}. Since 2023, many works describing themselves as ``LLM memory mechanisms''~\citep{zhong2023memorybankenhancinglargelanguage,packerMemGPTLLMsOperating2023,DBLP:conf/nips/Wang0CLYGW23} are more appropriately interpreted, under contemporary terminology, as early instances of agent memory. This reinterpretation arises from the historical ambiguity surrounding the very notion of an ``LLM agent.'' During 2023–2024, the community had no stable or coherent definition: in some cases, prompting an LLM to call a calculator already sufficed to qualify the system as an agent~\citep{wu2024mathchatconversetacklechallenging}; in other cases, agency required substantially richer capabilities such as explicit planning, tool use, memory, and reflective reasoning~\citep{ruan2023tptulargelanguagemodelbased}. Only recently has a more unified and structured definition begun to emerge (e.g., LLM-based agent = LLM + reasoning + planning + memory + tool use + self-improvement + multi-turn interaction + perception, as discussed by~\citet{zhang2025landscapeagenticreinforcementlearning}), though even this formulation is not universally applicable.
Against this historical backdrop, early systems such as MemoryBank~\citep{zhong2023memorybankenhancinglargelanguage} and MemGPT~\citep{packerMemGPTLLMsOperating2023} framed their contributions as providing \emph{LLM memory}. Yet what they fundamentally addressed were classical agentic challenges, for example enabling an LLM-based conversational agent to track user preferences, maintain dialogue-state information, and accumulate experience across multi-turn interactions. Under a modern and more mature understanding of agency, such systems are naturally categorized as instances of \emph{agent memory}.

That said, the subsumption is not absolute. A distinct line of research genuinely concerns \emph{LLM-internal memory}: managing the transformer’s key–value (KV) cache, designing long-context processing mechanisms, or modifying model architectures (e.g., RWKV~\citep{peng2023rwkvreinventingrnnstransformer}, Mamba~\citep{gu2024mambalineartimesequencemodeling,lieber2024jambahybridtransformermambalanguage}, diffusion-based LMs~\citep{nie2025largelanguagediffusionmodels}) to better retain information as sequence length grows. These works focus on intrinsic model dynamics and typically address tasks that do not require agentic behavior, and thus should be considered outside the scope of agent memory. 

\paragraph{Overlap}
Within our taxonomy, the majority of what has historically been called ``LLM memory'' corresponds to forms of agent memory. Techniques such as \emph{few-shot prompting}~\citep{prabhumoye2022fewshotinstructionpromptspretrained,ma2023fairnessguidedfewshotpromptinglarge} can be viewed as a form of long-term memory, where past exemplars or distilled task summaries serve as reusable knowledge incorporated through retrieval or context injection. \emph{Self-reflection} and iterative refinement methods~\citep{madaan2023self,mousavi2023ncriticsselfrefinementlargelanguage,han2025stitchtimesavesnine} naturally align with short-term, inside-trial memory, as the agent repeatedly leverages intermediate reasoning traces or outcomes from prior attempts within the same task. Even \emph{KV compression} and context-window management~\citep{yoonCompActCompressingRetrieved2024,jiangLLMLinguaCompressingPrompts2023}, when used to preserve salient information across the course of a single task, function as short-term memory mechanisms in an agentic sense. These techniques all support the agent’s ability to accumulate, transform, and reuse information throughout a task’s execution.

\paragraph{Distinctions}
In contrast, memory mechanisms that intervene directly in the model’s internal state—such as architectural modifications for longer effective context, cache rewriting strategies, recurrent-state persistence, attention-sparsity mechanisms, or externalized KV-store expansions—are more appropriately classified as \emph{LLM memory} rather than agent memory. Their goal is to expand or reorganize the representational capacity of the underlying model, not to furnish a decision-making agent with an evolving external memory base. They do not typically support cross-task persistence, environment-driven adaptation, or deliberate memory operations (e.g., formation, evolution, retrieval), and therefore lie outside the operational scope of agent memory as defined in this survey.

\subsubsection{Agent Memory vs. RAG}

At a conceptual level, \emph{agent memory} and \emph{retrieval-augmented generation} (RAG) exhibit substantial overlap: both systems construct, organize, and leverage auxiliary information stores to extend the capabilities of LLM/agents beyond their native parametric knowledge. For instance, structured representations such as knowledge graphs and indexing strategies appear in both communities’ methods, and recent developments in agentic RAG demonstrate how autonomous retrieval mechanisms can interact with dynamic databases in ways reminiscent of agent memory architectures~\citep{singh2025agenticretrievalaugmentedgenerationsurvey}. Indeed, the engineering stacks underlying many RAG and agent memory systems share common building blocks, including vector indices, semantic search, and context expansion modules. 

Despite these technological convergences, the two paradigms have \textit{historically} been distinguished by the contexts in which they are applied. Classical RAG techniques primarily augment an LLM with access to \textbf{static knowledge sources}, whether flat document stores, structured knowledge bases, or large corpora externally indexed to support retrieval on demand~\citep{zhang2025leanragknowledgegraphbasedgenerationsemantic,han2025retrievalaugmentedgenerationgraphsgraphrag}. These systems are designed to ground generation in up-to-date facts, mitigate hallucinations, and improve accuracy in knowledge-intensive tasks, but they generally do not maintain an internal, evolving memory of past interactions. In contrast, agent memory systems are instantiated within an agent’s \textbf{ongoing interaction with an environment}, continuously incorporating new information generated by the agent’s own actions and environmental feedback into a persistent memory base~\citep{wang2024agentworkflowmemory,DBLP:conf/aaai/Zhao0XLLH24,sun2025rearter}.

In early formulations the distinction between RAG and agent memory was relatively clear: RAG retrieved from externally maintained knowledge for a single task invocation, whereas agent memory evolved over multi-turn, multi-task interaction. However, this boundary has become increasingly blurred as retrieval systems themselves become more dynamic. For example, certain retrieval tasks continuously update relevant context during iterative querying (e.g., multi-hop QA settings where related context is progressively added). Interestingly, systems such as HippoRAG/HippoRAG2~\citep{gutiérrez2025hipporagneurobiologicallyinspiredlongterm,gutiérrez2025ragmemorynonparametriccontinual} have been interpreted by both RAG and memory communities as addressing long-term memory challenges for LLMs. Consequently, a more practical (though not perfectly separable) distinction lies in the \textbf{task domain}. RAG is predominantly applied to augment LLMs with large, externally sourced context for individual inference tasks, exemplified by classical multi-hop and knowledge-intensive benchmarks such as HotpotQA~\citep{yang2018hotpotqadatasetdiverseexplainable}, 2WikiMQA~\citep{ho2020constructingmultihopqadataset}, and MuSiQue~\citep{trivedi2022musiquemultihopquestionssinglehop}. By contrast, agent memory systems are typically evaluated in settings requiring sustained multi-turn interaction, temporal dependency, or environment-driven adaptation. Representative benchmarks include long-context dialogue evaluations such as LoCoMo~\citep{maharana2024evaluating} and LongMemEval~\citep{DBLP:conf/iclr/WuWYZCY25}, complex problem-solving and deep-research benchmarks such as GAIA~\citep{mialon2023gaia}, XBench~\citep{chen2025xbenchtrackingagentsproductivity}, and BrowseComp~\citep{wei2025browsecomp}, code-centric agentic tasks such as SWE-bench Verified~\citep{jimenez2023swe}, as well as lifelong learning benchmarks such as StreamBench~\citep{DBLP:conf/nips/WuTLCL24}. We provide a comprehensive summary of memory-related benchmarks in \Cref{sec:benchmarks}.

Nevertheless, even this domain-based distinction contains substantial gray areas. Many works self-described as agent memory systems are evaluated under long-document question-answering tasks such as HotpotQA~\citep{wang2025omem, DBLP:journals/corr/abs-2509-25911}, while numerous papers foregrounded as RAG systems in fact implement forms of agentic self-improvement, continually distilling and refining knowledge or skills over time. As a result, titles, methodologies, and empirical evaluations frequently blur the conceptual boundary between the two paradigms.
To further clarify these relationships, the following three paragraphs draw upon established taxonomies of RAG from~\citep{meiSurveyContextEngineering2025}: \emph{modular RAG}, \emph{graph RAG}, and \emph{agentic RAG}, and examine how the core techniques associated with each lineage manifest within both RAG and agent memory systems.

\paragraph{Modular RAG}
Modular RAG refers to architectures in which the retrieval pipeline is decomposed into clearly specified components, such as indexing, candidate retrieval, reranking, filtering, and context assembly, that operate in a largely static and pipeline-like fashion~\citep{singh2025agenticretrievalaugmentedgenerationsurvey}. These systems treat retrieval as a well-engineered, modular subsystem external to the LLM, designed primarily for injecting relevant knowledge into the model’s context window during inference. Within the agent memory perspective, the corresponding techniques typically appear in the \emph{retrieval stage}, where memory access is realized through vector search, semantic similarity matching, or rule-based filtering, as seen in popular agent memory frameworks like Memary~\citep{githubGitHubKingjulio8238Memary}, MemOS~\citep{li2025memosoperatingmemoryaugmentedgeneration}, and Mem0~\citep{Chhikara2025mem0}.

\paragraph{Graph RAG}
Graph RAG systems structure the knowledge base as a graph, ranging from knowledge graphs to concept graphs or document-entity relations, and leverage graph traversal or graph-based ranking algorithms to retrieve context~\citep{peng2024graphretrievalaugmentedgenerationsurvey}. This representation enables multi-hop relational reasoning, which has proven effective for knowledge-intensive tasks~\citep{edge2025localglobalgraphrag,han2025retrievalaugmentedgenerationgraphsgraphrag,dong2025youtugraphragverticallyunifiedagents}. In the context of agent memory, graph-structured memory arises naturally when agents accumulate relational insights over time, such as linking concepts, tracking dependencies among subtasks, or recording causal relations inferred through interaction. Several well-established practices include Mem0$^g$~\citep{Chhikara2025mem0}, A-MEM~\citep{xuAMEMAgenticMemory2025}, Zep~\citep{Rasmussen2025Zep}, and G-memory~\citep{Zhang2025GMemory}. Notably, graph-based agent memory systems may \emph{construct, extend, or reorganize} its internal graph throughout the agent’s operation. Consequently, graph-based retrieval forms the structural backbone for both paradigms, but only agent memory treats the graph as a living, evolving representation of experience. We provide further analysis on graph-based memory forms in \Cref{sec:token-level-2d} and also refer the readers to a relevant survey~\citep{liu2025graphaugmentedlargelanguagemodel}.

\paragraph{Agentic RAG}
Agentic RAG integrates retrieval into an autonomous decision-making loop, where an LLM agent actively controls when, how, and what to retrieve~\citep{singh2025agenticretrievalaugmentedgenerationsurvey,sun2025rearter}. These systems often employ iterative querying, multi-step planning, or self-directed search procedures, enabling the agent to refine its information needs through deliberate reasoning, as implemented in PlanRAG~\citep{lee2024planragplanthenretrievalaugmentedgeneration} and Self-RAG~\citep{asai2023selfraglearningretrievegenerate}. For a more detailed understanding of agentic RAG, we refer the readers to \cite{singh2025agenticretrievalaugmentedgenerationsurvey}. From the agent memory perspective, agentic RAG occupies the closest conceptual space: both systems involve autonomous interaction with an external information store, both support multi-step refinement, and both may incorporate retrieved insights into subsequent reasoning. The key distinction is that classical agentic RAG typically operates over an \emph{external} and often task-specific database, whereas agent memory maintains an \emph{internal, persistent, and self-evolving} memory base that accumulates knowledge across tasks~\citep{yanGeneralAgenticMemory2025,xuAMEMAgenticMemory2025}.


\subsubsection{Agent Memory vs. Context Engineering}
The relationship between \textit{agent memory} and \textit{context engineering} is best understood as an intersection of distinct operational paradigms rather than a hierarchical subsumption. Context engineering is a systematic design methodology that treats the context window as a constrained computational resource. It rigorously optimizes the information payload, including instructions, knowledge, state, and memory, to mitigate the asymmetry between massive input capacity and the model's generation capability~\citep{meiSurveyContextEngineering2025}. While agent memory focuses on the \textbf{cognitive modeling} of a persistent entity with an evolving identity, context engineering operates under a \textbf{resource management} paradigm. From the perspective of context engineering, agent memory is merely one variable within the context assembly function that requires efficient scheduling to maximize inference efficacy. Conversely, from the perspective of an agent, context engineering serves as the implementation layer that ensures cognitive continuity remains within the physical limits of the underlying model.

\paragraph{Overlap} 
The two fields converge significantly in the technical realization of working memory during long-horizon interactions and often employ functionally identical mechanisms to address the constraints imposed by a finite context window~\citep{huHiAgentHierarchicalWorking2025,DBLP:journals/corr/abs-2510-12635,kangACONOptimizingContext2025,yu2025memagent}. Both paradigms rely on advanced information compression~\citep{zhou2025mem1learningsynergizememory,wuReSumUnlockingLongHorizon2025}, organization~\citep{xuAMEMAgenticMemory2025,Zhang2025GMemory,anokhin2024arigraph}, and selection~\citep{DBLP:journals/corr/abs-2510-12635} techniques to preserve operational continuity over extended interaction sequences. For example, token pruning and importance-based selection methods~\citep{jiangLLMLinguaCompressingPrompts2023,liCompressingContextEnhance2023a} that are central to context engineering frameworks play a fundamental role in agentic memory systems by filtering noise and retaining salient information.
Similarly, the rolling summary technique serves as a shared foundational primitive, functioning simultaneously as a buffer management strategy and a transient episodic memory mechanism~\citep{yu2025memagent,luScalingLLMMultiturn2025}. In practice, the boundary between engineering the context and maintaining an agent's short-term memory effectively dissolves in these scenarios, as both rely on the same underlying summarization, dynamic information retrieval, and recursive state updates~\citep{tangTurnLimitsTraining2025,yoonCompActCompressingRetrieved2024}.

\paragraph{Distinctions}
The distinction becomes most pronounced when moving beyond short-term text processing to the broader scope of long-lived agents. Context engineering primarily addresses the \emph{structural organization} of the interaction interface between LLMs and their operational environment. This includes optimizing tool-integrated reasoning and selection pipelines~\citep{qin2024toolllm,schick2023toolformer,jiaAutoToolEfficientTool2025} and standardizing communication protocols, such as MCP~\citep{qiu2025alitageneralistagentenabling}. These methods focus on ensuring that instructions, tool calls, and intermediate states are correctly formatted, efficiently scheduled, and executable within the constraints of the context window. As such, context engineering operates at the level of \emph{resource allocation and interface correctness}, emphasizing syntactic validity and execution efficiency.

In contrast, agent memory defines a substantially broader cognitive scope. Beyond transient context assembly, it encompasses the persistent storage of factual knowledge~\citep{zhong2023memorybankenhancinglargelanguage}, the accumulation and evolution of experiential traces~\citep{DBLP:conf/aaai/Zhao0XLLH24,tang2025agentkbleveragingcrossdomain,Zhang2025MemGen}, and, in some cases, the internalization of memory into model parameters~\citep{self-param}. Rather than managing how information is presented to the model at inference time, agent memory governs what the agent \emph{knows}, what it \emph{has experienced}, and how these elements evolve over time. This includes consolidating repeated interactions into knowledge~\citep{DBLP:conf/acl/rmm2025}, abstracting procedural knowledge from past successes and failures~\citep{ouyang2025reasoningbankscalingagentselfevolving}, and maintaining a coherent identity across tasks and episodes~\citep{wang2024aipersona}.

From this perspective, context engineering constructs the external scaffolding that enables perception and action under resource constraints, whereas agent memory constitutes the internal substrate that supports learning, adaptation, and autonomy. The former optimizes the momentary interface between the agent and the model, while the latter sustains a persistent cognitive state that extends beyond any single context window.

\section{Form: What Carries Memory?}
\label{sec:what-memory}
As a starting point for organizing prior work, we begin by examining the most fundamental representational units out of which agent memory can be constructed. We first try to answer: what architectural or representational forms can agent memory take?

Across diverse agent systems, memory is not realized through a single, unified structure. Instead, different task settings call for different storage forms, each with its own structural properties. These architectures endow memory with distinct capabilities, shaping how an agent accumulates information over interactions and maintains behavioral consistency. They ultimately enable memory to fulfill its intended roles across varied task scenarios.

Based on where memory resides and in what form it is represented, we organize these memories into three categories:

\vspace{0.6em}
\begin{tcolorbox}[
  colback=selfevolagent_light!20,
  colframe=selfevolagent_light!80,
  colbacktitle=selfevolagent_light!80,
  coltitle=black,
  title={\bfseries\fontfamily{ppl}\selectfont{Three Major Memory Forms}},
  boxrule=2pt,
  arc=5pt,
  drop shadow,
  parbox=false,
  before skip=5pt,
  after skip=5pt,
  left=5pt,   
  right=5pt,
]
\begin{enumerate}
    \item \textbf{Token-level Memory} (\Cref{ssec:token}): Memory organized as explicit and discrete units that can be individually accessed, modified, and reconstructed. These units remain externally visible and can be stored in a structured form over time.
    
    \item \textbf{Parametric Memory} (\Cref{ssec:parametric}): Memory stored within the model parameters, where information is encoded through the statistical patterns of the parameter space and accessed implicitly during forward computation.

    \item \textbf{Latent Memory} (\Cref{ssec:latent}): Memory represented in the model’s internal hidden states, continuous representations, or evolving latent structures. It can persist and update during inference or across interaction cycles, capturing context-dependent internal states.
\end{enumerate}
\end{tcolorbox}

The three memory forms outlined above establish the core structural framework for understanding ``what carries memory''. Each form organizes, stores, and updates information in its own way, giving rise to distinct representational patterns and operational behaviors. With this structural taxonomy in place, we can more systematically examine why agents need memory (\Cref{sec:why-memory}) and how memory evolves, adapts, and shapes agent behavior over sustained interactions (\Cref{sec:how-memory}). This classification provides the conceptual foundation for the discussions that follow.

\subsection{Token-level Memory}\label{ssec:token}

\vspace{0.6em}
\begin{tcolorbox}[
  colback=selfevolagent_light!20,
  colframe=selfevolagent_light!80,
  colbacktitle=selfevolagent_light!80,
  coltitle=black,
  title={\bfseries\fontfamily{ppl}\selectfont{Definition of Token-level Memory}},
  boxrule=2pt,
  arc=5pt,
  drop shadow,
  parbox=false,
  before skip=5pt,
  after skip=5pt,
  left=5pt,   
  right=5pt,
]
Token-level memory stores information as persistent, discrete units that are externally accessible and inspectable. The token here is a broad representational notion: beyond text tokens, it includes visual tokens, audio frames—any discrete element that can be written, retrieved, reorganized, and revised outside model parameters.
\end{tcolorbox}

Because these units are explicit, token-level memory is typically transparent, easy to edit, and straightforward to interpret, making it a natural layer for retrieval, routing, conflict handling, and coordination with parametric and latent memory. Token-level memory is also the most common memory form and the one with the largest body of existing work.

Although all token-level memories share the property of being stored as discrete units, they differ significantly in how these units are organized. The structural organization of stored tokens plays a central role in determining how efficiently the agent can search, update, or reason over past information. To describe these differences, we categorize token-level memory by inter-unit structural organization, moving from no explicit topology to multi-layer topologies:

\begin{figure}[!t]
    \centering
    \includegraphics[width=1\linewidth]{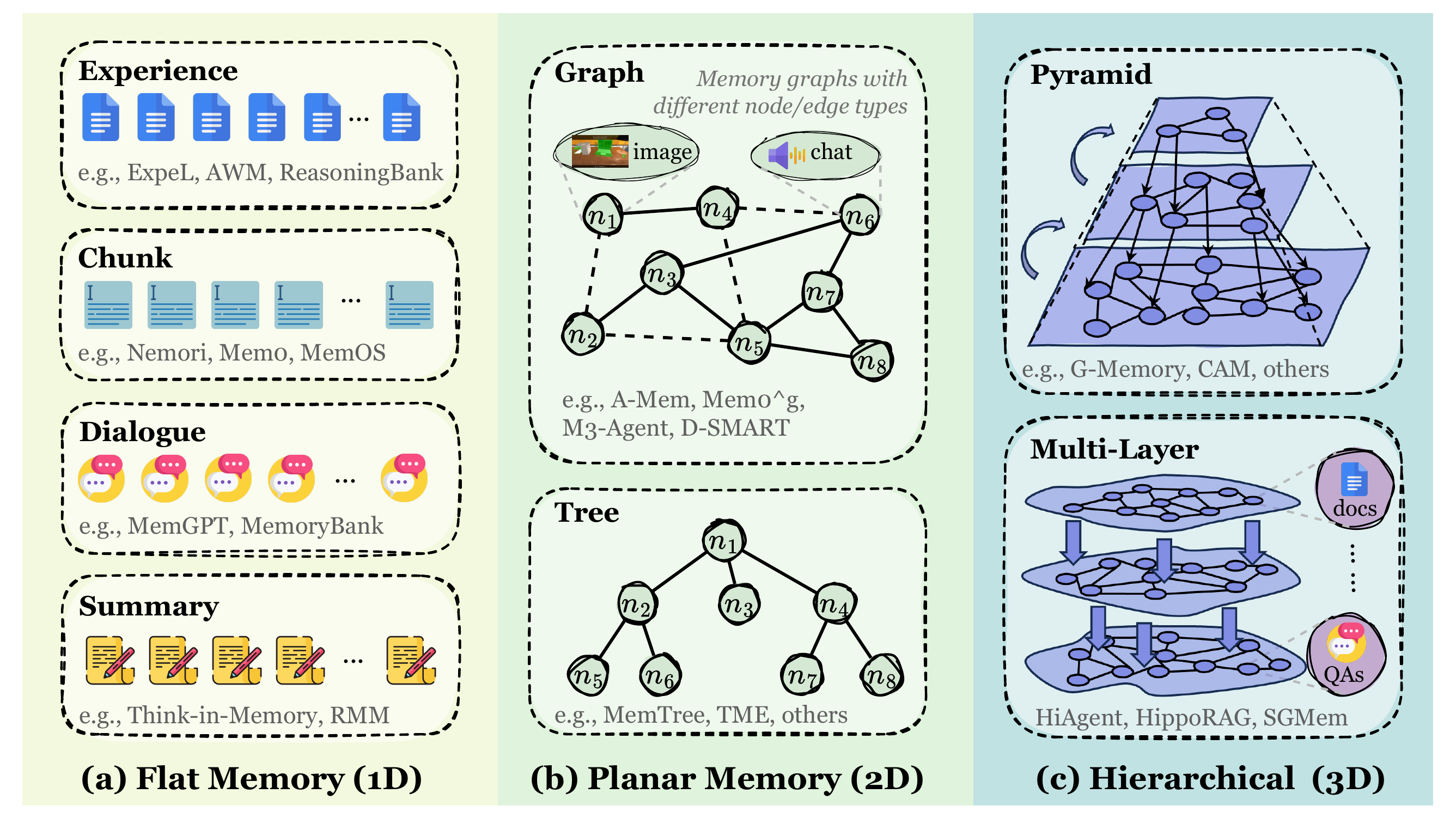}
    \caption{Taxonomy of token-level memory organized by topological complexity and dimensionality: 
    (a) \textbf{Flat Memory (1D)} stores information as linear sequences or independent clusters without explicit inter-unit topology, commonly used for \textit{Chunk} sets, \textit{Dialogue} logs, and \textit{Experience} pools. 
    (b) \textbf{Planar Memory (2D)} introduces a single-layer structured layout where units are linked via \textbf{Tree} or \textbf{Graph} structures to capture relational dependencies, supporting diverse node types such as images and chat records. 
    (c) \textbf{Hierarchical Memory (3D)} employs multi-level forms, such as \textbf{Pyramids} or \textbf{Multi-layer} graphs, to facilitate vertical abstraction and cross-layer reasoning between different data granularities, such as raw docs and synthesized QAs.}
    
    \label{fig:sec3-3.1}
\end{figure}

\vspace{0.6em}
\begin{tcolorbox}[
  colback=selfevolagent_light!20,
  colframe=selfevolagent_light!80,
  colbacktitle=selfevolagent_light!80,
  coltitle=black,
  title={\bfseries\fontfamily{ppl}\selectfont{Three Major Types of Token-level Memory}},
  boxrule=2pt,
  arc=5pt,
  drop shadow,
  parbox=false,
  before skip=5pt,
  after skip=5pt,
  left=5pt,   
  right=5pt,
]
\begin{enumerate}
    \item \textbf{Flat Memory (1D)}: No explicit inter-unit topology. Memories are accumulated as sequences or bags of units (e.g., snippets, trajectories, chunks)
    
    \item \textbf{Planar Memory (2D)}: A structured but single-layer organization within one plane: units are related by a graph, tree, table and so on, with no cross-layer relations. The structure is explicit, but not layered.

    \item \textbf{Hierarchical Memory (3D)}: Structured across multiple layers with inter-layer links, forming a volumetric or stratified memory 
\end{enumerate}
\end{tcolorbox}

The three types of token-level memory are clearly illustrated in \autoref{fig:sec3-3.1}. From Flat Memory with no topology, to Planar Memory with single-layer structural organization, to Hierarchical Memory with multi-layer interlinked structures, this organizational spectrum governs not only how token-level memory supports search, update, and reasoning, but also how the memory itself is structured and what capabilities it affords. In the subsections that follow, we introduce each organizational form in terms of its strengths and limitations, typical use cases, and representative work. The summary and comparison of representative token-level memory methods are presented in \autoref{tab:token-mem}.

It is worth noting that, following the idea introduced by ReAct~\citep{yao2023react}, a series of studies began focusing on long-horizon interaction tasks~\citep{DBLP:journals/corr/abs-2503-05592,DBLP:journals/corr/abs-2503-09516,DBLP:journals/corr/abs-2501-05366,DBLP:journals/corr/abs-2507-02592,DBLP:journals/corr/abs-2504-21776,DBLP:journals/corr/abs-2505-22648}. Many of these tasks introduce an explicit notion of memory, and because the memory is generally stored in plaintext form, they fall within the scope of token-level memory. Most of them emphasize how to compress or fold accumulated interaction traces so that agents can operate over long sequences without exceeding context limits~\citep{zhou2025mem1learningsynergizememory,DBLP:journals/corr/abs-2510-12635,wuReSumUnlockingLongHorizon2025,DBLP:journals/corr/abs-2510-11967,liDeepAgentGeneralReasoning2025,chen2025iterresearchrethinkinglonghorizonagents}. A more detailed discussion is provided in \Cref{ssec:working_mem} about working memory.

\subsubsection{Flat Memory (1D)}

\vspace{0.6em}
\begin{tcolorbox}[
  colback=selfevolagent_light!20,
  colframe=selfevolagent_light!80,
  colbacktitle=selfevolagent_light!80,
  coltitle=black,
  title={\bfseries\fontfamily{ppl}\selectfont{Definition of Flat (1D) Memory}},
  boxrule=2pt,
  arc=5pt,
  drop shadow,
  parbox=false,
  before skip=5pt,
  after skip=5pt,
  left=5pt,   
  right=5pt,
]
Flat Memory stores information as accumulations of discrete units, without explicitly modeling semantic or relational dependencies among them. These units may include text chunks, user profiles, experience trajectories, their corresponding vector representations, or multimodal entries. Relationships among these units are not encoded directly in the memory.
\end{tcolorbox}

To facilitate a clear and coherent presentation, we group prior work on flat memory according to their primary design objectives and technical emphases. This grouping serves \textbf{an organizational purpose} and does not imply that the resulting categories are strictly parallel or mutually exclusive. In practice, certain methods may be applicable to multiple categories, and some approaches involving multimodal information may be discussed in other sections when multimodality is not their central focus. Such an organization allows us to systematically review the literature while preserving flexibility in interpretation.

\setlength{\tabcolsep}{4pt} 
\renewcommand{\arraystretch}{1.1} 

\begin{center}
\footnotesize
\begin{longtable}{p{5cm}|llp{5.5cm}p{3.3cm}}
\caption{
Comparison of representative token-level memory methods.
We categorize existing works into three groups based on their topological complexity: \textbf{Flat Memory (1D)} for linear or independent records, \textbf{Planar Memory (2D)} for structured single-layer graphs/trees, and \textbf{Hierarchical Memory (3D)} for multi-level architectures.
Methods are characterized across four dimensions:
(1) \textbf{Multi} indicates multimodal capability, where {\color{green}\ding{52}} denotes support for modalities beyond text (e.g., visual) and {\color{red}\ding{55}} implies text-only;
(2) \textbf{Type} identifies the specific functional category of the memory (e.g., \textit{Fact} for factual memory, \textit{Exp} for experiential memory, \textit{Work} for working memory );
(3) \textbf{Memory Form} details the content of the stored units;
and (4) \textbf{Task} lists the primary application domains.
}\\

\toprule
{Method} & \makecell{Multi} & \makecell{Type} & \makecell{Memory Form} & {Task} \\
\midrule
\endfirsthead

\caption[]{
Comparison of representative token-level memory methods.
We categorize existing works into three groups based on their topological complexity: \textbf{Flat Memory (1D)} for linear or independent records, \textbf{Planar Memory (2D)} for structured single-layer graphs/trees, and \textbf{Hierarchical Memory (3D)} for multi-level architectures.
Methods are characterized across four dimensions:
(1) \textbf{Multi} indicates multimodal capability, where {\color{green}\ding{52}} denotes support for modalities beyond text (e.g., visual) and {\color{red}\ding{55}} implies text-only;
(2) \textbf{Type} identifies the specific functional category of the memory (e.g., \textit{Fact} for factual memory, \textit{Exp} for experiential memory, \textit{Work} for working memory );
(3) \textbf{Memory Structure} details the organization mechanism of the stored units;
and (4) \textbf{Task} lists the primary application domains.
(continued)
}\label{tab:token-mem}\\
\toprule
{Method} & \makecell{Multi} & \makecell{Type} & \makecell{Memory Structure} & {Task} \\
\midrule
\endhead

\midrule
\multicolumn{5}{r}{Continued on next page}\\
\bottomrule
\endfoot

\bottomrule
\endlastfoot

\multicolumn{5}{c}{\textit{Flat Memory Models}}\\
\midrule
Reflexion~\citep{shinn_reflexion_2023} & \color{red}\ding{55} & E\&W & Trajectory as short-term and feedback as long-term & QA, Reasoning, Coding \\
Memento~\citep{DBLP:journals/corr/abs-2508-16153} & \color{red}\ding{55} & Exp & Trajectory case (success/failure). & Reasoning \\
JARVIS-1~\citep{DBLP:journals/pami/WangCLJHZLHZYML25} & \color{green}\ding{52} & Exp & Plan-environment pairs. & Game \\
Expel~\citep{DBLP:conf/aaai/Zhao0XLLH24} & \color{red}\ding{55} & Exp & Insights and few-shot examples. & Reasoning \\
Buffer of Thoughts~\citep{yang2024buffer} & \color{red}\ding{55} & Exp & High-level thought-templates. & Game, Reasoning, Coding \\
SAGE~\citep{DBLP:journals/ijon/SAGE} & \color{red}\ding{55} & Exp & Dual-store with forgetting mechanism. & Game, Reasoning, Coding \\
ChemAgent~\citep{tang2025chemagentselfupdatinglibrarylarge} & \color{red}\ding{55} & Exp & Structured sub-tasks and principles. & Chemistry \\
AgentKB~\citep{tang2025agentkbleveragingcrossdomain} & \color{red}\ding{55} & Exp & 5-tuple experience nodes. & Coding, Reasoning \\
H\({}^{\mbox{2}}\)R~\citep{DBLP:journals/corr/abs-2509-12810} & \color{red}\ding{55} & Exp & Planning and Execution layers. & Game, Embodied Simulation \\
AWM~\citep{wang2024agentworkflowmemory} & \color{red}\ding{55} & Exp & Abstracted universal workflows. & Web \\
PRINCIPLES~\citep{DBLP:journals/corr/abs-2509-17459} & \color{red}\ding{55} & Exp & Rule templates from self-play. & Emotional Companion \\
ReasoningBank~\citep{ouyang2025reasoningbankscalingagentselfevolving} & \color{red}\ding{55} & Exp & Transferable reasoning strategy items. & Web \\
Voyager~\citep{wang_voyager_2024} & \color{green}\ding{52} & Exp & Executable skill code library. & Game \\
DGM~\citep{zhang_darwin_2025} & \color{red}\ding{55} & Exp & Recursive self-modifiable codebase. & Coding \\
Memp~\citep{fang2025mempexploringagentprocedural} & \color{red}\ding{55} & Exp & Instructions and abstract scripts. & Embodied Simulation, Travel Planning \\
UFO2~\citep{DBLP:journals/corr/abs-2504-14603} & \color{green}\ding{52} & Exp & System docs and interaction records. & Windows OS \\
LEGOMem~\citep{han_legomem_2025} & \color{red}\ding{55} & Exp & Vectorized task trajectories. & Office\\
ToolMem~\citep{xiao_toolmem_2025} & \color{red}\ding{55} & Exp & Tool capability. & Tool Calling \\
SCM~\citep{wang2025scmenhancinglargelanguage} & \color{red}\ding{55} & Fact & Memory stream and vector database. & Long-context \\
MemoryBank~\citep{zhong2023memorybankenhancinglargelanguage} & \color{red}\ding{55} & Fact & History and user profile. & Emotional Companion \\
MPC~\citep{DBLP:conf/acl/LeeHPP023} & \color{red}\ding{55} & Fact & Persona and summary vector pool. & QA
\\
RecMind~\citep{wang_recmind_2024} & \color{red}\ding{55} & Fact & User metadata and external knowledge. & Recommendation \\
InteRecAgent~\citep{DBLP:journals/tois/HuangLLYLX25} & \color{red}\ding{55} & Fact & User profiles and candidate item. & Recommendation \\
Ego-LLaVA~\citep{shenEncodeStoreRetrieveAugmentingHuman2024} & \color{green}\ding{52} & Fact & Language-encoded chunk embeddings. & Multimodal QA \\
ChatHaruhi~\citep{DBLP:journals/corr/abs-2308-09597} & \color{red}\ding{55} & Fact & Dialogue database from media. & Role-Playing \\
Memochat~\citep{lu2023memochattuningllmsuse} & \color{red}\ding{55} & Fact & Memos and categorized dialogue history. & Long-conv QA \\
RecursiveSum~\citep{wang2025recursivelysummarizingenableslongterm} & \color{red}\ding{55} & Fact & Recursive summaries of short dialogues. & Long-conv QA \\
MemGPT~\citep{packerMemGPTLLMsOperating2023} & \color{red}\ding{55} & Fact & Virtual memory (Main/External contexts). & Long-conv QA, Doc QA \\
RoleLLM~\citep{DBLP:conf/acl/WangPQLZWGGN00024} & \color{red}\ding{55} & Fact & Role-specific QA pairs. & Role-Playing \\
Think-in-memory~\citep{DBLP:journals/corr/abs-2311-08719} & \color{red}\ding{55} & Fact & Hash table of inductive thoughts. & Long-conv QA \\
PLA~\citep{DBLP:conf/coling/YuanSLWCL25} & \color{red}\ding{55} & Fact & Evolving records of history and summaries. & QA, Human Feedback \\
COMEDY~\citep{chen2025compress} & \color{red}\ding{55} & Fact & Single-model compressed memory format. & Summary, Compression, QA \\
Memoro~\citep{DBLP:conf/chi/ZulfikarCM24} & \color{green}\ding{52} & Fact & Speech-to-text vector embeddings. & User Study \\
Memory Sharing~\citep{DBLP:journals/corr/abs-2404-09982} & \color{red}\ding{55} & Fact & Query-Response pair retrieval. & Literary Creation, Logic, Plan Generation \\
Conv Agent\citep{DBLP:journals/corr/abs-2406-00057} & \color{red}\ding{55} & Fact & Chain-of-tables and vector entries. & QA \\
EM-LLM~\citep{DBLP:conf/iclr/FountasBOCLB025} & \color{red}\ding{55} & Fact & Episodic events with Bayesian boundaries. & Long-context \\
Memocrs~\citep{DBLP:conf/cikm/XiLL0T0024} & \color{red}\ding{55} & Fact & User metadata and knowledge. & Recommendation \\
SECOM~\citep{panSeCom2025} & \color{red}\ding{55} & Fact & Paragraph-level segmented blocks. & Long-conv QA \\
Mem0~\citep{Chhikara2025mem0} & \color{red}\ding{55} & Fact & Summary and original dialogue. & Long-conv QA \\
RMM~\citep{DBLP:conf/acl/rmm2025} & \color{red}\ding{55} & Fact & Reflection-organized flat entries. & Personalization \\
MEMENTO~\citep{DBLP:journals/corr/abs-2505-16348} & \color{green}\ding{52} & Fact & Interaction history entries. & Personalization \\
MemGuide~\citep{duMemGuideIntentDrivenMemory2025} & \color{red}\ding{55} & Fact & Dialogue-derived QA pairs. & Long-conv QA \\
MIRIX~\citep{wang2025mirixmultiagentmemoryllmbased} & \color{green}\ding{52} & Fact & Six optimized flat memory types. & Long-conv QA\\
SemanticAnchor~\citep{DBLP:journals/corr/SemanticAnchor} & \color{red}\ding{55} & Fact & Syntactic 5-tuple structure. & Long-conv QA \\
MMS~\citep{DBLP:journals/corr/abs-2508-15294} & \color{red}\ding{55} & Fact & Dual Retrieval and Context units. & Long-conv QA \\
Memory-R1~\citep{yan2025memory} & \color{red}\ding{55} & Fact & RL-managed mem0 architecture. & Long-conv QA \\
ComoRAG~\citep{DBLP:journals/corr/comorag} & \color{red}\ding{55} & Fact & Fact/Semantic/Plot units with probes. & Narrative QA \\
Nemori~\citep{DBLP:journals/corr/abs-2508-03341} & \color{red}\ding{55} & Fact & Predictive calibration store. & Long-conv QA \\
Livia~\citep{DBLP:journals/corr/abs-2509-05298} & \color{green}\ding{52} & Fact & Pruned interaction history. & Emotional Companion \\
MOOM~\citep{chen2025moommaintenanceorganizationoptimization} & \color{red}\ding{55} & Fact & Decoupled plot and character stores. & Role-Playing \\
Mem-$\alpha$~\citep{DBLP:journals/corr/abs-2509-25911} & \color{red}\ding{55} & Fact & Core, Semantic, and Episodic Mem. & Memory Management \\
Personalized Long term Interaction~\citep{DBLP:journals/corr/abs-2510-07925} & \color{red}\ding{55} & Fact & Hierarchical history and summaries. & Personalization \\
LightMem~\citep{fang2025lightmem} & \color{red}\ding{55} & Fact & Optimized Long/Short-term store. & Long-conv QA \\
MEXTRA~\citep{wang2025unveilingprivacyrisksllm} & \color{red}\ding{55} & Fact & Extracted raw dialogue data. & Privacy Attack \\
MovieChat~\citep{DBLP:conf/cvpr/SongCWZZWCG0ZLH24} & \color{green}\ding{52} & Fact & Short-term features and long-term persistence. & Video Understanding \\
MA-LMM~\citep{heMALMMMemoryAugmentedLarge2024} & \color{green}\ding{52} & Fact & Visual and Query memory banks. & Video Understanding \\
VideoAgent~\citep{wangVideoAgentLongFormVideo2024} & \color{green}\ding{52} & Fact & Temporal text descriptions and object tracking. & Video Understanding \\
Video-RAG~\citep{luo2025videoragvisuallyalignedretrievalaugmentedlong} & \color{green}\ding{52} & Fact & Visually-aligned information . & Video Understanding \\
KARMA~\citep{wangKARMAAugmentingEmbodied2025} & \color{green}\ding{52} & Fact & 3D scene graph and dynamic object states. & Embodied Task \\
Embodied VideoAgent~\citep{DBLP:journals/corr/abs-2501-00358} & \color{green}\ding{52} & Fact & Persistent object and sensor store. & MultiModal \\
Mem2Ego~\citep{DBLP:journals/corr/abs-2502-14254} & \color{green}\ding{52} & Fact & Map, landmark, and visited location stores. & Embodied Navigation \\
Context-as-Memory~\citep{DBLP:journals/corr/abs-2506-03141} & \color{green}\ding{52} & Fact & Generated context frames. & Video Generation \\
RCR-Router~\citep{DBLP:journals/corr/rcrrouter} & \color{red}\ding{55} & Fact & Budget-aware semantic subsets. & QA \\
ELL~\citep{cai2025experiencedriven} & \color{red}\ding{55} & Fact & Liflong memory and skills. & Lifelong Learning \\
MemRL~\citep{zhang2026memrlselfevolvingagentsruntime} & \color{red}\ding{55} & Exp & RL for memory management. & Web \\
ReMe~\citep{cao2025remembermerefineme}& \color{red}\ding{55} & Exp & Step level experience and insight. & Web \\
MMAG~\citep{zeppieri2025mmagmixedmemoryaugmentedgeneration} & \color{red}\ding{55} & Fact & Five interacting memory layers. & User Study \\
Hindsight~\citep{latimer2025hindsight2020buildingagent} & \color{red}\ding{55} & Fact & Retains, recalls, and reflects. & Long-conv QA \\
GAM~\citep{yan2025generalagenticmemorydeep} & \color{red}\ding{55} & Fact & Simple memory but search is guided. & Long-conv QA \\

\midrule
\multicolumn{5}{c}{\textit{Planar Memory Models}}\\
\midrule
D-SMART~\citep{Lei2025DSMART} & \color{red}\ding{55} & Fact & Structured memory with reasoning trees. & Long-conv QA \\
Reflexion~\citep{shinn_reflexion_2023} & \color{red}\ding{55} & Work & Reflective text buffer from experiences. & QA, Reasoning, Coding \\
PREMem~\citep{DBLP:journals/corr/PREMem} & \color{red}\ding{55} & Fact & Dynamic cross-session linked triples. & Long-conv QA \\
Query Reconstruct~\citep{xu2025memoryaugmentedqueryreconstructionllmbased} & \color{red}\ding{55} & Exp & Logic graphs built from knowledge bases. & KnowledgeGraph QA \\
KGT~\citep{sun2024knowledgegraphtuningrealtime} & \color{red}\ding{55} & Fact & KG node from query and feedback. & QA \\
Optimus-1~\citep{li2024optimus1hybridmultimodalmemory} & \color{green}\ding{52} & F\&E & Knowledge graph and experience pool. & Game \\
SALI~\citep{pan2024planningimaginationepisodicsimulation} & \color{green}\ding{52} & Exp & Topological graph with spatial nodes & Navigation \\
HAT~\citep{a2024enhancinglongtermmemoryusing} & \color{red}\ding{55} & Fact & Hierarchical aggregate tree. & Long-conv QA \\
MemTree~\citep{rezazadehIsolatedConversationsHierarchical2025} & \color{red}\ding{55} & Fact & Dynamic hierarchical conversation tree. & Long-conv QA \\
TeaFarm~\citep{ong2025lifelongdialogueagentstimelinebased} & \color{red}\ding{55} & Fact & Causal edges connecting memories. & Long-conv QA \\
COMET~\citep{kim2024commonsenseaugmentedmemoryconstructionmanagement} & \color{red}\ding{55} & Fact & Context-aware memory through graph. & Long-conv QA \\
Intrinsic Memory~\citep{yuen2025intrinsicmemoryagentsheterogeneous} & \color{red}\ding{55} & Fact & Private internal and shared external mem. & Planning \\
A-MEM~\citep{xuAMEMAgenticMemory2025} & \color{red}\ding{55} & Fact & Card-based connected mem. & Long-conv QA \\
Ret-LLM~\citep{modarressi2023ret} & \color{red}\ding{55} & Fact & Triplet table and LSH vectors. & QA \\
HuaTuo~\citep{wang2023huatuotuningllamamodel} & \color{red}\ding{55} & Fact & Medical Knowledge Graph. & Medical QA \\
M3-Agent~\citep{long2025seeing} & \color{green}\ding{52} & Fact & Multimodal nodes in graph structure. & Embodied QA \\
EMem~\citep{zhou2025ememsimplestrongbaselinelongterm} & \color{red}\ding{55} & Fact & Event-centric alternative with pagerank. & Long-conv QA \\
WorldMM~\citep{yeo2025worldmmdynamicmultimodalmemory} & \color{green}\ding{52} & Fact & Multiple complementary memories. & Video Understanding \\
Memoria~\citep{sarin2025memoriascalableagenticmemory} & \color{red}\ding{55} & Fact & Knowledge-graph profile and summary. & Long-conv QA \\
LingoEDU~\citep{zhou2026contextedusfaithfulstructured}& \color{red}\ding{55} & Fact & Relation tree of Elementary Discourse Units. & Long-conv QA \\
\midrule
\multicolumn{5}{c}{\textit{Hierarchical Memory Models}}\\
\midrule

GraphRAG~\citep{edge2025localglobalgraphrag} & \color{red}\ding{55} & Fact & Multi-level community graph indices. & QA, Summarization \\
H-Mem~\citep{Sun2025HMEM} & \color{red}\ding{55} & Fact & Decoupled index layers and content layers. & Long-conv QA \\
EMG-RAG~\citep{wang2024craftingpersonalizedagentsretrievalaugmented} & \color{red}\ding{55} & Fact & Three-tiered memory graph. & QA \\
G-Memory~\citep{Zhang2025GMemory} & \color{red}\ding{55} & Exp & Query-centric three-layer graph structure. & QA, Game, Embodied Task \\
Zep~\citep{Rasmussen2025Zep} & \color{red}\ding{55} & Fact & Temporal Knowledge Graphs. & Long-conv QA \\
SGMem~\citep{Wu2025SGMemSG} & \color{red}\ding{55} & Fact & Chunk Graph and Sentence Graph. & Long-conv QA \\
HippoRAG~\citep{gutiérrez2025hipporagneurobiologicallyinspiredlongterm} & \color{red}\ding{55} & Fact & Knowledge with query nodes. & QA \\
HippoRAG 2~\citep{gutiérrez2025ragmemorynonparametriccontinual} & \color{red}\ding{55} & Fact & KG with phrase and passage. & QA \\
AriGraph~\citep{anokhin2024arigraph} & \color{red}\ding{55} & Fact & Semantic and Episodic memory graph. & Game \\
Lyfe Agents~\citep{kaiya2023lyfeagentsgenerativeagents} & \color{red}\ding{55} & Fact & Working, Short \& Long-term layers. & Social Simulation \\
CAM~\citep{liCAMConstructivistView2025} & \color{red}\ding{55} & Fact & Multilayer graph with topic. & Doc QA \\
HiAgent~\citep{huHiAgentHierarchicalWorking2025} & \color{red}\ding{55} & E\&W & Goal graphs with recursive cluster. & Agentic Tasks \\
ILM-TR~\citep{tang2024enhancinglongcontextperformance} & \color{red}\ding{55} & Fact & Hierarchical Memory tree. & Long-context \\
CompassMem~\citep{hu2026memorymattersmoreeventcentric} & \color{red}\ding{55} & Fact & Hierarchical event-centric Memory. & QA \\
MAGMA~\citep{jiang2026magmamultigraphbasedagentic} & \color{red}\ding{55} & Fact & Semantic, temporal, causal, entity graphs. & Long-conv QA \\
EverMemOS~\citep{hu2026evermemosselforganizingmemoryoperating} & \color{red}\ding{55} & Fact & Reusable memories covering multi types. & Long-conv QA \\
RGMem~\citep{tian2025RGMem} & \color{red}\ding{55} & Fact & Renormalization Group-based memory. & Long-conv QA \\
MemVerse~\citep{liu2025memversemultimodalmemorylifelong} & \color{green}\ding{52} & Fact & Multimodal hierarchical knowledge graphs. & Reasoning, QA \\
\end{longtable}
\end{center}

\paragraph{Dialogue}
Some flat memory work focuses on storing and managing dialogue content. Early approaches primarily focused on preventing forgetting by storing raw dialogue history or generating recursive summaries to extend context windows~\citep{wang2025scmenhancinglargelanguage,lu2023memochattuningllmsuse,wang2025recursivelysummarizingenableslongterm,DBLP:conf/coling/YuanSLWCL25}. MemGPT~\citep{packerMemGPTLLMsOperating2023} introduces an operating-system metaphor with hierarchical management, inspiring subsequent works~\citep{li2025memosoperatingmemoryaugmentedgeneration,kang2025memoryosaiagent} to decouple active context from external storage for infinite context management.

To improve retrieval precision, the granularity and structure of memory units have become increasingly diverse and cognitively aligned. Some works, like COMEDY~\citep{chen2025compress}, Memory Sharing~\citep{DBLP:journals/corr/abs-2404-09982} and MemGuide~\citep{duMemGuideIntentDrivenMemory2025} compress information into compact semantic representations or query-response pairs to facilitate direct lookup, while others, like \citet{DBLP:journals/corr/abs-2406-00057} and MIRIX~\citep{wang2025mirixmultiagentmemoryllmbased} adopt hybrid structures ranging from vector-table combinations to multi-functional memory types. Furthermore, research has begun to define memory boundaries based on cognitive psychology, organizing information through syntactic tuples~\citep{DBLP:journals/corr/SemanticAnchor} or segmenting events based on Bayesian surprise and paragraph structures~\citep{DBLP:conf/iclr/FountasBOCLB025, panSeCom2025} , thereby matching human-like cognitive segmentation.

As conversational depth increases, memory evolves to store high-level cognitive processes and narrative complexities. Instead of mere factual records, systems like Think-in-Memory~\citep{DBLP:journals/corr/abs-2311-08719} and RMM~\citep{DBLP:conf/acl/rmm2025} store inductive thoughts and retrospective reflections to guide future reasoning. In complex scenarios such as role-playing or long narratives, approaches like ComoRAG~\citep{DBLP:journals/corr/comorag} and MOOM~\citep{chen2025moommaintenanceorganizationoptimization} decompose memory into factual, plot-level, and character-level components, ensuring the agent maintains a coherent persona and understanding across extended interactions.

Memory has transitioned from static storage to autonomous and adaptive optimization. Mem0\citep{Chhikara2025mem0} established standardized operations for memory maintenance, laying the foundation for intelligent control. Recent advances introduce reinforcement learning to optimize memory construction~\citep{yan2025memory,DBLP:journals/corr/abs-2509-25911}, while other mechanisms focus on dynamic calibration and efficiency, such as predicting missing information~\citep{DBLP:journals/corr/abs-2508-03341}, managing token budgets across multi-agent systems~\citep{DBLP:journals/corr/rcrrouter} , and reducing redundancy in long-term storage~\citep{fang2025lightmem}.

\paragraph{Preference}
Some memory systems focus on modeling a user’s evolving tastes, interests, and decision patterns, especially in recommendation scenarios where preference understanding is central. Unlike dialogue-centric memory, which focuses on maintaining conversational coherence, preference memory centers on identifying a user's tastes and tendencies. Early efforts such as RecMind ~\citep{wang_recmind_2024} separate user-specific information from external domain knowledge by storing both factual user attributes and item metadata. InteRecAgent ~\citep{DBLP:journals/tois/HuangLLYLX25} folds memory into the recommendation workflow but focuses more on the current candidate set, keeping user profiles and the active item pool to support context-aware recommendations. MR.Rec ~\citep{DBLP:journals/corr/abs-2510-14629} builds a memory index archiving the full interaction process, storing raw item information and per-category preference summaries. In conversational settings, Memocrs ~\citep{DBLP:conf/cikm/XiLL0T0024} proposes a more structured design with a user-specific memory tracking entities and user attitudes, and a general memory aggregating cross-user knowledge.

\paragraph{Profile} 
A subset of flat memory systems focuses on storing and maintaining stable user profiles, character attributes, or long-term identity information so that agents can behave consistently across turns and tasks. MemoryBank~\citep{zhong2023memorybankenhancinglargelanguage} represents one of the earliest frameworks in this direction: it organizes dialogue history and event summaries by timestamp, gradually building a user profile that supports accurate retrieval of identity-relevant information. AI Persona~\citep{wang2024aipersona} makes the memory system process information not only presented in the dialogue context but also from multi-dimensional human-AI interaction dimensions. MPC~\citep{DBLP:conf/acl/LeeHPP023} extends this idea by storing real-time persona information and dialogue summaries in a memory pool, keeping conversation behavior aligned with a consistent persona over long interactions. \citet{DBLP:journals/corr/abs-2510-07925} proposes a more comprehensive profile-maintenance mechanism, combining long-term and short-term memory with automatically generated summaries after each turn to form a mid-term context, allowing the user profile to evolve continuously through interaction.

In virtual role-playing settings, ChatHaruhi~\citep{DBLP:journals/corr/abs-2308-09597} extracts dialogue from novels and television scripts, enabling the model to maintain character-consistent behavior by retrieving memory. RoleLLM~\citep{DBLP:conf/acl/WangPQLZWGGN00024} takes a more structured approach by building question–answer pairs to capture character-specific knowledge.

\paragraph{Experience}
Distinct from the static, general knowledge, experience memory stems from the agent's dynamic accumulation during actual interaction tasks, encompassing specific observations, chains of thought, action trajectories, and environmental feedback. It is important to note that this section just provides a brief overview of experiential memory strictly from the perspective of token-level storage; a more comprehensive analysis and detailed discussion of this domain will be presented in \Cref{ssec:experiential_mem}.

The most fundamental form of experience memory involves the direct archival of historical behavioral trajectories. This paradigm enables agents to inform current decision-making by retrieving and reusing past instances, encompassing both successful and failed cases~\citep{DBLP:journals/corr/abs-2508-16153,DBLP:journals/pami/WangCLJHZLHZYML25}.

To address the limited generalizability inherent in raw trajectories, a significant body of research focuses on abstracting specific interactions into higher-level, generalized experiences. As one of the earliest and most influential approaches, Reflexion~\citep{shinn_reflexion_2023} distinguishes short-term memory as the trajectory history and long-term memory as the feedback produced by the self-reflection model. Certain studies compress complex interaction histories into universal workflows, rule templates, or high-level ``thought-templates'' to facilitate cross-problem transfer and reuse~\citep{wang2024agentworkflowmemory,DBLP:journals/corr/abs-2509-17459,yang2024buffer}. Other works emphasize the structural organization and dynamic maintenance of memory. These approaches ensure that stored insights remain adaptable to novel tasks and are efficiently updated by constructing domain-specific structured knowledge bases, employing hierarchical plan-execute memory architectures, or incorporating human-like forgetting and reflection mechanisms~\citep{tang2025chemagentselfupdatinglibrarylarge,tang2025agentkbleveragingcrossdomain,ouyang2025reasoningbankscalingagentselfevolving,DBLP:journals/corr/abs-2509-12810,DBLP:conf/aaai/Zhao0XLLH24,DBLP:journals/ijon/SAGE}.

In contexts involving programming or specific tool utilization, experience memory evolves into executable skills. Within this paradigm, agents consolidate exploration experiences into code repositories, procedural scripts, or tool-usage entries. Leveraging environmental feedback, these systems iteratively refine code quality or even dynamically modify their underlying logic to achieve self-evolution~\citep{DBLP:journals/tmlr/WangX0MXZFA24,DBLP:conf/acl/YinWPL0W25,fang2025mempexploringagentprocedural,xiao_toolmem_2025}. Furthermore, targeting complex environments such as operating systems, some studies distill successful execution records into reusable exemplars or vectorized representations, thereby facilitating an efficient pipeline from offline construction to online allocation~\citep{DBLP:journals/corr/abs-2504-14603,han_legomem_2025}.

\paragraph{Multimodal}
Multimodal memory systems store information in the form of discrete token-level units extracted from raw multimodal data, such as images, video frames, audio segments, and text, enabling agents to capture, compress, and retrieve knowledge across channels and over long spans of experience. In wearable and egocentric settings, early work such as Ego-LLaVA~\citep{shenEncodeStoreRetrieveAugmentingHuman2024} captures first-person video and converts it into lightweight language descriptions. Memoro~\citep{DBLP:conf/chi/ZulfikarCM24} follows a similar philosophy but uses speech-to-text to form embedding-based memory chunks. Building on this direction, Livia~\citep{DBLP:journals/corr/abs-2509-05298} incorporates long-term user memory into an AR system with emotional awareness, applying forgetting curves and pruning strategies. 

For video understanding, the emphasis shifts toward separating transient visual cues from enduring contextual information. MovieChat~\citep{DBLP:conf/cvpr/SongCWZZWCG0ZLH24} adopts a short-term/long-term split, storing recent frame features. MA-LMM~\citep{heMALMMMemoryAugmentedLarge2024} pushes this further with a dual-bank design—one storing raw visual features and the other retaining query embeddings. VideoAgent~\citep{wangVideoAgentLongFormVideo2024} adopts a more semantically organized approach, maintaining a temporal memory of textual clip descriptions alongside object-level memory that tracks entities across frames. In interactive video generation, Context-as-Memory~\citep{DBLP:journals/corr/abs-2506-03141} shows that simply storing previously generated frames as memory can also be highly effective. WorldMM~\citep{yeo2025worldmmdynamicmultimodalmemory} constructs multiple mutually reinforcing memory modules that capture information in both textual and visual modalities.

In embodied scenarios, memory becomes inherently tied to spatial structure and ongoing interaction. KARMA~\citep{wangKARMAAugmentingEmbodied2025} introduces a two-tier memory system: long-term memory stores static objects in a 3D scene graph, while short-term memory tracks object positions and state changes. Embodied VideoAgent~\citep{DBLP:journals/corr/abs-2501-00358} also builds persistent object memories but fuses them with first-person video and additional embodied sensors. Mem2Ego~\citep{DBLP:journals/corr/abs-2502-14254} extends this idea to navigation by separating global maps, landmark descriptions, and visitation histories into three distinct memory stores. Complementing these task-driven designs, MEMENTO~\citep{DBLP:journals/corr/abs-2505-16348} provides an evaluation framework that treats multimodal interaction history as an agent’s memory, enabling systematic assessment of how well embodied systems utilize accumulated perceptual experience.

\paragraph{Discussion}The primary advantage of Flat Memory is their simplicity and scalability: memory can be appended or pruned with minimal cost, and retrieval methods such as similarity search allow flexible access without requiring predefined structure. This makes them suitable for broad recall, episodic accumulation, and rapidly changing interaction histories. However, the lack of explicit relational organization means that coherence and relevance depend heavily on retrieval quality. As the memory grows, redundancy and noise can accumulate, and the model may retrieve relevant units without understanding how they relate, limiting compositional reasoning, long-horizon planning, and abstraction formation. Thus, topology-free collections excel at broad coverage and lightweight updates, but are constrained in tasks requiring structured inference or stable knowledge organization.

\subsubsection{Planar Memory (2D)}\label{sec:token-level-2d}

\vspace{0.6em}
\begin{tcolorbox}[
  colback=selfevolagent_light!20,
  colframe=selfevolagent_light!80,
  colbacktitle=selfevolagent_light!80,
  coltitle=black,
  title={\bfseries\fontfamily{ppl}\selectfont{Definition of Planar (2D) Memory}},
  boxrule=2pt,
  arc=5pt,
  drop shadow,
  parbox=false,
  before skip=5pt,
  after skip=5pt,
  left=5pt,   
  right=5pt,
]
Planar Memory introduces an explicit organizational topology among memory units, but only within a single structural layer, which for short called \textit{2D}. The topology may be a graph, tree, table, implicit connection structure and so on, where relationships such as adjacency, parent–child ordering, or semantic grouping are encoded within one plane, without hierarchical levels or cross-layer references.
\end{tcolorbox}

The core of Planar memory forms lies in breaking through a single storage pool by establishing explicit association mechanisms, achieving a leap from mere ``storage'' to ``organization''. 

\paragraph{Tree}
Tree structures organize information hierarchically and can handle different levels of abstraction. HAT~\citep{a2024enhancinglongtermmemoryusing} builds a Hierarchical Aggregate Tree by segmenting long interactions and then aggregating them step by step. This multi-level structure supports coarse-to-fine retrieval and performs better than flat vector indices in long-context question answering. To reduce dialogue fragmentation, MemTree~\citep{rezazadehIsolatedConversationsHierarchical2025} introduces a dynamic representation that infers hierarchical schemas from isolated conversation logs. It gradually summarizes concrete events into higher-level concepts, allowing agents to use both detailed memories and abstract knowledge.

\paragraph{Graph}
Graph structures dominate the landscape of 2D memory due to their ability to capture complex associations, causality, and temporal dynamics. Foundational works like Ret-LLM~\citep{modarressi2023ret} abstract external storage into addressable triple-based units, enabling the LLM to interact with a relation-centric table that functions like a lightweight knowledge graph. In the medical domain, HuaTuo~\citep{wang2023huatuotuningllamamodel} injects professional knowledge by integrating a structured corpus of Chinese medical knowledge graphs and clinical texts to fine-tune the base model. KGT~\citep{sun2024knowledgegraphtuningrealtime} introduces a real-time personalization mechanism where user preferences and feedback are encoded as nodes and edges in a user-specific knowledge graph. For reasoning-intensive tasks, PREMem~\citep{DBLP:journals/corr/PREMem} shifts part of the inference burden to the memory construction phase, deriving structured memory items and their evolution relations from raw dialogue. Similarly, Memory-augmented Query Reconstruction~\citep{xu2025memoryaugmentedqueryreconstructionllmbased} maintains a dedicated query memory that records past KG queries and reasoning steps, using retrieved records to reconstruct more accurate queries. Building on a timeline perspective, TeaFarm~\citep{ong2025lifelongdialogueagentstimelinebased} organizes dialogue history along segmented timelines and applies structured compression to manage lifelong context. COMET~\citep{kim2024commonsenseaugmentedmemoryconstructionmanagement} further refines conversational memory by using external commonsense bases to parse dialogue and dynamically update a context-aware persona graph with inferred hidden attributes. A-Mem~\citep{xuAMEMAgenticMemory2025} standardizes knowledge into card-like units. It organizes them by relevance and places related memories in the same box, which builds a complete memory network. Intrinsic Memory Agents~\citep{yuen2025intrinsicmemoryagentsheterogeneous} employ a partitioned architecture in which sub-agents maintain their own role-specific private memories while collaboratively reading and writing to a shared memory. Extending to multimodel agents, M3-Agent~\citep{long2025seeing} unifies image, audio, and text into an entity-centric memory graph. SALI~\citep{pan2024planningimaginationepisodicsimulation} constructs a Reality–Imagination Hybrid Memory, unifying real observations and imagined future scenarios into a consistent navigation graph.

\paragraph{Hybrid}
Complex tasks often require hybrid architectures that segregate distinct cognitive functions while sharing a common memory substrate. Optimus-1~\citep{li2024optimus1hybridmultimodalmemory} explicitly separates static knowledge into a hierarchical directed knowledge graph for planning, and dynamic interactions into an abstract multimodal experience Pool for reflection and self-improvement. D-SMART~\citep{Lei2025DSMART} combines a structured factual memory, implemented as a continuously updated knowledge graph, with a traversal-based reasoning tree.

\paragraph{Discussion}
The Planar Memory, by effectively establishing links between its nodes, enables memories to leverage collective synergies and thus encode more comprehensive contextual knowledge. Moreover, it supports retrieval mechanisms that go beyond simple iteration, including structured key–value lookups and relational traversal along graph edges. These capabilities make the form strong in storing, organizing, and managing memories. However, it also faces a critical limitation: Without a hierarchical storage mechanism, all memories must be consolidated into a single, monolithic module. As task scenarios grow in complexity and diversity, this redundant and flattened design becomes increasingly inadequate for robust performance. More importantly, the high construction and search costs significantly hinder its practical deployment.

\subsubsection{Hierarchical Memory (3D)}

\vspace{0.6em}
\begin{tcolorbox}[
  colback=selfevolagent_light!20,
  colframe=selfevolagent_light!80,
  colbacktitle=selfevolagent_light!80,
  coltitle=black,
  title={\bfseries\fontfamily{ppl}\selectfont{Definition of Hierarchical (3D) Memory}},
  boxrule=2pt,
  arc=5pt,
  drop shadow,
  parbox=false,
  before skip=5pt,
  after skip=5pt,
  left=5pt,   
  right=5pt,
]
Hierarchical memory organizes information across layers, using inter-level connections to shape the memories into a volumetric structured space.
\end{tcolorbox}

 Such hierarchies support representations at different degrees of abstraction—from raw observations, to compact event summaries, to higher-level thematic patterns. Cross-layer connections further yield a volumetric memory space through which the system can navigate not only laterally among units but also vertically across abstraction levels.

Hierarchical Memory moves beyond simple stratification, aiming to build complex systems with deep abstraction capabilities and dynamic evolutionary mechanisms. These works typically employ multi-level graph structures or neuroscience-inspired mechanisms to build a more human-like volumetric memory space, where information is richer and the connections between memory units are clearer and more explicit.

\paragraph{Pyramid}
This category constructs memory as multi-level pyramids, where information is progressively organized into higher layers of abstraction and queried in a coarse-to-fine manner. HiAgent~\citep{huHiAgentHierarchicalWorking2025} manages long-horizon tasks through a subgoal-centered hierarchical working memory, keeping detailed trajectories for the currently active subgoal while compressing completed subgoals into higher-level summaries that can be selectively retrieved when needed. GraphRAG~\citep{edge2025localglobalgraphrag} builds a multi-level graph index via community detection, recursively aggregating entity-level subgraphs into community-level summaries. Extending the idea of clustering memory nodes, Zep~\citep{Rasmussen2025Zep} formalizes agent memory as a Temporal Knowledge Graph, and it similarly performs community partitioning. ILM-TR~\citep{tang2024enhancinglongcontextperformance} employs a tree-structured, pyramidal index coupled with an Inner Loop mechanism, repeatedly querying summaries at different abstraction levels and updating a short-term memory buffer until the retrieved evidence and generated answer stabilize. To ensure controllable personalization, EMG-RAG~\citep{wang2024craftingpersonalizedagentsretrievalaugmented} organizes an Editable Memory Graph into three tiers, where a tree-like type and subclass index (L1, L2) sits above an entity-level memory graph (L3). In multi-agent systems, G-Memory~\citep{Zhang2025GMemory} structures shared experience using a three-tier graph hierarchy of insight, query, and interaction graphs. This design enables query-centric traversal to move vertically between high-level cross-trial insights and compact trajectories of concrete collaborations.

\paragraph{Multi-Layer}
These forms instead emphasize layered specialization, organizing memory into distinct modules or levels that focus on particular information types or functions. Lyfe Agents~\citep{kaiya2023lyfeagentsgenerativeagents} separates salient long-term records from low-value transient details, allowing the system to maintain a compact, behaviorally important layer of memories. H-Mem~\citep{Sun2025HMEM} explicitly arranges long-term dialogue memory into a multi-level hierarchy ordered by semantic abstraction, where lower layers store fine-grained interaction snippets and higher layers store increasingly compressed summaries. Biologically inspired architectures such as HippoRAG~\citep{gutiérrez2025hipporagneurobiologicallyinspiredlongterm} factor memory into an associative indexing component, implemented as an open knowledge graph, and an underlying passage store, using the graph layer to orchestrate multi-hop retrieval over stored content. Its successor, HippoRAG 2~\citep{gutiérrez2025ragmemorynonparametriccontinual}, extends this design into a non-parametric continual-learning setting, enriching the indexing layer with deeper passage integration and online LLM filtering. AriGraph~\citep{anokhin2024arigraph} separates memory by information type within a unified graph, combining a semantic knowledge-graph world model that encodes environment structure with an event-level component that links concrete observations back to the semantic backbone. Similarly, SGMem~\citep{Wu2025SGMemSG} adds a sentence-graph memory level on top of raw dialogue, representing histories as sentence-level graphs within chunked units. CAM~\citep{liCAMConstructivistView2025} layers the reading process itself by incrementally clustering overlapping semantic graphs into a hierarchical schemata structure. Recently, methods such as CompassMem~\citep{hu2026memorymattersmoreeventcentric} and MAGMA~\citep{jiang2026magmamultigraphbasedagentic} have begun exploring hierarchical composition strategies enriched with logical relations, aiming to make memory retrieval and utilization more efficient and comprehensive, so that memory can provide models with benefits beyond mere semantic information.

\paragraph{Discussion}
By placing memory nodes at the intersection of hierarchical and relational dimensions, Hierarchical Memory allows different memories to interact and form multi-dimensional synergies. This design helps the system encode knowledge that is more holistic and more deeply contextualized. The form also supports powerful retrieval: it enables complex, multi-path queries that move through relational networks within each layer and across abstraction levels between layers. This ability allows the system to retrieve task-relevant memories with high precision, leading to strong task performance.
However, the structure’s complexity and its dense information organization create challenges for both retrieval efficiency and overall effectiveness. In particular, ensuring that all stored memories remain semantically meaningful and designing the optimal three-dimensional layout of the system remain difficult and critical problems.

\subsection{Parametric Memory}\label{ssec:parametric}

{In contrast to token-level memory, which stores information as visible and editable discrete units, parametric memory stores information directly in the model’s parameters.} In this section, we examine methods that embed memory into learnable parameter spaces, allowing the model to internalize and recall information without referring to external storage.

Based on where the memory is stored relative to the core model parameters, we distinguish two primary forms of parametric memory:

\vspace{0.6em}
\begin{tcolorbox}[
  colback=selfevolagent_light!20,
  colframe=selfevolagent_light!80,
  colbacktitle=selfevolagent_light!80,
  coltitle=black,
  title={\bfseries\fontfamily{ppl}\selectfont{Two Major Types of Parametric Memory}},
  boxrule=2pt,
  arc=5pt,
  drop shadow,
  parbox=false,
  before skip=5pt,
  after skip=5pt,
  left=5pt,   
  right=5pt,
]
\begin{enumerate}
    \item \textbf{Internal Parametric Memory}: Memory encoded within the original parameters of the model (e.g., weights, biases). These methods directly adjust the base model to incorporate new knowledge or behavior.
    
    \item \textbf{External Parametric Memory}: Memory stored in additional or auxiliary parameter sets, such as adapters, LoRA modules, or lightweight proxy models. These methods introduce new parameters to carry memory without modifying the original model weights.
\end{enumerate}
\end{tcolorbox}

This distinction reflects a key design choice: whether memory is fully absorbed into the base model or attached modularly alongside it. In the subsections that follow, for each form we outline the implementation methods, analyze its strengths and limitations, and list representative systems or work. \autoref{tab:parametric-memory} provides an overview of representative parametric memory methods.

\begin{table*}[!t]

\caption{
Taxonomy of parametric memory methods.
We categorize existing works based on the \textit{storage location} relative to the core model: \textbf{Internal Parametric Memory} embeds knowledge directly into the original weights, while \textbf{External Parametric Memory} isolates information within auxiliary parameter sets.
Based on the training \textbf{phase}, we performed a secondary classification of the articles.
Methods are compared across three technical dimensions:
(1) \textbf{Type} defines the nature of the memory,
(2) \textbf{Task} specifies the target downstream application,
and (3) \textbf{Optimization} denotes the optimization strategy, such as \textit{SFT}, \textit{FT} (fine-tuning) , and \textit{PE} (prompt engineering).
}
\label{tab:parametric-memory}

\resizebox{\textwidth}{!}{
\begin{tabular}{p{6cm}|lp{6cm}l}
\toprule
{Method} & {Type} & {Task} & {Optimization}  \\
\midrule
\multicolumn{4}{c}{\textit{I. Internal Parametric Memory}}\\
\midrule
\multicolumn{4}{l}{\textbf{(a) Pre-Train Phase}}\\
TNL~\citep{lightening-attention} & Working & QA, Reasoning & SFT \\
StreamingLLM~\citep{attention-sink} & Working & QA, Reasoning & SFT \\
LMLM~\citep{zhao2025pretraininglimitedmemorylanguage} & Factual & QA, Factual Gen & SFT\\
HierMemLM~\citep{pt-hier-memory} & Factual & QA, Language Modeling & SFT\\
Function Token~\citep{zhang2025memoryretrievalconsolidationlarge} & Factual & Language Modeling & Pretrain \\

\midrule
\multicolumn{4}{l}{\textbf{(b) Mid-Train Phase}}\\
Agent-Founder~\citep{agent-founder} & Experiential & Tool Calling, Deep Research & SFT \\
Early Experience~\citep{agent-early-experience} & Experiential & Tool Calling, Embodied Simulation, Reasoning, Web  & SFT \\
\midrule
\multicolumn{4}{l}{\textbf{(c) Post-Train Phase}}\\
Character-LM~\citep{character-lm} & Factual & Role Playing & SFT  \\
CharacterGLM~\citep{characterGLM} & Factual & Role Playing & SFT \\
SELF-PARAM~\citep{self-param}  & Factual & QA, Recommendation & KL Tuning \\
Room~\citep{machine-memory-system}  & Experiential & Embodied Task & RL\\
KnowledgeEditor~\citep{edit-fact} & Factual & QA, Fact Checking & FT \\ 
Mend~\citep{mend}& Factual & QA, Fact Checking, Model Editing & FT \\ 
PersonalityEdit~\cite{personality-edit}  & Factual & QA, Model Editing & FT, PE \\ 
APP~\citep{app}  & Factual & QA & FT \\ 
DINM~\citep{dinm} & Experiential & QA, Detoxification & FT \\ 
AlphaEdit~\citep{fang2025alphaeditnullspaceconstrainedknowledge} & Factual & QA & FT\\
\midrule
\multicolumn{4}{c}{\textit{II. External Parametric Memory}} \\
\midrule
\multicolumn{4}{l}{\textbf{(a) Adapter-based Modules}}\\
MLP-Memory~\citep{mlp-memory}  & Factual & QA, Classification, Textual Entailment & SFT \\
K-Adapter~\citep{DBLP:conf/acl/WangTDWHJCJZ21} & Factual & QA, Entity Typing,  Classification & SFT\\
WISE~\citep{DBLP:conf/nips/0104L0XY0X0C24}  & Factual & QA, Hallucination Detection & SFT\\
ELDER~\citep{DBLP:conf/aaai/00010W0025} & Factual & Model Editing & SFT\\
T-Patcher~\citep{huang2023transformerpatchermistakeworthneuron} & Factual & QA & FT\\
Sparse Memory FT~\citep{lin2025continuallearningsparsememory} & Factual & QA & SFT\\
Memory Decoder~\citep{cao2025memorydecoder} & Factual & QA, Language Modeling & SFT\\
MemLoRA~\citep{bini2025memloradistillingexpertadapters} & Factual & QA & SFT\\

\midrule
\multicolumn{4}{l}{\textbf{(b) Auxiliary LM-based Modules}}\\
MAC~\citep{amortize-context}  & Factual & QA & SFT  \\
Retroformer~\citep{yao2024retroformer}  & Experiential & QA, Web Navigation & RL\\
\bottomrule
\end{tabular}
}
\end{table*}

\subsubsection{Internal Parametric Memory}\label{ssec:internal}

{Internal parameter memory injects domain knowledge, personalized knowledge, or priors required by downstream tasks into the model.} We also regard enhancing the model's long-context capability as injecting a prior. The timing of memory injection can be the pre-training phase, continued pre-training phase, mid-training phase, or post-training phase. The memory stored in internal parameters does not add extra parameters or additional modules.

\paragraph{Pre-Train} 
Some works introduce memory mechanisms during the pre-training phase, aiming to address the issue that long-tail world knowledge is difficult to compress into the limited model parameters.
LMLM~\citep{zhao2025pretraininglimitedmemorylanguage} and HierMemLM~\citep{pt-hier-memory} store the memory for knowledge retrieval in the model during the pre-training phase, while storing the knowledge itself in an external knowledge base. 
Some works also optimize the computational efficiency of attention to enhance long-window memory capability~\citep{attention-sink,lightening-attention,lightening-attention-2,flash-attention-2,flash-attention-3}.

\paragraph{Mid-Train}
During the continued pre-training phase, some works incorporate generalizable experience from downstream tasks. For instance, ~\citet{agent-founder} and ~\citet{agent-early-experience} integrate agent experience. Some works improve the long-window performance or efficiency of LLMs during the mid-training phase, enabling the model to maintain more short-term memory with longer windows in memory-aided tasks~\citep{big-bird,longformer}.

\paragraph{Post-Train} Other works incorporate memory during the post-training phase to adapt to downstream tasks. Some works enable LLMs to memorize personalized user history or styles. Some works allow LLMs to learn from the successes or failures of past similar task executions. Character-LM~\citep{character-lm} and CharacterGLM~\citep{characterGLM} fine-tunes the LLM into different characteristics. During the post-training phase, SELF-PARAM~\citep{self-param} injects additional knowledge through KL divergence distillation without requiring extra parameters. Room~\citep{machine-memory-system} stores knowledge externally while save experience internally. KnowledgeEditor~\citep{edit-fact} modifies internal parameters, aiming to alter only the knowledge that requires editing. MEND~\citep{mend} achieves fast knowledge editing by using small networks to modify the gradients of large models. PersonalityEdit~\citep{personality-edit} proposes an LLM personality editing dataset based on personality theories in psychology. APP~\citep{app} employs multiple training objectives to ensure that adjacent knowledge is minimally disturbed during knowledge editing. DINM~\citep{dinm} proposes a model editing method that enables the model to learn to reject such dangerous requests without affecting its normal functions.

\paragraph{Discussion} 
The advantages of internal parameters lie in their simple structure, which does not add extra inference overhead or deployment costs to the vanilla model. Their drawback is the difficulty in updating internal parameters: storing new memory requires retraining, which is costly and prone to forgetting old memory. Therefore, internal parameter memory is more suitable for large-scale storage of domain knowledge or task priors, rather than short segments of personalized memory or working memory.

\subsubsection{External Parametric Memory}\label{ssec:exteranl}

Storing memory as tokens outside LLMs leads to insufficient understanding of token-form memory content in the input window by the model. Meanwhile, storing memory in the parameters of LLMs has issues, such as difficulty in updating and conflicts with pre-trained knowledge. Some works adopt a compromise approach, which \textbf{introduces memory through external parameters }without altering the original parameters of LLMs.

\paragraph{Adapter} A common line of external parametric memory methods relies on modules that are attached to a frozen base model. MLP-Memory~\citep{mlp-memory} integrates RAG knowledge with Transformer decoders through MLP. K-Adapter~\citep{DBLP:conf/acl/WangTDWHJCJZ21} injects new knowledge by training task-specific adapter modules while keeping the original backbone unchanged, enabling continual knowledge expansion without interfering with pre-trained representations. WISE~\citep{DBLP:conf/nips/0104L0XY0X0C24} further introduces a dual-parameter memory setup—separating pre-trained knowledge and edited knowledge—and a routing mechanism that dynamically selects which parameter memory to use at inference time, thus mitigating conflicts during lifelong editing. ELDER~\citep{DBLP:conf/aaai/00010W0025} advances this direction by maintaining multiple LoRA modules and learning a routing function that adaptively selects or blends them based on input semantics, improving robustness and scalability in long-term editing scenarios. Collectively, these methods leverage additional parameter subspaces to store and retrieve memory in a modular and reversible manner, avoiding the risks of catastrophic interference associated with directly modifying the core model weights.

\paragraph{Auxiliary LM} Beyond Adapter-based storage, another line of work adopts a more architecturally decoupled form of external parametric memory, where memory is stored in a separate model or external knowledge module. MAC~\citep{amortize-context} compresses the information from a new document into a compact modulation through an amortization network, and stores it in a memory bank. Retroformer~\citep{yao2024retroformer} proposes a learning paradigm for memorizing the experiences of successes or failures in past task executions.

\paragraph{Discussion}
This external parametric memory approach provides a balance between adaptability and model stability. Because memory is encoded into additional parameter modules, it can be added, removed, or replaced without interfering with the base model’s pre-trained representation space. This supports modular updates, task-specific personalization, and controlled rollback, while avoiding the catastrophic forgetting or global weight distortion that may occur in full model fine-tuning.

However, this approach also comes with limitations. External parameter modules must still integrate with the model’s internal representation flow, meaning that their influence is indirect and mediated through the model’s attention and computation pathways. As a result, the effectiveness of memory injection depends on how well the external parameters can interface with internal parametric knowledge.

\begin{figure}[!t]
    \centering
    \includegraphics[width=0.9\linewidth]{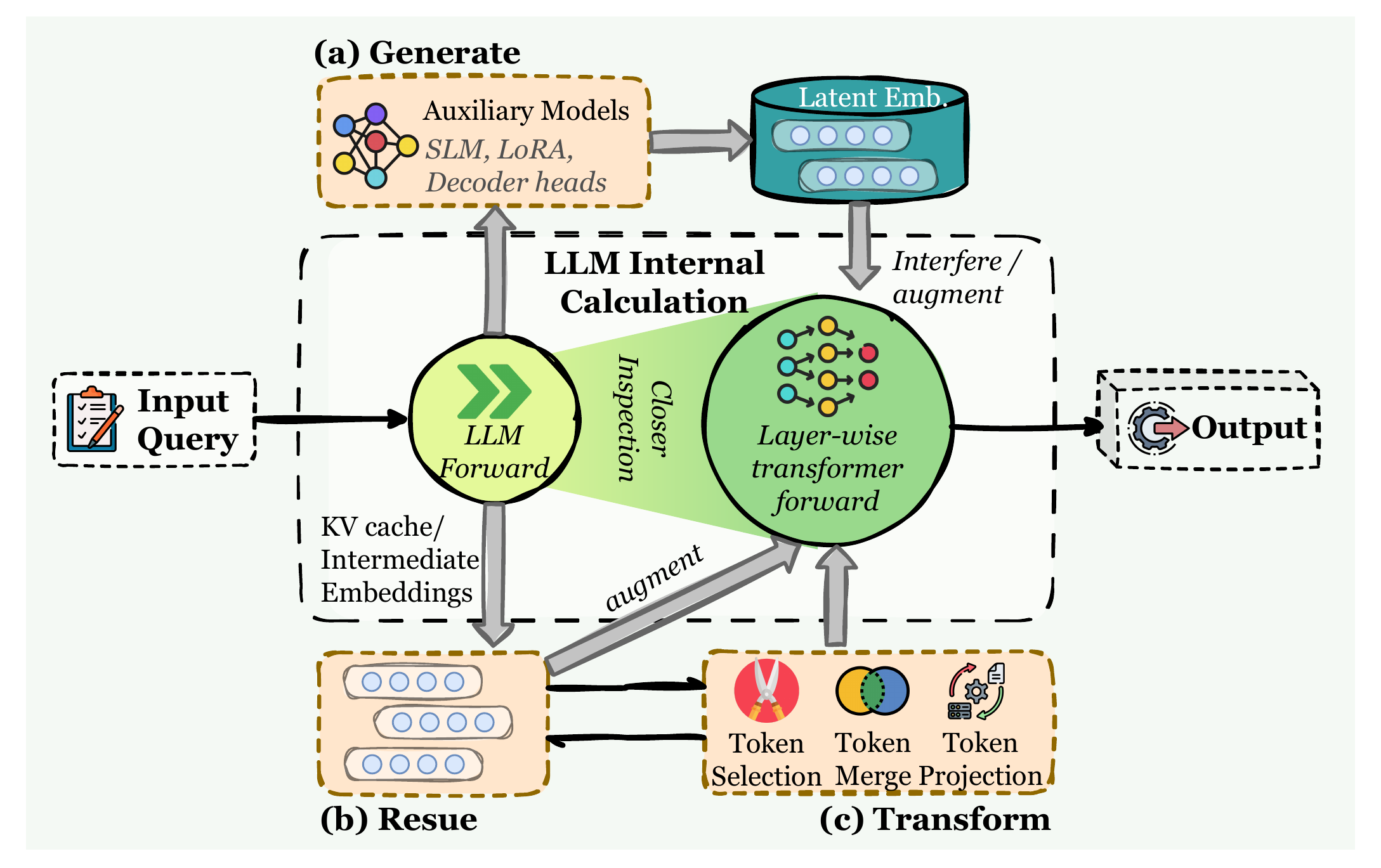}
    \caption{
    Overview of Latent Memory integration in LLM agents. Unlike explicit text storage, latent memory operates within the model's internal representational space. The framework is categorized by the origin of the latent state: (a) \textbf{Generate}, where auxiliary models synthesize embeddings to interfere with or augment the LLM's forward pass; (b) \textbf{Reuse}, which directly propagates prior computational states such as KV caches or intermediate embeddings; and (c) \textbf{Transform}, which compresses internal states through token selection, merging, or projection to maintain efficient context.
    }
    \label{fig:sec3-3.3}
\end{figure}

\subsection{Latent Memory}\label{ssec:latent}

\vspace{0.6em}
\begin{tcolorbox}[
  colback=selfevolagent_light!20,
  colframe=selfevolagent_light!80,
  colbacktitle=selfevolagent_light!80,
  coltitle=black,
  title={\bfseries\fontfamily{ppl}\selectfont{Definition of Latent Memory}},
  boxrule=2pt,
  arc=5pt,
  drop shadow,
  parbox=false,
  before skip=5pt,
  after skip=5pt,
  left=5pt,   
  right=5pt,
]
Latent memory refers to memory that is carried implicitly in the model’s internal representations (e.g., KV cache, activations, hidden states, latent embeddings), rather than being stored as explicit, human-readable tokens or dedicated parameter sets. 
\end{tcolorbox}

Latent avoids exposing memory in plaintext and introduces practically less inference latency, while potentially offering better performance gains by preserving fine-grained contextual signals within the model’s own representational space.

As shown in \autoref{fig:sec3-3.3}, we organize prior work by the origin of latent memory, which means how the latent state is formed and introduced into the agent. We summarize the works in this part in \autoref{tab:latent-memory}.

\vspace{0.6em}
\begin{tcolorbox}[
  colback=selfevolagent_light!20,
  colframe=selfevolagent_light!80,
  colbacktitle=selfevolagent_light!80,
  coltitle=black,
  title={\bfseries\fontfamily{ppl}\selectfont{Three Major Types of Latent Memory}},
  boxrule=2pt,
  arc=5pt,
  drop shadow,
  parbox=false,
  before skip=5pt,
  after skip=5pt,
  left=5pt,   
  right=5pt,
]
\begin{enumerate}
    \item \textbf{Generate}: latent memory is produced by an independent model or a module, and then supplied to the agent as reusable internal representations.
    
    \item \textbf{Reuse}: latent memory is directly carried over from prior computation, most prominently KV-cache reuse (within or across turns), as well as recurrent or stateful controllers that propagate hidden states.

    \item \textbf{Transform}: existing latent state is transformed into new representations(e.g., distillation, pooling, or compression), so the agent can retain essentials while reducing latency and context footprint.
\end{enumerate}
\end{tcolorbox}

\begin{table*}[!t]
\centering
\caption{
Taxonomy of latent memory methods.
We categorize existing works based on the \textit{origin} of the latent state: 
\textbf{Generate} synthesizes memory via auxiliary modules, 
\textbf{Reuse} propagates internal computational states, 
and \textbf{Transform} compresses, modifies or restructs existing latent state.
Methods are compared across three technical dimensions:
(1) \textbf{Form} specifies the specific data type of the latent memory,
(2) \textbf{Type} defines the nature of the recorded content (e.g., Working, Factual, and Experiential),
and (3) \textbf{Task} denotes the target downstream application.
}
\label{tab:latent-memory}

\renewcommand{\arraystretch}{0.97} 

\resizebox{\textwidth}{!}{
\begin{tabular}{l|lll}
\toprule
{Method} & {Form} & {Type} & {Task} \\
\midrule
\multicolumn{4}{c}{\textit{I. Generate}}\\
\midrule
\multicolumn{4}{l}{\textbf{(a) Single Modal}}\\
Gist~\citep{DBLP:conf/nips/Mu0G23} & Gist Tokens & Working & Long-context Compression \\
Taking a Deep Breath~\citep{DBLP:conf/emnlp/LuoZXWLLCS24} & Sentinel Tokens & Working & Long-context QA \\
SoftCoT~\citep{DBLP:conf/acl/00010ZM25} & Soft Tokens & Working & Reasoning \\
CARE~\citep{DBLP:journals/corr/abs-2508-15253} & Memory Tokens & Working & QA, Fact Checking \\
AutoCompressor~\citep{DBLP:conf/emnlp/ChevalierWAC23} & Summary Vectors &Working & QA, Compression \\
MemoRAG~\citep{DBLP:conf/www/Qian0ZMLD025} & Global Semantic States & Working & QA, Summary \\
MemoryLLM~\citep{Wang2024MEMORYLLM} & Persistent Tokens & Factual & Long-conv QA, Model Editing \\
M+~\citep{DBLP:journals/corr/abs-2502-00592} & Cross-layer Token Pools & Factual & QA \\
LM2~\citep{DBLP:journals/corr/abs-2502-06049} & Matrix Slots & Working & QA, Reasoning \\
Titans~\citep{DBLP:journals/corr/abs-2501-00663} & Neural Weights (MLP) & Working & QA, Language Modeling\\
MemGen~\citep{Zhang2025MemGen} & LoRA Fragments & Working, Exp. & QA, Math, Code, Embodied Task, Reasoning \\
EMU~\citep{DBLP:conf/iclr/NaSM24} & Embeddings w/ Returns & Factual & Game \\
TokMem~\citep{wu2025tokmemtokenizedproceduralmemory} & Memory Tokens & Exp. & Funcation calling \\
Nested Learning~\citep{behrouz2025nested} & Nested Optimization & Factual & Language Modeling \\
Memoria~\citep{park2024memoria} & Three memory layers with engrams & Factual & Language Modeling \\

\midrule
\multicolumn{4}{l}{\textbf{(b) Multi-Modal}}\\
CoMem~\citep{Wu2025CoMEM} & Multimodal Embeddings & Factual & Multimodal QA \\
ACM~\citep{DBLP:journals/corr/abs-2510-09038} & Trajectory Embeddings & Working & Web \\
Time-VLM~\citep{DBLP:journals/corr/abs-2502-04395} & Patch Embeddings & Working & Video Understanding \\
Mem Augmented RL  ~\citep{DBLP:conf/iros/MezghaniSLMBBA22} & Novelty State Encoder & Working & Visual Navigation \\
MemoryVLA~\citep{DBLP:journals/corr/abs-2508-19236} & Perceptual States & Factual, Working & Embodied Task \\
XMem~\citep{cheng2022xmemlongtermvideoobject} & Key-Value Embeddings & Working & Video Segmentation \\

\midrule
\multicolumn{4}{c}{\textit{II. Reuse}} \\
\midrule
Memorizing Transformers~\citep{DBLP:conf/iclr/WuRHS22} & External KV Cache & Working & Language Modeling \\
SirLLM~\citep{DBLP:conf/acl/YaoL024} & Entropy-selected KV & Factual & Long-conv QA \\
Memory${}^{\mbox{3}}$~\citep{DBLP:journals/corr/abs-2407-01178} & Critical KV Pairs & Factual & QA \\
FOT~\citep{DBLP:conf/nips/TworkowskiSPWMM23} & Memory-Attention KV & Working & QA, Few-shot learning, Language Modeling \\
LONGMEM~\citep{DBLP:conf/nips/Wang0CLYGW23} & Residual SideNet KV & Working & Language Modeling and Understanding \\

\midrule
\multicolumn{4}{c}{\textit{III. Transform}} \\
\midrule
Scissorhands~\citep{DBLP:conf/nips/LiuDLWXXKS23} & Pruned KV & Working & Image classification \& generation \\
SnapKV~\citep{DBLP:conf/nips/LiHYVLYCLC24} & Aggregated Prefix KV & Working & Language Modeling \\
PyramidKV~\citep{DBLP:journals/corr/abs-2406-02069} & Layer-wise Budget & Working & Language Modeling \\
RazorAttention~\citep{DBLP:conf/iclr/TangLLHKHYW25} & Compensated Window & Working & Language Modeling \\
H2O~\citep{DBLP:conf/nips/Zhang00CZC0TRBW23} & Heavy Hitter Tokens & Working & QA, Language Modeling  \\
R${}^3$Mem~\citep{wang2025R3Mem} & Virtual memory tokens with reversible compression & Working & QA, Language Modeling  \\

\bottomrule
\end{tabular}
}
\end{table*}

\subsubsection{Generate}
\label{ssec:generate}

A major line of work builds memory by \textbf{generating new latent representations} rather than reusing or transforming existing activations. In this paradigm, the model or an auxiliary encoder creates compact continuous states. These states may appear as special tokens in the sequence or as standalone vectors. They summarize the essential information from long contexts, task trajectories, or multimodal inputs. The generated latent summaries are then stored, inserted, or used as conditions for later reasoning or decision-making. This enables the system to operate beyond its native context length, maintain task-specific intermediate states, and retain knowledge across episodes without revisiting the original input. Although the concrete forms vary across studies, the underlying idea remains consistent. Memory is explicitly produced through learned encoding or compression, and the resulting latent states serve as reusable memory units that support future inference.

This design choice may also raise potential ambiguity with parametric memory, particularly since many methods rely on separately trained models to generate latent representations. In this chapter, however, our classification is grounded in the form of memory rather than the learning mechanism. Crucially, although these approaches generate memory through learned encoding, the produced latent representations are explicitly instantiated and reused as independent memory units, rather than being directly embedded into the model’s parameters or forward-pass activations. We will return to this distinction when discussing individual methods in detail.

\paragraph{Single Modal} In the single-modal setting, a major group of methods focuses on long-context processing and language modeling, where models generate a small set of internal representations to replace long raw inputs ~\citep{DBLP:conf/nips/Mu0G23,DBLP:conf/emnlp/LuoZXWLLCS24,DBLP:conf/acl/00010ZM25,DBLP:conf/emnlp/ChevalierWAC23,DBLP:conf/www/Qian0ZMLD025,Wang2024MEMORYLLM,DBLP:journals/corr/abs-2502-00592}. A typical strategy is to compress long sequences into a few internal tokens or continuous vectors that can be reused during later inference. For example, Gist~\citep{DBLP:conf/nips/Mu0G23} train a language model to produce a set of gist tokens after processing a long prompt. \citet{DBLP:conf/emnlp/LuoZXWLLCS24} introduce a special sentinel token at each chunk boundary and encourage the model to aggregate local semantics into that token. SoftCoT ~\citep{DBLP:conf/acl/00010ZM25} follows a similar direction by generating instance-specific soft tokens from the last hidden state. CARE ~\citep{DBLP:journals/corr/abs-2508-15253} further extends the latent tokens by training a context assessor that compresses retrieved RAG documents into compact memory tokens.

Work such as AutoCompressor~\citep{DBLP:conf/emnlp/ChevalierWAC23} and MemoRAG~\citep{DBLP:conf/www/Qian0ZMLD025} emphasizes vectorized or standalone latent representations. AutoCompressor~\citep{DBLP:conf/emnlp/ChevalierWAC23} encodes entire long documents into a small number of summary vectors serving as soft prompts, while MemoRAG~\citep{DBLP:conf/www/Qian0ZMLD025} uses an LLM to produce compact hidden-state memories capturing global semantic structure. These approaches not only abstract away from raw text but also transform retrieved or contextualized information into new latent memory units optimized for reuse. To support more persistent memory, MemoryLLM ~\citep{Wang2024MEMORYLLM} embeds a set of dedicated memory tokens within the model’s latent space. M+ ~\citep{DBLP:journals/corr/abs-2502-00592} extends this idea into a cross-layer long-term memory architecture. LM2~\citep{DBLP:journals/corr/abs-2502-06049} follows a related but structurally distinct direction by introducing matrix-shaped latent memory slots into every layer.

A different branch of work internalizes the generation of latent memory within the model’s parameter dynamics. Although these works rely on parameterized modules, their operational memory units remain latent representations, placing them firmly within this category. Titans ~\citep{DBLP:journals/corr/abs-2501-00663} compresses long-range information into an online-updated MLP weight, producing latent vectors during inference. MemGen ~\citep{Zhang2025MemGen} dynamically generates latent memory during decoding: two LoRA adapters determine where to insert memory fragments and what latent content to insert. EMU~\citep{DBLP:conf/iclr/NaSM24} trains a state encoder to produce latent embeddings annotated with returns and desirability.

\paragraph{Multi Modal} In multimodal settings, generative latent memory extends to images, audios and videos, encoding them as compact latent representations. CoMem~\citep{Wu2025CoMEM} uses a VLM to compress multimodal knowledge into a set of embeddings that act as plug-and-play memory. Similarly, \citet{DBLP:journals/corr/abs-2510-09038} compresses entire GUI interaction trajectories into fixed-length embeddings and injects them into the VLM input space. For temporal modeling, Time-VLM~\citep{DBLP:journals/corr/abs-2502-04395} divides video or interaction streams into patches and generates a latent embedding for each patch. 

In vision-based navigation, \citet{DBLP:conf/iros/MezghaniSLMBBA22} learns a state encoder that maps visual observations into a latent space and constructs an episodic memory containing only novel observations. MemoryVLA~\citep{DBLP:journals/corr/abs-2508-19236} maintains a Perceptual–Cognitive Memory Bank that stores both perceptual details and high-level semantics as transformer hidden states. In long-video object segmentation, XMem\citep{cheng2022xmemlongtermvideoobject} encodes each frame into key–value latent embeddings and organizes them into a multi-stage memory comprising perceptual, working, and long-term components.

\paragraph{Discussion}
These single-modal and multimodal approaches share the same fundamental principle: first generate compact latent representations, then maintain and retrieve them as memory entries. The model can actively construct highly information-dense representations tailored to the task, capturing key dynamics, long-range dependencies, or cross-modal relations with minimal storage cost. It also avoids repeatedly processing the full context, enabling more efficient reasoning across extended interactions.

However, the drawbacks are equally evident. The generation process itself may introduce information loss or bias, and the states can drift or accumulate errors over multiple read–write cycles. Moreover, training a dedicated module to generate latent representations introduces additional computational overhead, data requirements, and engineering complexity.

\subsubsection{Reuse}
\label{ssec:reuse}
In contrast to methods that generate new latent representations, another line of work directly \textbf{reuses the model’s internal activations, primarily the key–value (KV) cache}, as latent memory. These approaches do not transform(modify, compress) the stored KV pairs and instead treat the raw activations from forward passes as reusable memory entries. The main challenge is to determine which KV pairs to keep, how to index them, and how to retrieve them efficiently under long-context or continual-processing demands.

From a cognitive perspective, \citet{DBLP:journals/corr/abs-2501-02950} provides conceptual grounding by framing biological memory as a key–value system, where keys function as retrieval addresses and values encode stored content—an abstraction closely aligned with KV-based memory in modern LLMs. Memorizing Transformers ~\citep{DBLP:conf/iclr/WuRHS22} explicitly store past KV pairs and retrieve them via K-nearest-neighbor search during inference. 
FOT~\citep{DBLP:conf/nips/TworkowskiSPWMM23} extends this line of work by introducing memory-attention layers that perform KNN-based retrieval over additional KV memories during inference. LONGMEM ~\citep{DBLP:conf/nips/Wang0CLYGW23} similarly augments long-range retrieval, employing a lightweight residual SideNet that treats historical KV embeddings as a persistent memory store. These systems demonstrate how retrieval-aware organization of latent KV states can substantially enhance access to distant information.

\paragraph{Discussion}
Reuse-type latent memory methods highlight the effectiveness of directly leveraging the model’s own internal activations as memory, showing that carefully curated KV representations can serve as a powerful and efficient substrate for long-range retrieval and reasoning.

Their greatest strength lies in preserving the full fidelity of the model’s internal activations, ensuring that no information is lost through pruning or compression. This makes them conceptually simple, easy to integrate into existing forms, and highly faithful to the model’s original computation. However, raw KV caches grow rapidly with context length, which increases memory consumption and can make retrieval less efficient. The effectiveness of reuse therefore depends heavily on indexing strategies.

\subsubsection{Transform}
\label{ssec:transform}
{Transform-type latent memory methods focus on \textbf{modifying, compressing, or restructuring existing latent states} rather than generating entirely new ones or directly reusing raw KV caches.} These approaches treat KV caches and hidden activations as malleable memory units, reshaping them through selection, aggregation, or structural transformation. In doing so, they occupy a conceptual middle ground between generate-type and reuse-type memory: the model does not create fresh latent representations, but it also does more than simply replay stored KV pairs.

A major line of work focuses on compressing KV caches while preserving essential semantics. Some methods reduce memory usage by keeping only the most influential tokens. Scissorhands~\citep{DBLP:conf/nips/LiuDLWXXKS23} prunes tokens based on attention scores when cache capacity is exceeded, whereas SnapKV~\citep{DBLP:conf/nips/LiHYVLYCLC24} aggregates high-importance prefix KV representations via a head-wise voting mechanism. PyramidKV ~\citep{DBLP:journals/corr/abs-2406-02069} reallocates KV budgets across layers. SirLLM ~\citep{DBLP:conf/acl/YaoL024} builds on this perspective by estimating token importance with a token-entropy criterion and selectively retaining only informative KV entries. Memory${}^{\mbox{3}}$ ~\citep{DBLP:journals/corr/abs-2407-01178} only stores the most critical attention key–value pairs, significantly shrinking storage requirements. RazorAttention ~\citep{DBLP:conf/iclr/TangLLHKHYW25} introduces a more explicit compression scheme: it computes the effective attention span of each head, retains only a limited local window, and uses compensation tokens to preserve information from discarded entries. From a more efficiency-oriented perspective, H2O ~\citep{DBLP:conf/nips/Zhang00CZC0TRBW23} adopts a simpler eviction strategy, retaining only the most recent tokens along with special H2 tokens to reduce memory footprint.

\paragraph{Discussion}
These methods demonstrate how latent memory can be transformed, through selection, retrieval enhancement, or compressed re-encoding, into more effective memory representations, enabling LLMs to extend their usable context length and improve reasoning performance without relying on raw cache reuse.

Their main advantage lies in producing more compact and information-dense memory representations, which reduce storage cost and enable efficient retrieval over long contexts. By reshaping latent states, these methods allow the model to access distilled semantic signals that may be more useful than raw activations. However, transformation introduces the risk of information loss, and the compressed states can become harder to interpret or verify compared with directly reused KV caches. The additional computation required for pruning, aggregation, or re-encoding also increases system complexity.

\begin{figure}[!t]
    \centering
    \includegraphics[width=\linewidth]{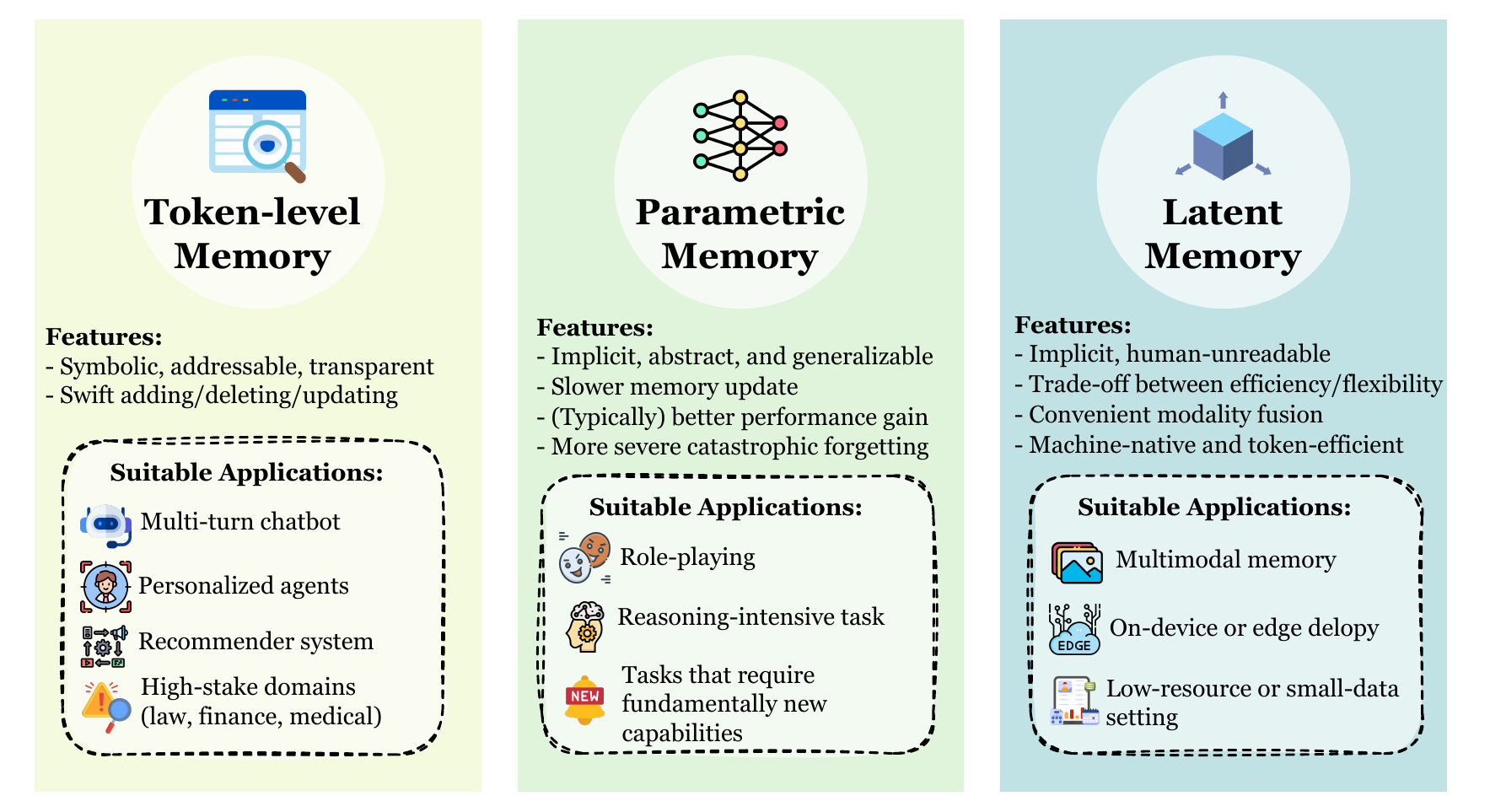}
    \caption{Overview of three complementary memory paradigms for LLM agents. Token-level, parametric, and latent memories differ in their representational form, update dynamics, interpretability, and efficiency, leading to distinct strengths, limitations, and application domains in long-horizon and interactive agent systems.}
    \label{fig:sec3-comparison}
\end{figure}

\subsection{Adaptation}
As shown above, such a large body of work has focused on agent memory, clearly demonstrating that memory mechanisms are essential for agent systems~\citep{zhang2025ruc_survey}. The choice of memory type in an agent system reflects how designers expect the agent to behave in a given task. Designers are not simply asking the agent to remember certain information, but also implicitly expressing how they want that information to shape the agent’s behavior. Therefore, choosing the right type of memory for a task is far more than a simple combinatorial choice.

In this section, we start from the features of each memory type and discuss which tasks and scenarios they are best suited for in an ideal setting, as shown in \autoref{fig:sec3-comparison}. We hope this discussion can offer useful ideas and guidance for making practical choices. The examples illustrate only one possible form of memory in these idealized settings and do not imply that other memory types lack unique advantages in the same scenarios.

\paragraph{Token-level Memory}
Token-level memory remains \emph{symbolic}, \emph{addressable}, and \emph{transparent}, making it particularly well suited for scenarios where explicit reasoning, controllability, and accountability are essential. This type of memory excels in real-time, high-frequency update settings, where an agent must continuously track and revise information, and where the knowledge itself exhibits a clear structure that can be explicitly modeled. Its externalizability allows memory to be easily inspected, audited, transferred, or revised, making it especially suitable for domains requiring precise add/delete/update operations. The high level of interpretability further ensures that an agent's decision process can be traced back to concrete memory units, a crucial property in high-stakes applications. Moreover, token-level memory provides long-term stability and avoids catastrophic forgetting, enabling agents to accumulate reliable knowledge over extended time horizons. Another practical advantage is that token-level memory is often implemented as a plug-and-play module, allowing it to be readily integrated with the latest closed-source or open-source foundation models without modifying their internal parameters.

\textbf{Possible Scenarios:}
\begin{itemize}[itemsep=-0.2em]
    \item Chatbots and multi-turn dialogue systems.~\citep{zhong2023memorybankenhancinglargelanguage,lu2023memochattuningllmsuse,Chhikara2025mem0}
    \item Long-horizon or life-long agents requiring stable memory.~\citep{wang2024aipersona, DBLP:journals/corr/abs-2510-07925}
    \item User-specific personalization profiles.~\citep{wang2024aipersona,DBLP:conf/acl/LeeHPP023}
    \item Recommendation systems.~\citep{wang_recmind_2024, DBLP:journals/tois/HuangLLYLX25, DBLP:conf/cikm/XiLL0T0024}
    \item Enterprise or organizational knowledge bases.
    \item Legal, compliance, and other high-stakes domains requiring verifiable provenance.
\end{itemize}

\paragraph{Parametric Memory}
Compared with symbolic memory, parametric memory is \emph{implicit, abstract}, and \emph{generalizable}, making it naturally suited to tasks requiring conceptual understanding and broad pattern induction. It is particularly effective when the agent must rely on general knowledge or rules that apply across diverse contexts, because such regularities can be internalized as distributed representations without requiring explicit external lookup. This internalization supports fluid reasoning and end-to-end processing, enabling the model to generalize systematically to unseen tasks or problem variations. Consequently, parametric memory is better aligned with tasks demanding structural insight, robust abstraction, and deeply ingrained behavioral or stylistic patterns.

\textbf{Possible Scenarios:}
\begin{itemize}[itemsep=-0.2em]
    \item Role-playing or persona-consistent behaviors.~\citep{character-lm,characterGLM}
    \item Mathematical reasoning, coding, games, and structured problem-solving.
    \item Human alignment and normative behavioral priors.
    \item Stylized, professional, or domain-expert responses.
\end{itemize}

\paragraph{Latent Memory}
Unlike token-level or parametric memory, latent memory sits between explicit data and fixed model weights, enabling a unique balance of flexibility and efficiency. Its low readability provides intrinsic privacy protection, making latent representations suitable for sensitive information processing. At the same time, their high expressive capacity permits rich semantic encoding with minimal information loss, allowing agents to capture subtle correlations across modalities or tasks. Latent memory also supports efficient inference-time retrieval and integration, enabling agents to inject large quantities of compact knowledge. This memory type therefore prioritizes performance and scalability over interpretability, achieving high knowledge density and compression ideal for constrained or highly dynamic environments.

\textbf{Possible Scenarios:}
\begin{itemize}[itemsep=-0.2em]
    \item Multimodal or fully integrated agent architectures.~\citep{DBLP:journals/corr/abs-2508-19236,cheng2022xmemlongtermvideoobject,Wu2025CoMEM}
    \item On-device or edge deployment and cloud-serving environments.
    \item Encrypted or privacy-sensitive application domains.
    
\end{itemize}

\section{Functions: Why Agents Need Memory?}
\label{sec:why-memory}
The transition from large language models as general-purpose, stateless text processors to autonomous, goal-directed agents is not merely an incremental step but a fundamental paradigm shift. This shift exposes the critical limitation of statelessness. By definition, an agent must persist, adapt, and interact coherently over time. Achieving this relies not merely on a large context window but fundamentally on the capacity for \textbf{memory}. This section addresses the \textit{functions}, or \textit{fundamental purpose}, of agent memory, prioritizing the question of \textit{why it is essential} over \textit{how it is implemented}. We posit that agent memory is not a monolithic component but a set of distinct functional capabilities, each serving a unique objective in enabling persistent, intelligent behavior.

\begin{figure}[!t]
    \centering
    \includegraphics[width=\linewidth]{img/sec4-all-new.pdf}
    \caption{
    The functional taxonomy of agent memory. We organize memory capabilities based on their \textit{functions} (purpose) into three primary pillars spanning two temporal domains: (1) \textbf{Factual Memory} serves as a persistent declarative knowledge base to ensure interaction \textit{consistency}, \textit{coherence}, and \textit{adaptability}; (2) \textbf{Experiential Memory} encapsulates procedural knowledge to enable \textit{continual learning} and \textit{self-evolution} across episodes; and (3) \textbf{Working Memory} provides mechanisms for the active management of transient context.}
    \label{fig:sec4-all}
\end{figure}

To provide a systematic analysis, this section organizes the \textit{why} of memory around a functional taxonomy that maps directly to an agent's core requirements. At the highest level, we distinguish between two temporal categories: \textbf{long-term memory}, which serves as the persistent, cross-session store for accumulated knowledge, and \textbf{short-term memory}, which functions as the transient, in-session workspace for active reasoning. 
This high-level temporal split is further resolved into three primary functional pillars, which form the structure of our analysis. An overview of this taxonomy is provided in \autoref{fig:sec4-all}.

\vspace{0.6em}
\begin{tcolorbox}[
  colback=selfevolagent_light!20,
  colframe=selfevolagent_light!80,
  colbacktitle=selfevolagent_light!80,
  coltitle=black,
  title={\bfseries\fontfamily{ppl}\selectfont{Three Primary Memory Functions}},
  boxrule=2pt,
  arc=5pt,
  drop shadow,
  parbox=false,
  before skip=5pt,
  after skip=10pt,
  left=5pt,   
  right=5pt,
]
\begin{enumerate}
    \item \textbf{Factual Memory} (\Cref{ssec:factual_memory}): The agent's declarative knowledge base, established to ensure consistency, coherence, and adaptability by recalling explicit facts, user preferences, and environmental states. 
    This system answers the question: ``What does the agent know?''
    \item \textbf{Experiential Memory} (\Cref{ssec:experiential_mem}): The agent's procedural and strategic knowledge, accumulated to enable continual learning and self-evolution by abstracting from past trajectories, failures, and successes. 
    This system answers: ``How does the agent improve?''
    \item \textbf{Working Memory} (\Cref{ssec:working_mem}): The agent's capacity-limited, dynamically controlled scratchpad for active context management during a single task or session. This system answers: ``What is the agent thinking about now?''
\end{enumerate}

\end{tcolorbox}

These three memory systems are not isolated but form a dynamic, interconnected architecture that defines the agent's \textbf{cognitive loop}. The cycle begins with \textit{encoding}, in which the outcomes of the agent's interactions, such as newly acquired facts or the results of a failed plan, are consolidated into long-term memory through summarization, reflection, or abstraction. \textit{Processing} subsequently occurs within working memory, which functions as the active workspace for immediate inference. To support this reasoning, the system relies on \textit{retrieval} to populate the workspace with relevant context and skills drawn from the persistent stores of factual and experiential memory. 
This encoding-processing-retrieval sequence constitutes the central architectural pattern enabling agents to learn from the past simultaneously and reason in the present.

\subsection{Factual Memory}
\label{ssec:factual_memory}

Factual memory refers to the capacity of an agent to store and retrieve explicit, declarative \textbf{facts} about past events, user-specific information, and the state of the external environment.
This information encompasses a wide range of content, including dialogue history, user preferences, and relevant properties of the external world.
By allowing the agent to exploit historical information when interpreting current inputs, factual memory serves as the cornerstone for context awareness, personalized responses, and extended task planning.

To understand the structural composition of agent memory, we draw upon the cognitive science framework of \textit{declarative memory}~\citep{riedelDeclarativeMemory2015}.
In neuroscience, declarative memory denotes long-term storage for information that can be consciously accessed and is commonly analyzed in terms of two major components: \textit{episodic} and \textit{semantic} memory~\citep{squireMemorySystemsBrain2004}.
\textit{Episodic memory} stores personally experienced events associated with specific temporal and spatial contexts---the \textit{what}, \textit{where}, and \textit{when} of an episode~\citep{tulvingEpisodicSemanticMemory1972,tulvingEpisodicMemoryMind2002}. Its central characteristic is the capacity to mentally re-experience past events.
\textit{Semantic memory} retains general factual knowledge, concepts, and word meanings independent of the specific occasion on which they were acquired~\citep{squireMemorySystemsBrain2004}.
While supported by a unitary declarative system in the human brain, these components represent distinct levels of abstraction.

In agent systems, this biological distinction is operationalized not as a rigid dichotomy but as a processing \textit{continuum}.
Systems typically initiate this process by logging concrete interaction histories as episodic traces, such as dialogue turns, user actions, and environment states~\citep{zhong2023memorybankenhancinglargelanguage,wang_recmind_2024,Chhikara2025mem0}.
Subsequent processing stages apply summarization~\citep{wang2025recursivelysummarizingenableslongterm,chen2025compress}, reflection~\citep{DBLP:conf/acl/rmm2025,park2023generativeagentsinteractivesimulacra,wang2025recursivelysummarizingenableslongterm}, entity extraction~\citep{gutiérrez2025hipporagneurobiologicallyinspiredlongterm}, and fact induction~\citep{Rasmussen2025Zep}.
The resulting abstractions are stored in structures such as vector databases~\citep{zhong2023memorybankenhancinglargelanguage}, key-value stores, or knowledge graphs~\citep{Rasmussen2025Zep,sun2024knowledgegraphtuningrealtime}, governed by procedures for deduplication and consistency checking.
Through this sequence, raw event streams are gradually transformed into reusable semantic fact bases.

Functionally, this architecture ensures that the agent exhibits three fundamental properties during interaction: \textbf{consistency}, \textbf{coherence}, and \textbf{adaptability}.
\begin{itemize}
    \item \textbf{Consistency} implies stable behavior and self-presentation over time. By maintaining a persistent internal state regarding user-specific facts and its own commitments, the agent avoids contradictions and arbitrary changes of stance.
    \item \textbf{Coherence} is reflected in robust context awareness. The agent can recall and integrate relevant interaction history, refer to past user inputs, and preserve topical continuity, ensuring responses form a logically connected dialogue rather than isolated utterances.
    \item \textbf{Adaptability} demonstrates the ability to personalize behavior based on stored user profiles and historical feedback. Consequently, response style and decision-making progressively align with the user's specific needs and characteristics.
\end{itemize}

For exposition, we further organize factual memory according to the primary entity it refers to. This entity-centric taxonomy, together with representative methods and their technical design choices, is systematically summarized in \autoref{tab:factual_mem}.
This perspective highlights two central application domains:

\vspace{0.6em}
\begin{tcolorbox}[
  colback=selfevolagent_light!20,
  colframe=selfevolagent_light!80,
  colbacktitle=selfevolagent_light!80,
  coltitle=black,
  title={\bfseries\fontfamily{ppl}\selectfont{Two Types of Factual Memory}},
  boxrule=2pt,
  arc=5pt,
  drop shadow,
  parbox=false,
  before skip=5pt,
  after skip=5pt,
  left=5pt,   
  right=5pt,
]

\begin{itemize}
    \item \textbf{User factual memory} (\Cref{user_factual_mem}) denotes facts that sustain the consistency of interactions between humans and agents, including identities, stable preferences, task constraints, and historical commitments.
    \item \textbf{Environment factual memory} (\Cref{env_factual_mem}) denotes facts that sustain consistency with respect to the external world, such as document states, resource availability, and the capabilities of other agents.
\end{itemize}
\end{tcolorbox}

{
\label{tab:factual_mem}
\footnotesize


\begin{xltabular}{\textwidth}{p{5cm}|llXl}
\caption{
Taxonomy of factual memory methods. We categorize existing works based on the primary target entity: \textbf{User Factual Memory} focuses on sustaining interaction consistency, while \textbf{Environment Factual Memory} ensures consistency with the external world. Methods are compared across three technical dimensions: (1) \textbf{Carrier} (\Cref{sec:what-memory}) identifies the storage medium, (2) \textbf{Structure} follows the taxonomy of token-level memory (\Cref{ssec:token}), and (3) \textbf{Optimization} denotes the integration strategy, where \textit{PE} encompasses prompt engineering and inference-time techniques without parameter updates, distinct from gradient-based methods like \textit{SFT} and \textit{RL}.
} \\
\toprule
{Method} & {Carrier} & {Structure} & {Task} & {Optimization} \\
\midrule
\endfirsthead

\caption[]{Taxonomy of factual memory methods. (continued)}\\
\toprule
{Method} & {Carrier} & {Form} & {Task}  & {Optimization}  \\
\midrule
\endhead

\midrule
\multicolumn{5}{r}{Continued on next page}\\
\bottomrule
\endfoot

\bottomrule
\endlastfoot

\multicolumn{5}{c}{\textit{I. User factual Memory}}\\
\midrule
\multicolumn{5}{l}{\textbf{(a) Dialogue Coherence}}\\
MemGPT~\citep{packer2023memgpt} & Token-level & 1D & Long-term dialogue & PE \\
TiM~\citep{DBLP:journals/corr/abs-2311-08719}  & Token-level & 2D & QA & PE \\
MemoryBank~\citep{zhong2023memorybankenhancinglargelanguage}  & Token-level & 1D & Emotional Companion & PE \\
AI Persona~\citep{wang2024aipersona}   & Token-level & 1D & Emotional Companion & PE \\
Encode-Store-Retrieve~\citep{shenEncodeStoreRetrieveAugmentingHuman2024}  & Token-level & 1D & Multimodal QA & PE \\
Livia~\citep{DBLP:journals/corr/abs-2509-05298}   & Token-level & 1D & Emotional Companion & PE \\
mem0~\citep{Chhikara2025mem0}  & Token-level & 1D & Long-term dialogue, QA & PE \\
RMM~\citep{DBLP:conf/acl/rmm2025} & Token-level & 2D & Personalization & PE, RL \\
D-SMART~\citep{Lei2025DSMART}& Token-level & 2D & Reasoning & PE \\
Comedy~\citep{chen2025compress} & Token-level & 1D & Summary, Compression, QA & PE \\
MEMENTO~\citep{DBLP:journals/corr/abs-2505-16348}  & Token-level & 1D & Embodied, Personalization & PE \\
O-Mem~\citep{wang2025omem}  & Token-level & 3D & Personalized Dialogue & PE \\
DAM-LLM~\citep{lu2025dynamicaffectivememorymanagement} & Token-level & 1D & Emotional Companion & PE \\
MemInsight~\citep{salama2025meminsightautonomousmemoryaugmentation} & Token-level & 1D & Personalized Dialogue & PE \\
EMem~\citep{zhou2025ememsimplestrongbaselinelongterm} & Token-level & 1D & Personalized Dialogue & PE \\
RGMem~\citep{tian2025RGMem} & Token-level & 1D & Long-conv QA & PE \\
Memoria~\citep{sarin2025memoriascalableagenticmemory} & Token-level & 1D  & Long-conv QA & PE \\
MemVerse~\citep{liu2025memversemultimodalmemorylifelong} & \color{green}\ding{52} & Fact & Multimodal hierarchical knowledge graphs. & Reasoning, QA \\

\midrule
\multicolumn{5}{l}{\textbf{(b) Goal Consistency}}\\
RecurrentGPT~\citep{zhou2023recurrentgpt}  & Token-level & 1D & Long-Context Generation, Personalized Interactive Fiction & PE \\
Memolet~\citep{yenMemoletReifyingReuse2024} & Token-level & 2D & QA, Document Reasoning & PE \\
MemGuide~\citep{duMemGuideIntentDrivenMemory2025}  & Token-level & 1D & Long-conv QA & PE, SFT \\
SGMem~\citep{Wu2025SGMemSG} & Token-level& 2D & Long-context & PE\\
A-Mem~\citep{xuAMEMAgenticMemory2025}  & Token-level & 2D & QA, Reasoning & PE \\
M3-agent~\citep{long2025seeing}  & Token-level & 2D & Multimodal QA & PE, SFT \\
WorldMM~\citep{yeo2025worldmmdynamicmultimodalmemory} & Token-level & 1D & Multimodal QA & PE \\
EverMemOS~\citep{hu2026evermemosselforganizingmemoryoperating} & Token-level & 1D & Long-conv QA & PE \\

\midrule
\multicolumn{5}{c}{\textit{II. Environment factual Memory}}\\
\midrule

\multicolumn{5}{l}{\textbf{(a) Knowledge Persistence}}\\
MemGPT~\citep{packer2023memgpt} & Token-level & 1D & Document QA & PE \\
CALYPSO~\citep{zhuCALYPSOLLMsDungeon2023} & Token-level & 1D & Tabletop Gaming & PE \\ 
AriGraph~\citep{anokhin2024arigraph} & Token-level & 3D & Game, Multi-op QA & PE \\
HippoRAG~\citep{gutiérrez2025hipporagneurobiologicallyinspiredlongterm} & Token-level & 3D & QA & PE \\
WISE~\citep{DBLP:conf/nips/0104L0XY0X0C24}&Parametric&/&Document Reasoning, QA&SFT\\
MemoryLLM~\citep{Wang2024MEMORYLLM} & Parametric &/&Document Reasoning&SFT\\
Memoria~\citep{park2024memoria} & latent & / & Language Modeling & PE \\
Zep~\citep{Rasmussen2025Zep} & Token-level & 3D & Document analysis & PE \\
MemTree~\citep{rezazadehIsolatedConversationsHierarchical2025} & Token-level & 2D & Document Reasoning, Dialogue & PE \\
LMLM~\citep{zhao2025pretraininglimitedmemorylanguage}  & Token-level & 1D & QA & SFT \\
M+ ~\citep{DBLP:journals/corr/abs-2502-00592}& Latent&/&Document Reasoning, QA&SFT\\
CAM~\citep{liCAMConstructivistView2025} & Token-level & 3D & Multi-hop QA & SFT, RFT \\
MemAct~\citep{DBLP:journals/corr/abs-2510-12635} & Token-level & 1D & Multi-obj QA & RL \\
Mem-$\alpha$~\citep{DBLP:journals/corr/abs-2509-25911}&Token-Level&1D&Document Reasoning&RL\\
WebWeaver~\citep{liWebWeaverStructuringWebScale2025} & Token-level & 1D & Deep Research & SFT \\
MemLoRA~\citep{bini2025memloradistillingexpertadapters} & Parametric & / & QA & SFT\\
Memory Decoder~\citep{cao2025memorydecoder} & Parametric & / & QA, Language Modeling & SFT\\

\midrule
\multicolumn{5}{l}{\textbf{(b) Shared Access}}\\
GameGPT~\citep{chenGameGPTMultiagentCollaborative2023} & Token-level & 1D & Game Development & PE \\
Generative Agent~\citep{park2023generativeagentsinteractivesimulacra} & Token-level & 2D & Social Simulation & PE \\
S³~\citep{gaoS^mbox3SocialnetworkSimulation2023}  & Token-level & 1D & Social Simulation & PE \\
Memory Sharing~\citep{DBLP:journals/corr/abs-2404-09982}  & Token-level & 1D & Document Reasoning & PE \\
MetaGPT~\citep{hong2023metagpt} & Token-level & 1D & Software Development & PE \\
G-Memory~\citep{zhang2025g}  & Token-level & 3D & QA & PE \\
OASIS ~\citep{yang2025oasisopenagentsocial}&Token-level, Parametric&1D& Social Simulation& PE \\
\bottomrule
\end{xltabular}

}
\subsubsection{User factual memory}
\label{user_factual_mem}

User factual memory persists verifiable facts about a specific user across sessions and tasks, including identity, preferences, routines, historical commitments, and salient events.

 Its primary function is to prevent characteristic failure modes of stateless interaction, such as coreference drift, repeated elicitation, and contradictory responses, thereby reducing interruptions to long-horizon goals~\citep{DBLP:conf/acl/rmm2025,zhong2023memorybankenhancinglargelanguage}.
Engineering practice typically comprises selection and compression, structured organization, retrieval and reuse, and consistency governance, aiming to sustain \textit{long-range dialogic and behavioral coherence} under bounded access cost.

\paragraph{Dialogue Coherence} Dialogue coherence requires an agent to preserve conversational context, user-specific facts, and a stable persona over extended periods. This ensures that later turns remain sensitive to earlier disclosures and affective cues, rather than degrading into repeated clarifications or inconsistent replies. To achieve this, modern systems implement user factual memory through two complementary strategies: \textit{heuristic selection} and \textit{semantic abstraction}.

To navigate finite context windows efficiently, a primary strategy is to \textit{selectively} retain and rank interaction histories.
Rather than retaining all raw logs, systems~\citep{DBLP:journals/corr/abs-2509-05298,zhong2023memorybankenhancinglargelanguage,park2023generativeagentsinteractivesimulacra,Lei2025DSMART} maintain structured stores of past interactions, ranking entries by metrics such as \textit{relevance}, \textit{recency}, \textit{importance}, or \textit{distinctiveness}.
By filtering retrieval based on these scores, high-value items are preserved and periodically condensed into higher-level summaries, conditioning subsequent responses to maintain continuity without overwhelming the agent's working memory.

Beyond mere selection, advanced frameworks emphasize the \textit{transformation and abstraction} of raw dialogue fragments into higher-level semantic representations.
Approaches such as Think in Memory~\citep{DBLP:journals/corr/abs-2311-08719} and Reflective Memory Management~\citep{DBLP:conf/acl/rmm2025} convert raw interaction traces into \textit{thought} representations or reflections via iterative update operations. This allows the agent to query a stable semantic memory, keeping later replies topically consistent and less repetitive. Similarly, COMEDY~\citep{chen2025compress} employs a single language model to generate, compress, and reuse memory while updating compact user profiles. 
These methods effectively stabilize \textit{persona} and \textit{preference} expression over long conversational histories by decoupling memory storage from the raw token surface form.

\paragraph{Goal Consistency}
Goal consistency requires an agent to maintain and refine an explicit task representation over time. This ensures that clarifying questions, information requests, and actions remain strictly aligned with the primary objective, minimizing intent drift.

To mitigate such drift, systems utilize factual memory to dynamically track and update the task state.
Approaches like RecurrentGPT~\citep{zhou2023recurrentgpt}, Memolet~\citep{yenMemoletReifyingReuse2024}, and MemGuide~\citep{duMemGuideIntentDrivenMemory2025} retain confirmed information while highlighting unresolved elements. By guiding retrieval based on task intent, these methods help agents satisfy missing constraints and maintain focus across sessions.

For complex, long-horizon tasks, memory forms are often \textit{structured} to facilitate localized retrieval centered on the active goal~\citep{Wu2025SGMemSG}.
For instance, A-Mem~\citep{xuAMEMAgenticMemory2025} organizes memories as an interconnected graph of linked notes, while H-Mem~\citep{limbacherHMemHarnessingSynaptic2020} employs associative mechanisms to recall prerequisite facts when subsequent steps depend on prior observations.

In embodied scenarios, factual memory grounds agent behavior in user-specific habits and environmental context.
Systems such as M3-Agent~\citep{long2025seeing} and MEMENTO~\citep{DBLP:journals/corr/abs-2505-16348} persist data on household members, object locations, and routines, reusing this information to minimize redundant exploration and repeated instructions.
Similarly, Encode-Store-Retrieve~\citep{shenEncodeStoreRetrieveAugmentingHuman2024} processes egocentric visual streams into text-addressable entries, allowing agents to answer questions based on past visual experiences without requiring user repetition.

 \paragraph{Summary}
Collectively, these mechanisms transform ephemeral interaction traces into a persistent cognitive substrate.
By integrating retrieval-based ranking with generative abstraction, user factual memory upgrades the system from simple similarity matching to the active maintenance of explicit goals and constraints.
This foundation yields a dual benefit: it fosters a sense of familiarity and trust through long-term behavioral coherence, while simultaneously enhancing operational efficiency by increasing task success rates, reducing redundancy, and lowering error recovery overhead.

\subsubsection{Environment factual memory}
\label{env_factual_mem}

Environment factual memory pertains to entities and states \textit{external} to the user, encompassing long documents, codebases, tools, and interaction traces.

This memory paradigm addresses incomplete factual recall and unverifiable provenance, minimizes contradictions and redundancy in multi-agent collaboration, and stabilizes long-horizon tasks in heterogeneous environments. The central objective is to furnish an updatable, retrievable, and governable external fact layer, providing a stable reference across sessions and stages.
Concretely, we categorize existing implementations along two complementary dimensions: \textit{knowledge persistence} and \textit{multi-agent shared access}.

\paragraph{Knowledge Persistence}
Knowledge memory refers to persistent representations of world knowledge and domain-specific knowledge that support long document analysis, factual question answering, multihop reasoning, and reliable retrieval of code and data resources.

In terms of \textit{knowledge organization}, existing research focuses on structuring external data to enhance reasoning capabilities. For instance, HippoRAG~\citep{gutiérrez2025hipporagneurobiologicallyinspiredlongterm} utilizes knowledge graphs to facilitate evidence propagation, while MemTree~\citep{rezazadehIsolatedConversationsHierarchical2025} employs a dynamic hierarchical structure to optimize aggregation and targeted access in growing corpora. Regarding storage form, LMLM~\citep{zhao2025pretraininglimitedmemorylanguage} explicitly decouples factual knowledge from model weights by externalizing it into a database, thereby enabling direct knowledge edits and provenance verification without retraining. In narrative domains, CALYPSO~\citep{zhuCALYPSOLLMsDungeon2023} distills lengthy game contexts into bite-sized prose, preserving critical story state accessibility.

In scenarios requiring continuous \textit{knowledge updates}, parameter-centric approaches integrate persistence directly into the \textit{model architecture}. Methods such as MEMORYLLM~\citep{Wang2024MEMORYLLM}, M+~\citep{DBLP:journals/corr/abs-2502-00592}, and WISE~\citep{DBLP:conf/nips/0104L0XY0X0C24} incorporate trainable memory pools or side networks to absorb new information. Rather than relying solely on static external retrieval, these designs focus on the challenge of model editing, allowing agents to adapt to dynamic environments and correct obsolete facts while preserving the stability of the pre-trained backbone.

\paragraph{Shared Access} 
Shared memory establishes a visible and manageable common factual foundation for \textit{multi-agent collaboration}, serving to align goals, carry intermediate artifacts, and eliminate redundant work. By maintaining a centralized repository of past queries and responses, frameworks such as Memory Sharing~\citep{gaoMemorySharingLarge2024} enable agents to access and build on peers' accumulated insights asynchronously. This mechanism ensures that individual agents directly benefit from collective knowledge, thereby suppressing contradictory conclusions and enhancing overall system efficiency.

For complex project coordination, systems such as MetaGPT~\citep{hong2023metagpt} and GameGPT~\citep{chenGameGPTMultiagentCollaborative2023} utilize shared message pools as central workspaces for publishing plans and partial results. Similarly, G-Memory~\citep{zhang2025g} employs hierarchical memory graphs as a unified coordination medium. These architectures facilitate consistency maintenance around the current project state, which reduces communication overhead and enables the extraction of reusable workflows from historical collaborations.

In the domain of social simulation, platforms like Generative Agents~\citep{park2023generativeagentsinteractivesimulacra} and S${}^3$~\citep{gaoS^mbox3SocialnetworkSimulation2023}, alongside large-scale simulators such as OASIS~\citep{yang2025oasisopenagentsocial} and AgentSociety~\citep{piaoAgentSocietyLargeScaleSimulation2025}, model the global environment and public interaction logs as a shared memory substrate. This substrate is incrementally updated and observed by the population, allowing information to diffuse naturally among agents and supporting coherent, history-aware social dynamics at scale.

\paragraph{Summary} environment factual memory furnishes a continuously updatable, auditable, and reusable external fact layer. On the knowledge axis, it improves completeness, interpretability, and editability of factual recall through structured organization and long-term memory modules. On the collaboration axis, it maintains cross-agent and cross-stage consistency through sharing and governance, thereby enabling robust decision-making and execution under long horizons, multiple actors, and multi-source information.

\begin{figure}[!t]
    \centering
    \includegraphics[width=\linewidth]{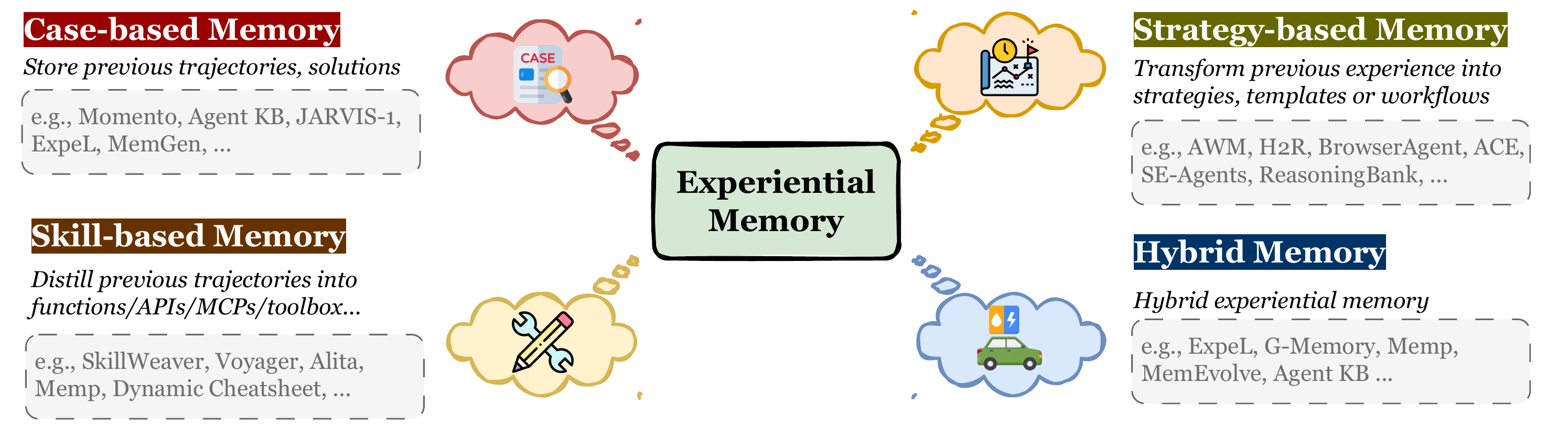}
    \caption{
    Taxonomy of experiential memory paradigms. We classify approaches based on the \textit{abstraction level} of stored knowledge: (1) \textbf{Case-based Memory}  preserves raw trajectories and solutions as concrete exemplars; (2) \textbf{Strategy-based Memory}  abstracts experiences into high-level strategies, templates, or workflows; (3) \textbf{Skill-based Memory} distills procedural knowledge into executable functions and APIs; and (4) \textbf{Hybrid Memory}  integrates multiple representations. Together, these systems mirror human \textit{procedural memory} to enable continual learning and self-evolution. This figure draws inspiration from \cite{gaoSurveySelfEvolvingAgents2025b}.
    }
    \label{fig:experiential_mem}
\end{figure}

\subsection{Experiential Memory}
\label{ssec:experiential_mem}
Experiential memory encapsulates the mechanism by which agents encode historical trajectories, distilled strategies, and interaction outcomes into durable, retrievable representations. Unlike working memory, which manages transient context, experiential memory focuses on the long-term accumulation and transfer of knowledge \textit{across} distinct episodes.

Theoretically grounded in cognitive science, this paradigm parallels human \textit{nondeclarative memory}, specifically the \textit{procedural} and \textit{habit} systems~\citep{squireMemorySystemsBrain2004,segerCriticalReviewHabit2011}. Biological systems rely on distributed neural circuits for implicit skill acquisition~\citep{reberNeuralBasisImplicit2013}. In contrast, agentic experiential memory typically employs explicit data structures, such as vector databases or symbolic logs. This implementation difference grants agents a unique capability absent in biological counterparts: the ability to introspect, edit, and reason over their own procedural knowledge.

Crucially, experiential memory serves as a foundation for \textbf{continual learning} and \textbf{self-evolution} in the \textit{era of experience}~\citep{suttonWelcomeEraExperience2025,gaoSurveySelfEvolvingAgents2025b}. By maintaining a repository of structured experiences, agents achieve a non-parametric path to adaptation and avoid the prohibitive costs of frequent parametric updates. This mechanism effectively closes the learning loop by converting interaction feedback into reusable knowledge. Through this process, agents rectify past errors, abstract generalizable heuristics, and compile routine behaviors. Consequently, such adaptation minimizes redundant computations and refines decision-making over time~\citep{DBLP:conf/aaai/Zhao0XLLH24,shinn_reflexion_2023}.

To systematically analyze existing literature, we classify experiential memory based on the \textit{abstraction level} of the stored information.
An overview of this abstraction-based taxonomy and representative paradigms is illustrated in \autoref{fig:experiential_mem}.
Representative methods under this abstraction-based taxonomy, together with their storage carriers, representation forms, and optimization strategies, are summarized in \autoref{tab:experiential_mem}.

\vspace{0.6em}
\begin{tcolorbox}[
  colback=selfevolagent_light!20,
  colframe=selfevolagent_light!80,
  colbacktitle=selfevolagent_light!80,
  coltitle=black,
  title={\bfseries\fontfamily{ppl}\selectfont{Three Types of Experiential Memory}},
  boxrule=2pt,
  arc=5pt,
  drop shadow,
  parbox=false,
  before skip=5pt,
  after skip=5pt,
  left=5pt,   
  right=5pt,
]

\begin{itemize}
    \item \textbf{Case-based Memory} (\Cref{case_based_mem}) stores minimally processed records of historical episodes, prioritizing high informational fidelity to support direct replay and imitation. By retaining the original alignment between situations and outcomes, it serves as a repository of concrete, verifiable evidence that functions as in-context exemplars for evidence-driven learning.
    \item \textbf{Strategy-based Memory} (\Cref{strategy_based_mem}) distills transferable reasoning patterns, workflows, and high-level insights from past trajectories to guide planning across diverse scenarios. Acting as a cognitive scaffold, it decouples decision-making logic from specific contexts, thereby enhancing cross-task generalization and constraining the search space for complex reasoning.
    \item \textbf{Skill-based Memory} (\Cref{skill_based_mem}) encapsulates executable procedural capacities, ranging from atomic code snippets to standardized API protocols, that operationalize abstract strategies into verifiable actions. This category serves as the agent's active execution substrate, enabling the modular expansion of capabilities and the efficient handling of tool-use environments.
\end{itemize}
\end{tcolorbox}

{
\label{tab:experiential_mem}
\footnotesize
\begin{xltabular}{\textwidth}{p{4.5cm}|llXc}
\caption{
Taxonomy of experiential memory methods.
We categorize existing works based on the \textit{abstraction level} of stored knowledge: \textbf{Case-based Memory} preserves raw records for direct replay, \textbf{Strategy-based Memory} distills abstract heuristics for planning, and \textbf{Skill-based Memory} compiles executable capabilities for action. 
Methods are compared across three technical dimensions: 
(1) \textbf{Carrier} (\Cref{sec:what-memory}) identifies the storage medium, (2) \textbf{Form} specifies the representation format of the experience,
and (3) \textbf{Optimization} denotes the integration strategy, where \textit{PE} encompasses prompt engineering and inference-time techniques without parameter updates, distinct from gradient-based methods like \textit{SFT} and \textit{RL}.
} \label{tab:experiential_mem} \\

\toprule
{Method} & {Carrier} & {Form} & {Task}  & {Optimization}  \\
\midrule
\endfirsthead

\caption[]{Taxonomy of experiential memory methods.
We categorize existing works based on the \textit{abstraction level} of stored knowledge: \textbf{Case-based Memory} preserves raw records for direct replay, \textbf{Strategy-based Memory} distills abstract heuristics for planning, and \textbf{Skill-based Memory} compiles executable capabilities for action. (continued)}\\
\toprule
{Method} & {Carrier} & {Form} & {Task}  & {Optimization}  \\
\midrule
\endhead

\midrule
\multicolumn{5}{r}{Continued on next page}\\
\bottomrule
\endfoot

\bottomrule
\endlastfoot

\multicolumn{5}{c}{\textit{I. Case-based Memory}}\\
\midrule
Expel~\citep{DBLP:conf/aaai/Zhao0XLLH24} & Token-level  & Solution  & Reasoning &  PE   \\
Synapse~\citep{zheng_synapse_2024} & Token-level & Solution & Web Interaction, Instruction-guided Web Task &  PE \\
Fincon~\citep{yu2024fincon} & Token-level & Solution &  Financial &  PE \\
MapCoder~\citep{islam_mapcoder_2024} & Token-level & Solution &  Coding & PE \\
Memento~\citep{DBLP:journals/corr/abs-2508-16153}  & Token-level & Trajectory & Reasoning  & RL \\
COLA~\citep{zhao_cola_2025} & Token-level & Trajectory  &  GUI, Web Navigation, Reasoning & PE\\
Continuous Memory~\citep{DBLP:journals/corr/abs-2510-09038} & Latent & Trajectory & GUI  & SFT \\
JARVIS-1~\citep{DBLP:journals/pami/WangCLJHZLHZYML25} & Token-level & Trajectory & Game, GUI Interaction  & PE \\
MemGen~\citep{Zhang2025MemGen} & Latent & Trajectory & Web Search, Embodied Simulation, Reasoning, Math, Code  & RL, SFT \\
Early Experience~\citep{agent-early-experience} & Parametric &  Trajectory &Embodied Simulation, Reasoning, Web Navigation  & SFT \\
DreamGym~\citep{chen2025scalingagentlearningexperience} & Token-level & Trajectory & Web Interaction, Embodied Simulation, Shopping & RL \\
MemRL~\citep{zhang2026memrlselfevolvingagentsruntime} & Token-level & Trajectory &Coding, Embodied Simulation, Reasoning  & RL \\

\midrule
\multicolumn{5}{c}{\textit{II. Strategy-based Memory}}\\
\midrule
Reflexion~\citep{shinn2023reflexion} & Token-level & Insight & Embodied Simulation, Reasoning, Coding & PE \\
Buffer of Thoughts~\citep{yang2024buffer} & Token-level & Pattern & Game, Reasoning, Coding &  PE \\
AWM~\citep{wang2024agentworkflowmemory} & Token-level & Workflow & Web Interaction, Instruction-guided Web Task &  PE \\
RecMind~\citep{wang_recmind_2024} & Token-level & Pattern &  Recommendation &  PE \\
H\({}^{\mbox{2}}\)R~\citep{DBLP:journals/corr/abs-2509-12810} & Token-level & Insight & Game, Embodied Simulation & 
  PE \\
ReasoningBank ~\citep{ouyang2025reasoningbankscalingagentselfevolving} & Token-level & Insight & Web Interaction, Instruction-guided Web Task &  PE\\
R2D2~\citep{huang_r2d2_2025} & Token-level & Insight &  Web Interaction &  PE \\
BrowserAgent~\citep{yu_browseragent_2025} & Token-level & Insight &  General QA, Web search &  RL, SFT \\
Agent KB~\citep{tang2025agentkbleveragingcrossdomain}  & Token-level & Workflow & Code, Reasoning &    PE \\

ToolMem~\citep{xiao_toolmem_2025} & Token-level & Insight & Reasoning, Image Generation & PE \\
PRINCIPLES~\citep{DBLP:journals/corr/abs-2509-17459} & Token-level & Pattern & Emotional Companion &  PE\\
SE-Agent~\citep{sun2025seagent} & Token-level & Insight & Coding  & PE\\
ACE~\citep{DBLP:journals/corr/abs-2510-04618} & Token-level & Insight & Coding, Tool calling, Financial &  PE \\
Flex~\citep{cai2025flexcontinuousagentevolution} & Token-level & Insight & Math, Chemistry, Biology & PE \\
AgentEvolver~\citep{zhai2025agentevolverefficientselfevolvingagent} & Parametric & Pattern & Tool-augmented Task & RL \\
Dynamic Cheatsheet~\citep{suzgun2025dynamiccheatsheettesttimelearning} & Token-level & Insight & Math, Reasoning, Game & PE \\
Training-Free GRPO~\citep{cai2025trainingfreegrouprelativepolicy}  & Token-level & Insight & Math, Reasoning, Web Search & PE \\
MemEvolve~\citep{zhang2025memevolvemetaevolutionagentmemory} & Token-level & Solution,Insight & Web Search, Reasoning  & PE \\

\midrule
\multicolumn{5}{c}{\textit{III. Skill-based Memory}}\\
\midrule
CREATOR~\citep{qian_creator_2023} & Token-level & Function and Script & Reasoning, Math & PE \\
Gorilla~\citep{patil_gorilla_2024} & Token-level & API & Tool calling & SFT \\
ToolRerank~\citep{zheng_toolrerank_2024}  & Token-level & API &  Tool calling & PE \\
Voyager~\citep{wang_voyager_2024} & Token-level & Code Snippet & Game &  PE \\
RepairAgent~\citep{bouzenia_repairagent_2024} & Token-level & Function and Script & Coding & PE \\
COLT~\citep{qu_colt_2024} & Token-level & API & Tool calling & SFT \\
ToolLLM~\citep{qin2024toolllm} & Token-level & API & Tool Calling &  SFT \\
LEGOMem~\citep{han_legomem_2025} & Token-level & Function and Script & Office &   PE \\
Darwin Gödel Machine~\citep{zhang_darwin_2025} & Token-level & Code Snippet & Code &  PE \\
Huxley-Gödel Machine~\citep{wang2025huxleygodelmachinehumanlevelcoding} & Token-level & Code Snippet & Code & PE \\
Memp\(^{\mbox{p}}\)~\citep{fang2025mempexploringagentprocedural} & Token-level & Function and Script & Embodied Simulation, Travel Planning &  PE\\
SkillWeaver~\citep{skillweaver} &  Token-level & Function and Script & Web Interaction, Instruction-guided Web Task &  PE \\
Alita~\citep{qiu2025alitageneralistagentenabling} & Token-level & MCP & Math, Reasoning, VQA &   PE \\
Alita-G~\citep{qiuAlitaGSelfEvolvingGenerative2025a} & Token-level & MCP & Math, Reasoning, VQA &   PE \\
LearnAct~\citep{DBLP:journals/corr/abs-2504-13805} & Token-level & Function and Script & Mobile GUI & PE \\
ToolGen~\citep{wang2025toolgen} & Parametric & API & Tool calling & SFT \\
MemTool~\citep{lumer2025memtool} & Token-level & MCP & Tool calling & SFT \\
ToolRet~\citep{shiRetrievalModelsArent2025} & Token-level & API & Web, Code, Tool Retrieval & SFT \\
DRAFT~\citep{quExplorationMasteryEnabling2025} & Token-level & API & Tool calling & PE \\
ASI~\citep{wang2025inducingprogrammaticskillsagentic} & Token-level & Functions and Scripts & Web Interaction & PE \\

\end{xltabular}

}

\subsubsection{Case-based Memory}
\label{case_based_mem}
Case-based memory stores minimally processed records of historical \textit{events}, prioritizing fidelity to ensure that episodes can be replayed or reused as in-context exemplars. Unlike strategy templates or skill modules, cases avoid extensive abstraction, thereby preserving the original alignment between situations and solutions.

\paragraph{Trajectories}
This category preserves interaction sequences to enable replay and evidence-driven learning. To optimize retrieval in text-based environments, Memento~\citep{DBLP:journals/corr/abs-2508-16153} employs soft Q-learning to dynamically refine the probability of selecting high-utility past trajectories. In multimodal settings, JARVIS-1~\citep{DBLP:journals/pami/WangCLJHZLHZYML25}, EvoVLA~\citep{liu2025evovlaselfevolvingvisionlanguageactionmodel} and Auto-scaling Continuous Memory~\citep{DBLP:journals/corr/abs-2510-09038} retain visual context, with the former storing survival experiences in Minecraft and the latter compressing GUI history into continuous embeddings. Furthermore, the early experience paradigm~\citep{agent-early-experience} constructs reward-free, agent-generated interaction traces and integrates them into model parameters via mid-training to enhance generalization.

\paragraph{Solutions}
This category treats memory as a repository of proven solutions. ExpeL~\citep{DBLP:conf/aaai/Zhao0XLLH24} autonomously gathers experience through trial-and-error, storing successful trajectories as exemplars while extracting textual insights to guide future actions. Synapse~\citep{zheng_synapse_2024} similarly injects abstracted state-action episodes as contextual examples to align problem-solving patterns. In program synthesis, MapCoder ~\citep{islam_mapcoder_2024} keeps relevant example code as a playbook-like case that multi-agent pipelines retrieve and adapt to improve reliability on complex tasks. In the financial domain, FinCon~\citep{yu2024fincon} maintains an episodic memory of past actions, PnL trajectories, and belief updates to facilitate robust cross-round decision-making.

\paragraph{Summary}
Case-based memory offers high informational fidelity and provides verifiable evidence for imitation. However, the reliance on raw data imposes challenges regarding retrieval efficiency and context window consumption. Distinguished from executable skills or abstract strategies, cases do not encompass orchestration logic or function interfaces. Instead, they serve as the factual substrate upon which higher-level reasoning operates.

\subsubsection{Strategy-based Memory}
\label{strategy_based_mem}
Unlike case libraries that retain \textit{what happened}, strategy-based memory extracts transferable knowledge of \textit{how to act}, encompassing reusable reasoning patterns, task decompositions, insights, abstractions, and cross-situational workflows. It elevates experiences into editable, auditable, and composable high-level knowledge, thereby reducing dependence on lengthy trajectory replay and improving cross-task generalization and efficiency. 
We focus on non-code or weakly code-based templates and workflows in this section, while executable functions, APIs, MCP protocols, and code snippets are classified under \Cref{skill_based_mem}.
Based on the granularity and structural complexity of the retained knowledge, we categorize strategy-based memory into three distinct types: atomic \textbf{Insights}, sequential \textbf{Workflows}, and schematic \textbf{Patterns}.

\paragraph{Insights}
This category of approaches focuses on distilling discrete pieces of knowledge, such as granular decision \textit{rules} and reflective \textit{heuristics}, from past trajectories.
H${}^2$R~\citep{DBLP:journals/corr/abs-2509-12810} explicitly decouples planning-level and execution-level memories, enabling high-level planning insights and low-level operational rules to be retrieved separately for fine-grained transfer in multi-task scenarios. 
R2D2~\citep{huang_r2d2_2025} integrates remembering, reflecting, and dynamic decision-making for web navigation, deriving corrective insights from both failed and successful cases to inform subsequent episodes.
For long-horizon web automation, BrowserAgent~\citep{yu_browseragent_2025} persists key conclusions as explicit memory to stabilize extended chains of reasoning and mitigate context drift.

\paragraph{Workflows}
Distinct from atomic, static insights, \textit{workflows} encapsulate strategies as structured \textit{sequences} of actions—executable routines abstracted from prior trajectories to guide multi-step execution at inference time.
Agent Workflow Memory (AWM)~\citep{wang2024agentworkflowmemory} induces reusable workflows on Mind2Web ~\citep{deng_mind2web_2023} and WebArena ~\citep{zhou2023webarena} and uses them as high-level scaffolds to guide subsequent generation, improving success rates and reducing steps without updating base model weights. This demonstrates that strategy templates can act as a top-level controller that complements case-level evidence.
Agent KB~\citep{tang2025agentkbleveragingcrossdomain} establishes a unified knowledge base that treats workflows as transferable procedural knowledge. It employs hierarchical retrieval, accessing workflows first to structure the strategic approach and enabling problem-solving logic reuse across diverse agent architectures.

\paragraph{Patterns}
At a higher level of abstraction, reasoning patterns function as \textit{cognitive templates} that encapsulate the structure of problem-solving, enabling agents to tackle complex reasoning tasks by instantiating these \textit{generalizable skeletons}. Buffer of Thoughts~\citep{yang2024buffer} maintains a meta-buffer of thought templates that are retrieved and instantiated to solve new problems. Similarly, ReasoningBank~\citep{ouyang2025reasoningbankscalingagentselfevolving} abstracts both successes and failures into reusable reasoning units, facilitating test-time expansion and robust learning. RecMind's self-inspiring planning algorithm~\citep{wang_recmind_2024} generates intermediate self-guidance to structure subsequent planning and tool use. In the domain of dialogue agents, PRINCIPLES~\citep{DBLP:journals/corr/abs-2509-17459} builds a synthetic strategy memory via offline self-play to guide strategy planning at inference, thereby eliminating the need for additional training. 
These advances indicate a paradigmatic shift from descriptive rules to portable reasoning structures.

\paragraph{Summary}
Strategy-based memory, which encompasses insights, workflows, and patterns, serves as a high-level scaffold to guide generative reasoning. 
Unlike case-based memory that relies on retrieving specific, raw trajectories which may be noisy or context-dependent, this form of memory distills generalizable schemas to effectively constrain the search space and improve robustness on unseen tasks. 
However, a key distinction is that these strategies function as structural guidelines rather than executable actions; they direct the planning process but do not interact with the environment directly. 
This limitation necessitates skill-based memory, discussed in the following section, which stores callable capabilities and tools. 
Ultimately, robust agents typically synergize these components: strategies provide the abstract planning logic, while skills handle the grounded execution.

\subsubsection{Skill-based Memory}
\label{skill_based_mem}
Skill memory captures an agent’s procedural capacity and operationalizes abstract strategy into verifiable actions. It encodes what the agent can do, complements declarative knowledge of what the agent knows, and anchors the perception–reasoning–action loop by providing invocable, testable, and composable executables. Recent evidence shows that language models can learn when and how to call tools and scale reliably with large tool repertoires, establishing skill memory as the execution substrate of modern agents.

Skill memory spans a continuum from internal, fine-grained code to externalized, standardized interfaces. The unifying criteria are straightforward: skills must be \textbf{callable} by the agent, their outcomes must be verifiable to support learning, and they must compose with other skills to form larger routines.

\paragraph{Code Snippets}
Executable code stored as reusable snippets offers the fastest path from experience to capability. In open-ended tasks, agents distill successful sub-trajectories into interpretable programs and reuse them across environments. Voyager ~\citep{wang_voyager_2024} exemplifies this pattern with an ever-growing skill library; the Darwin Gödel Machine ~\citep{zhang_darwin_2025} goes further by safely rewriting its own code under empirical validation, yielding self-referential and progressively more capable skill sets.

\paragraph{Functions and Scripts}
Abstracting complex behaviors into modular functions or scripts enhances reusability and generalization. Recent advancements empower agents to autonomously create specialized tools for problem-solving~\citep{qian_creator_2023,yuan_craft_2024}, and to refine tool-use capabilities through demonstrations and environmental feedback across diverse domains such as mobile GUIs, web navigation, and software engineering~\citep{fang2025mempexploringagentprocedural,skillweaver,bouzenia_repairagent_2024}. Furthermore, emergent mechanisms for procedural memory enable agents to distill execution trajectories into retrievable scripts, facilitating efficient generalization to novel scenarios~\citep{DBLP:journals/corr/abs-2504-13805,han_legomem_2025}.

\paragraph{APIs}
APIs serve as the universal interface for encapsulated skills. While earlier work focused on fine-tuning models to correctly invoke tools~\citep{schick2023toolformer,patil_gorilla_2024}, the exponential growth of API libraries has shifted the primary bottleneck to retrieval. Standard information retrieval methods often fail to capture the functional semantics of tools~\citep{shiRetrievalModelsArent2025}. Consequently, recent approaches have moved towards learning-based retrieval and reranking strategies that account for tool documentation quality, hierarchical relationships, and collaborative usage patterns to bridge the gap between user intent and executable functions~\citep{zheng_toolrerank_2024,gao_ptr_2024,qu_colt_2024,quExplorationMasteryEnabling2025}.

\paragraph{MCPs}
To reduce protocol fragmentation in API-based ecosystems, the Model Context Protocol provides an open standard that unifies how agents discover and use tools and data, including code-execution patterns that load tools on demand and cut context overhead ~\citep{qiu2025alitageneralistagentenabling,qiuAlitaGSelfEvolvingGenerative2025a}. Broad platform support indicates a convergence toward a common interface layer.

Beyond standard executables, research explores learnable memories of tool capabilities to handle uncertain neural tools, parametric integration that embeds tool symbols to unify retrieval and calling, and architecture-as-skill perspectives where specialized agents are callable modules within a modular design space ~\citep{xiao_toolmem_2025,wang2025toolgen,zhao_cola_2025}. Collectively, these strands reframe skill memory as a learnable, evolving, and orchestrable capability layer.

\paragraph{Summary}
In conclusion, skill-based memory constitutes the active execution substrate of the agent, evolving from static code snippets and modular scripts to standardized APIs and learnable architectures. It bridges the gap between abstract planning and environmental interaction by operationalizing insights from case-based and strategy-based memories into verifiable procedures. As mechanisms for tool creation, retrieval, and interoperability (e.g., MCP) mature, skill memory moves beyond simple storage, enabling a continuous loop of capability synthesis, refinement, and execution that drives open-ended agent evolution.

\subsubsection{Hybrid memory}
\label{experiential_hybrid_mem}
Advanced agent architectures increasingly adopt a \textbf{hybrid} design that integrates multiple forms of experiential memory to balance grounded evidence with generalizable logic. By maintaining a spectrum of knowledge spanning raw episodes, distilled rules, and executable skills, these systems dynamically select the most appropriate memory format, ensuring both retrieval precision and broad generalization across contexts.

A prominent direction involves coupling \textit{case-based} and \textit{strategy-based} memories to facilitate complementary reasoning. For example, ExpeL~\citep{DBLP:conf/aaai/Zhao0XLLH24} synergizes concrete trajectories with abstract textual insights, allowing agents to recall specific solutions while applying general heuristics. Agent KB~\citep{tang2025agentkbleveragingcrossdomain} employs a hierarchical structure where high-level workflows guide planning and specific solution paths provide execution details. Similarly, R2D2~\citep{huang_r2d2_2025} integrates a replay buffer of historical traces with a reflective mechanism that refines decision strategies from past errors, effectively bridging case retrieval and strategic abstraction. Complementing these, Dynamic Cheatsheet~\citep{suzgun2025dynamiccheatsheettesttimelearning} prevents redundant computation by storing accumulated strategies and problem-solving insights for immediate reuse at inference time.

Furthermore, recent frameworks strive to unify the lifecycle of memory, incorporating Skill-based components or establishing comprehensive cognitive architectures~\citep{sun2025sophiapersistentagentframework,cai2025experiencedriven}. In scientific reasoning, ChemAgent~\citep{tang2025chemagentselfupdatinglibrarylarge} constructs a self-updating library that pairs execution cases with decomposable skill modules, enabling the model to refine its chemical reasoning through accumulated experience. Taking a holistic approach, LARP~\citep{yan2023larplanguageagentroleplay} establishes a cognitive architecture for open-world games that harmonizes semantic memory for world knowledge, episodic memory for interaction cases, and procedural memory for learnable skills, ensuring consistent role-playing and robust decision-making. Finally, evolutionary systems like G-Memory~\citep{Zhang2025GMemory} and Memp~\citep{fang2025mempexploringagentprocedural} implement dynamic transitions, where repeated successful cases are gradually compiled into efficient skills, automating the shift from heavy retrieval to rapid execution. A recent effort, MemVerse~\citep{liu2025memversemultimodalmemorylifelong} combines both parametric memory and token-level prcedural memory.

\subsection{Working Memory}
\label{ssec:working_mem}
In cognitive science, working memory is defined as a capacity-limited, dynamically controlled mechanism that supports higher-order cognition by selecting, maintaining, and transforming task-relevant information in the moment~\citep{baddeleyWorkingMemoryTheories2012}. Beyond mere temporary storage, it implies active control under resource constraints.
This perspective is grounded in frameworks such as the multicomponent model and the embedded-processes account, both of which emphasize attentional focus, interference control, and bounded capacity~\citep{cowanWorkingMemoryUnderpins2014}.

When transposed to LLMs, the standard context window functions primarily as a passive, read-only buffer. Although the model can consume the window's contents during inference, it lacks explicit mechanisms to select, sustain, or transform the current workspace dynamically.
Recent behavioral evidence suggests that current models do not exhibit human-like working memory characteristics, underscoring the necessity for explicitly engineered, operable working memory mechanisms~\citep{huangLanguageModelsNot2025}.

Throughout this section, we define working memory as the set of mechanisms for the \textbf{active management and manipulation} of context within a single episode~\citep{DBLP:journals/corr/abs-2510-12635}.
The objective is to transform the context window from a passive buffer into a controllable, updatable, and interference-resistant workspace.
This transition offers immediate benefits: it increases the density of task-relevant information under fixed attention budgets, suppresses redundancy and noise, and enables the rewriting or compression of representations to preserve coherent chains of thought. We categorize these mechanisms based on the interaction dynamics.
Representative working memory approaches under this interaction-based taxonomy, together with their storage carriers, task domains, and optimization strategies, are systematically summarized in \autoref{tab:working_mem_taxonomy}.

\vspace{0.6em}
\begin{tcolorbox}[
  colback=selfevolagent_light!20,
  colframe=selfevolagent_light!80,
  colbacktitle=selfevolagent_light!80,
  coltitle=black,
  title={\bfseries\fontfamily{ppl}\selectfont{Two Types of Working Memory}},
  boxrule=2pt,
  arc=5pt,
  drop shadow,
  parbox=false,
  before skip=5pt,
  after skip=10pt,
  left=5pt,   
  right=5pt,
]

\begin{itemize}
    \item \textbf{Single-turn Working Memory} (\Cref{single_turn_working_mem}) focuses on \textbf{input condensation and abstraction}. In this setting, the system must process massive immediate inputs such as long documents or high-dimensional multimodal streams within a single forward pass. The goal is to dynamically filter and rewrite evidence to construct a bounded computational scratchpad, thereby maximizing the effective information payload per token.
    \item \textbf{Multi-turn Working Memory} (\Cref{multi_turn_working_mem}) addresses \textbf{temporal state maintenance}. In sequential interactions, the challenge is to prevent historical accumulation from overwhelming the attention mechanism. This involves maintaining task states, goals, and constraints through a continuous loop of reading, executing, and updating, ensuring that intermediate artifacts are folded and consolidated across turns.
    
\end{itemize}
\end{tcolorbox}

In summary, working memory for LLMs represents a paradigm shift towards active, within-episode context management. By aligning with the cognitive requirement of active manipulation, it suppresses interference and provides a practical solution to the engineering constraints of long-context inference.

\begin{table*}[!t] 
\centering 
\caption{
Taxonomy of working memory methods. We categorize approaches into \textbf{Single-turn} and \textbf{Multi-turn} settings based on interaction dynamics.
Methods are compared across three technical dimensions: 
(1) \textbf{Carrier} (\Cref{sec:what-memory}) identifies the storage medium, (2) \textbf{Task} specifies the evaluation domain or application scenario, and (3) \textbf{Optimization} denotes the integration strategy, where PE encompasses prompt engineering and inference-time techniques without parameter updates, distinct from gradient-based methods like SFT and RL.
} 
\label{tab:working_mem_taxonomy} 
\resizebox{\textwidth}{!}{
\begin{tabular}{p{6.5cm}|lp{8cm}c}
\toprule {Method} & {Carrier} & {Task} & {Optimization} \\

\midrule 
\multicolumn{4}{c}{\textit{I. Single-turn Working Memory}}\\ 
\midrule 

\multicolumn{4}{l}{\textbf{(a) Input Condensation}}\\

Gist~\citep{DBLP:conf/nips/Mu0G23} & Latent  & Instruction Fine-tuning & SFT \\ 
ICAE~\citep{geIncontextAutoencoderContext2024} & Latent  & Language Modeling, Instruction Fine-tuning & Pretrain, LoRA\\ 
AutoCompressors~\citep{DBLP:conf/emnlp/ChevalierWAC23} & Latent  & Langague Modeling & SFT \\ 
LLMLingua~\citep{jiangLLMLinguaCompressingPrompts2023} & Token-level  & Reasoning, Conversation, Summarization & PE \\ 
LongLLMLingua~\citep{jiangLongLLMLinguaAcceleratingEnhancing2024} & Token-level  & Multi-doc QA, Long-context, Multi-hop QA & PE \\ 
CompAct~\citep{yoonCompActCompressingRetrieved2024} & Token-level  & Document QA & SFT \\ 
HyCo2~\citep{liaoHardSoftHybrid2025} & Hybrid  & Summarization, Open-domain QA, Multi-hop QA & SFT \\
Sentence-Anchor~\citep{tarasov2025sentenceanchoredgistcompressionlongcontext} & Latent & Document QA & SFT\\
MELODI~\citep{chen2024melodiexploringmemorycompression} & Hybrid & Pretraining & Pretrain \\
R${}^3$Mem~\citep{wang2025R3Mem} & Latent & Document QA, Language Modeling & PEFT  \\

\midrule
\multicolumn{4}{l}{\textbf{(b) Observation Abstraction}}\\
Synapse~\citep{zheng_synapse_2024} & Token-level  & Computer Control, Web Navigation & PE \\
VideoAgent~\citep{wangVideoAgentLongFormVideo2024} & Token-level  & Long-term Video Understanding & PE \\ 
MA-LMM~\citep{heMALMMMemoryAugmentedLarge2024} & Latent  & Long-term Video Understanding & SFT \\
Context as Memory~\citep{DBLP:journals/corr/abs-2506-03141} & Token-level  & Long-term Video Generation & PE \\

\midrule
\multicolumn{4}{c}{\textit{II. Multi-turn Working Memory}}\\
\midrule

\multicolumn{4}{l}{\textbf{(c) State Consolidation}}\\
MEM1~\citep{zhou2025mem1learningsynergizememory} & Latent & Retrieval, Open-domain QA, Shopping & RL \\ 
MemGen~\citep{Zhang2025MemGen} & Latent & Reasoning, Embodied Action, Web Search, Coding & RL \\ 
MemAgent~\citep{yu2025memagent} & Token-level & Long-term Doc. QA & RL \\
ReMemAgent~\citep{shi2025lookreasonforwardrevisitable} & Token-level& Long-term Doc. QA & RL \\
ReSum~\citep{wuReSumUnlockingLongHorizon2025} & Token-level & Long-horizon Web Search & RL \\ 
MemSearcher~\citep{yuanMemSearcherTrainingLLMs2025} & Token-level & Multi-hop QA & SFT, RL \\
ACON~\citep{kangACONOptimizingContext2025} & Token-level & App use, Multi-objective QA & PE \\
IterResearch~\citep{chen2025iterresearchrethinkinglonghorizonagents} & Token-level & Reasoning, Web Navigation, Long-Horizon QA & RL \\
SUPO~\citep{lu2025scalingllmmultiturnrl} & Token-level & Long-horizon task & RL\\
AgentDiet~\citep{xiao2025improvingefficiencyllmagent} & Token-level & Long-horizon task & PE\\
SUMER~\citep{zheng2025goaldirectedsearchoutperformsgoalagnostic} & Token-level & QA & RL\\
Sculptor~\citep{liSculptorEmpoweringLLMs2025} & Token-level & Multi-Needle QA & PE,RL \\
AgeMem~\citep{yuAgenticMemoryLearning2026} &  Token-level & QA, Embodied Action & PE,RL \\

\midrule
\multicolumn{4}{l}{\textbf{(d) Hierarchical Folding}}\\
HiAgent~\citep{huHiAgentHierarchicalWorking2025} & Token-level & Long-horizon Agent Task & PE \\ 
Context-Folding~\citep{DBLP:journals/corr/abs-2510-11967} & Token-level & Deep Research, SWE & RL \\
AgentFold~\citep{ye2025agentfold} & Token-level & Web Search & SFT \\
DeepAgent~\citep{liDeepAgentGeneralReasoning2025} & Token-level & Tool Use, Shopping, Reasoning & RL \\ 

\midrule
\multicolumn{4}{l}{\textbf{(e) Cognitive Planning}}\\
SayPlan~\citep{ranaSayPlanGroundingLarge2023} & Token-level &  3D Scene Graph, Robotics & PE \\ 
KARMA~\citep{wangKARMAAugmentingEmbodied2025} & Token-level & Household & PE \\ 
Agent-S~\citep{agasheAgentOpenAgentic2025} & Token-level & Computer Use & PE \\ 
PRIME~\citep{DBLP:journals/corr/PRIME} & Token-level & Multi-hop QA, Knowledge-intensive Reasoning & PE \\ 

\bottomrule 
\end{tabular} 
} 

\end{table*}

\subsubsection{Single-turn Working Memory}
\label{single_turn_working_mem}
Single-turn working memory addresses the challenge of processing massive immediate inputs, including long documents~\citep{DBLP:conf/emnlp/ChevalierWAC23} and high-dimensional multimodal streams~\citep{wangVideoAgentLongFormVideo2024}, within a single forward pass.
Rather than passively consuming the entire context, the objective is to actively construct a writable workspace. This involves filtering and transforming raw information to increase density and operability under fixed attention and memory budgets~\citep{jiangLLMLinguaCompressingPrompts2023,jiangLongLLMLinguaAcceleratingEnhancing2024}.
We categorize these mechanisms into \textit{input condensation}, which reduces physical token count, and \textit{observation abstraction}, which transforms data into structured semantic representations.

\paragraph{Input Condensation}
Input condensation techniques aim to preprocess the context to minimize token usage while preserving essential information~\citep{jiangLLMLinguaCompressingPrompts2023}. These methods generally fall into three paradigms: hard, soft, and hybrid condensation~\citep{liaoHardSoftHybrid2025}.

\textit{Hard condensation} discretely selects tokens based on importance metrics. Approaches like LLMLingua~\citep{jiangLLMLinguaCompressingPrompts2023} and LongLLMLingua~\citep{jiangLongLLMLinguaAcceleratingEnhancing2024} estimate token perplexity to discard predictable or task-irrelevant content, while CompAct~\citep{yoonCompActCompressingRetrieved2024} adopts an iterative strategy to retain segments that maximize information gain. Although efficient, hard selection risks severing syntactic or semantic dependencies.
\textit{Soft condensation} encodes variable-length contexts into dense latent vectors (memory slots). Methods such as Gist~\citep{DBLP:conf/nips/Mu0G23}, In-Context Autoencoder (ICAE)~\citep{geIncontextAutoencoderContext2024}, and AutoCompressors~\citep{DBLP:conf/emnlp/ChevalierWAC23} train models to compress prompts into valid summary tokens or distinct memory embeddings. This achieves high compression ratios but requires additional training and may obscure fine-grained details. 
\textit{Hybrid} approaches like HyCo2~\citep{liaoHardSoftHybrid2025} attempt to reconcile these \textit{trade-offs} by combining global semantic adapters (soft) with token-level retention probabilities (hard).

\paragraph{Observation Abstraction} While condensation focuses on reduction, \textit{observation abstraction} aims to transform raw observations into structured formats that facilitate reasoning.
This mechanism maps dynamic, high-dimensional observation spaces into fixed-size memory states, preventing agents from being overwhelmed by raw data.

In complex interactive environments, abstraction converts verbose inputs into concise state descriptions.
Synapse~\citep{zheng_synapse_2024} rewrites unstructured HTML DOM trees into task-relevant state summaries to guide GUI automation. Similarly, in multimodal settings, processing every frame of a video stream is computationally prohibitive. Working memory mechanisms address this by extracting semantic structures: Context as Memory~\citep{DBLP:journals/corr/abs-2506-03141} filters frames based on field-of-view overlap, VideoAgent~\citep{wangVideoAgentLongFormVideo2024} converts streams into temporal event descriptions, and MA-LMM~\citep{heMALMMMemoryAugmentedLarge2024} maintains a bank of visual features. These methods effectively rewrite high-dimensional, redundant streams into low-dimensional, semantically rich representations operable within a limited context window for efficient processing.

\paragraph{Summary}
Single-turn working memory functions as an \textit{active compression layer} that maximizes the utility of the context window for immediate reasoning. By employing input condensation and observation abstraction, these mechanisms effectively increase the information density of the operational workspace, ensuring that critical evidence is retained despite capacity constraints. However, this optimization is strictly \textit{intra-turn}; it addresses the breadth and complexity of static inputs rather than the temporal continuity of dynamic interactions.

\subsubsection{Multi-turn Working Memory}
\label{multi_turn_working_mem}
Multi-turn working memory addresses a fundamentally different problem space than the single-turn setting. In long-horizon interactions, the primary bottleneck shifts from instantaneous context capacity to the continuous maintenance of \textbf{task state} and \textbf{historical relevance}. Even with extended context windows, the accumulation of history inevitably saturates attention budgets, increases latency, and induces goal drift~\citep{luScalingLLMMultiturn2025}.
To mitigate this, working memory in multi-turn settings functions as an externalized state carrier, organizing a continuous loop of reading, evaluation, and writing. The objective is to preserve critical state information accessible and consistent within a bounded resource budget. We categorize these mechanisms by their state management strategies: \textit{state consolidation}, \textit{hierarchical folding}, and \textit{cognitive planning}.

\paragraph{State Consolidation} In continuous interaction streams, state consolidation maps an ever-growing trajectory into a fixed-size state space through dynamic updates.
Treating interaction as a streaming environment, MemAgent~\citep{yu2025memagent}, and MemSearcher~\citep{yuanMemSearcherTrainingLLMs2025} employ recurrent mechanisms to update fixed-budget memory and discard redundancy, answering queries from a compact, evolving state. ReSum~\citep{wuReSumUnlockingLongHorizon2025} extends this by periodically distilling history into reasoning states, utilizing reinforcement learning to optimize summary-conditioned behavior for indefinite exploration.

Beyond heuristic summarization, ACON~\citep{kangACONOptimizingContext2025} frames state consolidation as an optimization problem, jointly compressing environment observations and interaction histories into a bounded condensation and iteratively refining compression guidelines from failure cases. IterResearch~\citep{chen2025iterresearchrethinkinglonghorizonagents} further adopts an MDP-inspired formulation with iterative workspace reconstruction, where an evolving report serves as persistent memory and periodic synthesis mitigates context suffocation and noise contamination in long-horizon research.

Regarding state representation, approaches vary to ensure constant-size footprints. MEM1~\citep{zhou2025mem1learningsynergizememory} maintains a shared internal state that merges new observations with prior memory. Distinct from explicit text, MemGen~\citep{Zhang2025MemGen} injects latent memory tokens directly into the reasoning stream.

\paragraph{Hierarchical Folding} For complex, long-horizon tasks, state maintenance requires structure beyond linear summarization. Hierarchical folding decomposes the task trajectory based on \textit{subgoals}, maintaining fine-grained traces only while a subtask is active, and \textit{folding} the completed sub-trajectory into a concise summary upon completion.

This \textit{decompose-then-consolidate} strategy allows the working memory to expand and contract dynamically. HiAgent~\citep{huHiAgentHierarchicalWorking2025} instantiates this by using subgoals as memory units, retaining only active action–observation pairs and writing back a summary after subgoal completion. Context-Folding~\citep{DBLP:journals/corr/abs-2510-11967} and AgentFold~\citep{ye2025agentfold} extend this by making the folding operation a learnable policy, training agents to autonomously determine when to branch into sub-trajectories and how to abstract them into high-level states. DeepAgent~\citep{liDeepAgentGeneralReasoning2025} further applies this to tool-use reasoning, compressing interactions into structured episodic and working memories to support fine-grained credit assignment. By replacing finished sub-trajectories with stable high-level abstractions, these methods preserve essential context while keeping the active window small.

\paragraph{Cognitive Planning} At the highest level of abstraction, working memory creates and maintains an externalized \textit{plan} or \textit{world model}. The state functions not merely as a summary of the past, but as a forward-looking structure that guides future actions.

PRIME~\citep{DBLP:journals/corr/PRIME} integrates retrieval directly into the planning loop, ensuring that memory updates actively support complex reasoning steps.
In embodied and agentic environments, treating the language model as a high-level planner elevates the plan to the core of working memory. Approaches like SayPlan employ 3D scene graphs as queryable environmental memory to scale planning across large spaces~\citep{ranaSayPlanGroundingLarge2023}. In GUI and household tasks, systems like Agent-S~\citep{agasheAgentOpenAgentic2025} and KARMA~\citep{wangKARMAAugmentingEmbodied2025} stabilize long-horizon performance by anchoring reasoning to a hierarchical plan, using memory-augmented retrieval to bridge long-term knowledge with short-term execution.

By making plans and structured environment representations the readable and writable core of working memory, agents can maintain goal consistency and revise strategies robustly against perception failures~\citep{songLLMPlannerFewShotGrounded2023}.

\paragraph{Summary}
Multi-turn working memory pivots on the construction of an operable \textbf{state carrier} rather than the retention of raw history. By integrating \textit{state consolidation} to compress continuous streams, \textit{hierarchical folding} to structure sub-trajectories, and \textit{cognitive planning} to anchor future actions, these mechanisms effectively decouple reasoning performance from interaction length. This paradigm enables agents to maintain temporal coherence and goal alignment over indefinite horizons while adhering to strict computational and memory constraints.

\section{Dynamics: How Memory Operates and Evolves?}
\label{sec:how-memory}

\begin{figure}[!h]
    \centering
    \includegraphics[width=0.85\linewidth]{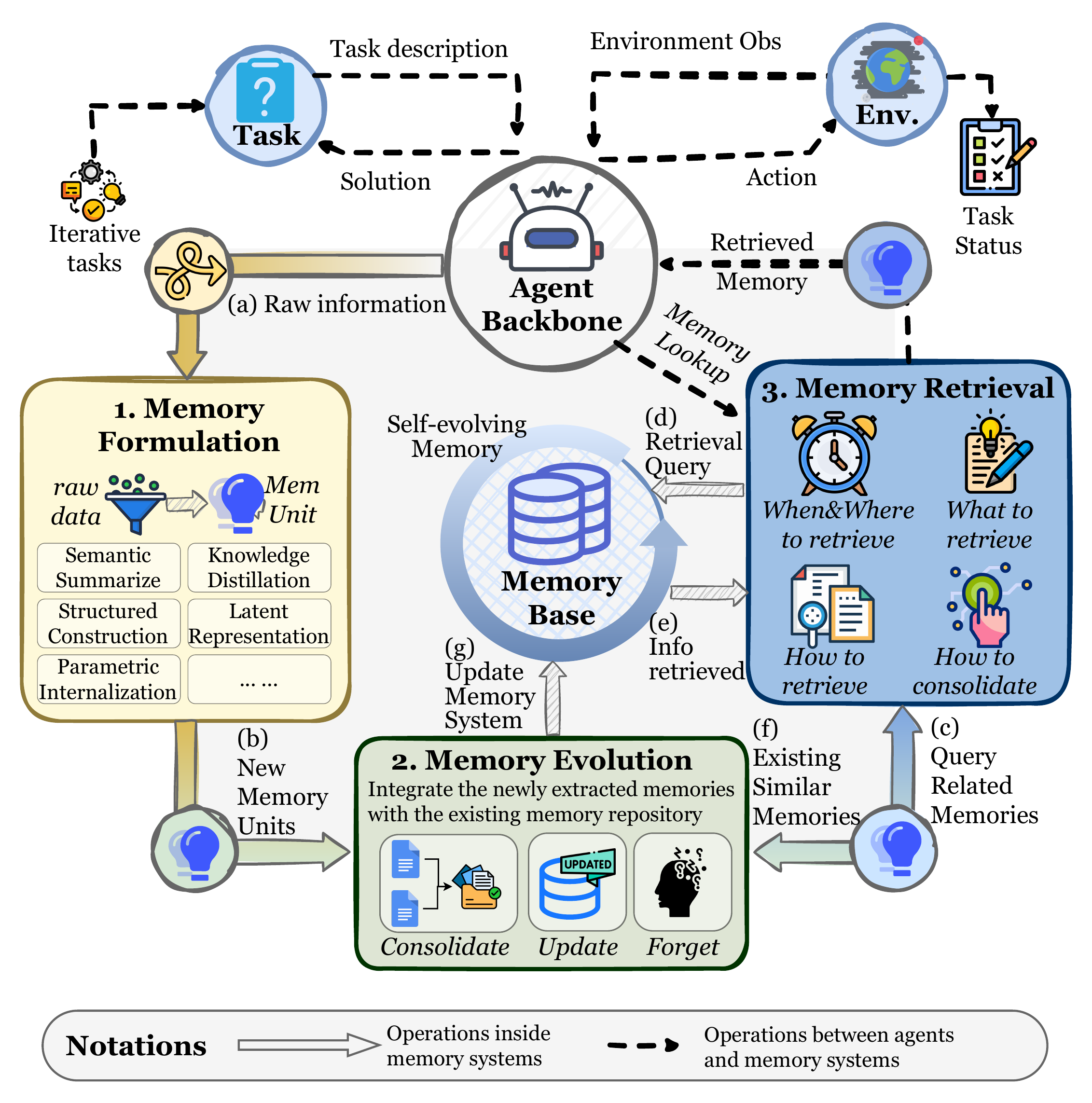}
    \caption{
    The operational dynamics of agent memory. We decouple the complete memory lifecycle into three fundamental processes that drive the system's adaptability and self-evolution: (1) \textbf{Memory Formation} transforms raw interactive experiences into information-dense knowledge units by selectively identifying patterns with long-term utility; (2) \textbf{Memory Evolution} dynamically integrates new memories into the existing repository through \textit{consolidation}, \textit{updating}, and \textit{forgetting} mechanisms to ensure the knowledge base remains coherent and efficient; and (3) \textbf{Memory Retrieval} executes context-aware queries to access specific memory modules, thereby optimizing reasoning performance with precise information support. The alphabetical order denotes the sequence of operations within the memory systems.
    }
    \label{fig:sec5-whole}
\end{figure}

The preceding sections introduced the architectural forms (\Cref{sec:what-memory}) and functional roles (\Cref{sec:why-memory}) of memory, outlining a relatively static conceptual framework for agent memory. However, such a static view overlooks the inherent dynamism that fundamentally characterizes agentic memory. Unlike knowledge that is statically encoded in model parameters or fixed databases, an agentic memory system can dynamically construct and update its memory store and perform customized retrieval conditioned on different queries. This adaptive capability is crucial for enabling agents to self-evolve and engage in lifelong learning.

Accordingly, this section investigates the paradigm shift from static storage to dynamic memory management and utilization. This paradigm shift reflects the foundational operational advantages of agentic memory over static database approaches. In practice, an agentic memory system can autonomously extract refined, generalizable knowledge based on reasoning traces and environmental feedback. By dynamically fusing and updating this newly extracted knowledge with the existing memory base, the system ensures continuous adaptation to evolving environments and mitigates cognitive conflicts. Based on the constructed memory base, the system executes targeted retrieval from designated memory modules at precise moments, thereby enhancing reasoning effectively. To systematically analyze ``how'' the memory system operates and evolves, we examine the complete memory lifecycle by decomposing it into three fundamental processes. \autoref{fig:sec5-whole} provides a holistic illustration of this dynamic memory lifecycle, highlighting how memory formation, evolution, and retrieval interact to support adaptive and self-evolving agent behavior. 

\vspace{0.5em}
\begin{tcolorbox}[
  colback=selfevolagent_light!20,
  colframe=selfevolagent_light!80,
  colbacktitle=selfevolagent_light!80,
  coltitle=black,
  title={\bfseries\fontfamily{ppl}\selectfont{Three Fundamental Process in Memory Systems}},
  boxrule=2pt,
  arc=5pt,
  drop shadow,
  parbox=false,
  before skip=5pt,
  after skip=10pt,
  left=5pt,   
  right=5pt,
]

\begin{enumerate}
    \item \textbf{Memory Formation}(\Cref{ssec:formation}): This process focuses on transforming raw experience into information-dense knowledge. Instead of passively logging all interaction history, the memory system selectively identifies information with long-term utility, such as successful reasoning patterns or environmental constraints. This part answers the question: ``How to extract the memory?''.
    \item \textbf{Memory Evolution}(\Cref{ssec:evolution}): This process represents the dynamic evolution of the memory system. It focuses on integrating newly formed memories with the existing memory base. Through mechanisms such as the consolidation of correlated entries, conflict resolution, and adaptive pruning, the system ensures that the memory remains generalizable, coherent, and efficient in an ever-changing environment. This part answers the question: ``How to refine the memory?''.
    \item \textbf{Memory Retrieval}(\Cref{ssec:retrieval}): This process determines the quality of the retrieved memory. Conditioned on the context, the system constructs a task-aware query and uses a carefully designed retrieval strategy to access the appropriate memory bank. The retrieved memory is therefore both semantically relevant and functionally critical for reasoning. This part answers the question: ``How to utilize the memory?''.
\end{enumerate}
\end{tcolorbox}

These three processes are not independent; rather, they form an interconnected cycle that drives the dynamic evolution and operation of the memory system. Memory extracted during the memory formation stage is integrated and updated with the existing memory base during the memory evolution stage. Leveraging the memory base constructed through these first two stages, the memory retrieval stage enables targeted access to optimize reasoning. In turn, reasoning outcomes and environmental feedback feed back into memory formation to extract new insights and into memory evolution to refine the memory base. Collectively, these components enable LLMs to transition from static conditional generators into dynamic systems that continuously learn from and respond to changing environments.

\subsection{Memory Formation}
\label{ssec:formation}

We define memory formation as the process of encoding raw contexts (e.g., dialogues or images) into compact knowledge. The necessity for memory formation emerges from the scaling limitations inherent in processing lengthy, noisy, and highly redundant raw contexts. Full-context prompting often encounters computational overhead, prohibitive memory footprints, and degraded reasoning performance on out-of-distribution input lengths. To mitigate these issues, recent memory systems distill essential information into efficiently storable and precisely retrievable representations, enabling more efficient and effective inference. 

Memory formation is not independent of the preceding sections. Depending on the task type, the memory formation process selectively extracts different architectural memories described in \Cref{sec:what-memory} to fulfill the corresponding functions outlined in \Cref{sec:why-memory}. Based on the granularity of information compression and the logic of encoding, we categorize the memory formation process into five distinct types.
\autoref{tab:formation} summarizes representative methods under each category, comparing their sub-types, representation forms, and key mechanisms.

\begin{tcolorbox}[
  colback=selfevolagent_light!20,
  colframe=selfevolagent_light!80,
  colbacktitle=selfevolagent_light!80,
  coltitle=black,
  title={\bfseries\fontfamily{ppl}\selectfont{Five Categories of Memory Formation Operations}},
  boxrule=2pt,
  arc=5pt,
  drop shadow,
  parbox=false,
  before skip=5pt,
  after skip=10pt,
  left=5pt,   
  right=5pt,
]

\begin{itemize} 
    \item \textbf{Semantic Summarization}(\Cref{ssec:summarization}) transforms lengthy raw data into compact summaries, filtering out redundancy while preserving global, high-level semantic information to reduce contextual overhead. 
    \item \textbf{Knowledge Distillation} (\Cref{ssec:knowledgedistill}) extracts specific cognitive assets, ranging from factual details to experiential planning strategies. 
    \item \textbf{Structured Construction}(\Cref{ssec:structured}) organizes amorphous source data into explicit topological representations, such as knowledge graphs or hierarchical trees, to enhance the explainability of memory and support multi-hop reasoning. 
    \item \textbf{Latent Representation}(\Cref{ssec:latentrepre}) encodes raw experiences directly into machine-native formats (e.g., vector embeddings or KV states) within a continuous latent space. 
    \item \textbf{Parametric Internalization}(\Cref{ssec:parametricinter}) consolidates external memories directly into the model's weight space through parameter updates, effectively transforming retrievable information into the agent's intrinsic competence and instincts. 
\end{itemize}
\end{tcolorbox}

Although we categorize these methods into five types, we argue that these memory formation strategies are not mutually exclusive. A single algorithm can integrate multiple memory formation strategies and translate knowledge across different representations~\citep{li2025memosoperatingmemoryaugmentedgeneration}.

\begin{table*}[!t]
\centering
\caption{
Taxonomy of memory formation methods. We classify approaches based on the memory formation operations.
Methods are analyzed across three technical dimensions:
(1) \textbf{Sub-Type} identifies the specific variation or scope,
(2) \textbf{Representation Form} specifies the output format, and
(3) \textbf{Key Mechanism} denotes the core algorithmic strategy.
}
\label{tab:formation}
\resizebox{\textwidth}{!}{
\begin{tabular}{l| l l l}
\toprule
Method & Sub-Type & Representation Form & Key Mechanism \\
\midrule
\multicolumn{4}{c}{\textit{I. Semantic Summarization}} \\
\midrule
MemGPT \citep{packerMemGPTLLMsOperating2023} & Incremental & Textual Summary & Merging new chunks into the working context \\
Mem0 \citep{Chhikara2025mem0} & Incremental & Textual Summary & LLM-driven summarization \\
Mem1 \citep{zhou2025mem1learningsynergizememory} & Incremental & Textual Summary & RL-optimized summarization (PPO) \\
MemAgent \citep{yu2025memagent} & Incremental & Textual Summary & RL-optimized summarization (GRPO) \\
MemoryBank \citep{zhong2023memorybankenhancinglargelanguage} & Partitioned & Textual Summary & Daily/Session-based segmentation \\
ReadAgent \citep{lee2024humaninspiredreadingagentgist} & Partitioned & Textual Summary & Semantic clustering before summarization \\
LightMem \citep{fang2025lightmem} & Partitioned & Textual Summary & Topic-clustered summarization \\
DeepSeek-OCR \citep{wei2025deepseekocr} & Partitioned & Visual Token Mapping & Optical 2D mapping compression \\
FDVS \citep{you2024towards} & Partitioned & Multimodal Summary & Multi-source signal integration (Subtitle/Object) \\
LangRepo \citep{kahatapitiya2025language} & Partitioned & Multimodal Summary & Hierarchical video clip aggregation \\
\midrule
\multicolumn{4}{c}{\textit{II. Knowledge Distillation}} \\
\midrule
TiM \citep{DBLP:journals/corr/abs-2311-08719} & Factual & Textual Insight & Abstraction of dialogue into thoughts \\
RMM \citep{Tan2025MemoTime} & Factual & Topic Insight & Abstraction of dialogue into topic-based memory\\
MemGuide \citep{duMemGuideIntentDrivenMemory2025} & Factual & User Intent & Capturing high-level user intent \\
M3-Agent \citep{long2025seeing} & Factual & Text-addressable Facts & Compressing egocentric visual observations \\
AWM \citep{wang2024agentworkflowmemory} & Experiential & Workflow Patterns & Workflow extraction from success trajectories \\
Mem$^p$ \citep{fang2025mempexploringagentprocedural} & Experiential & Procedural Knowledge & Distilling gold trajectories into abstract procedures \\
ExpeL \citep{DBLP:conf/aaai/Zhao0XLLH24} & Experiential & Experience Insight & Contrastive reflection and successful practices \\
R2D2 \citep{huang_r2d2_2025} & Experiential & Reflective Insight & Reflection on reasoning traces vs. ground truth \\
$H^{2}R$ \citep{DBLP:journals/corr/abs-2509-12810} & Experiential & Hierarchical Insight & Two-tier reflection (Plan \& Subgoal) \\
Memory-R1 \citep{yan2025memory} & Experiential & Textual Knowledge & RL-trained LLMExtract module \\
Mem-$\alpha$ \citep{DBLP:journals/corr/abs-2509-25911} & Experiential & Textual Insight & Learnable insight extraction policy \\
\midrule
\multicolumn{4}{c}{\textit{III. Structured Construction}} \\
\midrule
KGT \citep{sun2024knowledgegraphtuningrealtime} & Entity-Level & User Graph & Encoding user preferences as nodes/edges \\
Mem0$^g$ \citep{Chhikara2025mem0} & Entity-Level & Knowledge Graph & LLM-based entity and triplet extraction \\
D-SMART \citep{Lei2025DSMART} & Entity-Level & Dynamic Memory Graph & Constructing an OWL-compliant graph \\
GraphRAG \citep{edge2025localglobalgraphrag} & Entity-Level & Hierarchical KG & Community detection and iterative summarization \\
AriGraph \citep{anokhin2024arigraph} & Entity-Level & Semantic+Episodic Graph & Dual-layer (Semantic nodes + Episodic links) \\
Zep \citep{Rasmussen2025Zep} & Entity-Level & Temporal KG & 3-layer graph (Episodic, Semantic, Community) \\
RAPTOR \citep{Sarthi2024RAPTOR} & Chunk-Level & Tree Structure & Recursive GMM clustering and summarization \\
MemTree \citep{rezazadehIsolatedConversationsHierarchical2025} & Chunk-Level & Tree Structure & Bottom-up insertion and summary updates \\
H-MEM \citep{Sun2025HMEM} & Chunk-Level & Hierarchical JSON & Top-down 4-level hierarchy organization \\
A-MEM \citep{xuAMEMAgenticMemory2025} & Chunk-Level & Networked Notes & Discrete notes with semantic links \\
PREMem \citep{DBLP:journals/corr/PREMem} & Chunk-Level & Reasoning Patterns & Cross-session reasoning pattern clustering \\
CAM \citep{liCAMConstructivistView2025} & Chunk-Level & Hierarchical Graph & Disentangling overlapping clusters via replication \\
G-Memory \citep{Zhang2025GMemory} & Chunk-Level & Hierarchical Graph & 3-tier graph (interaction, query, insight) \\
\midrule
\multicolumn{4}{c}{\textit{IV. Latent Representation}} \\
\midrule
MemoryLLM \citep{Wang2024MEMORYLLM} & Textual & Latent Vector & Self-updatable latent embeddings \\
M+ \citep{DBLP:journals/corr/abs-2502-00592} & Textual & Latent Vector & Cross-layer long-term memory tokens \\
MemGen \citep{Zhang2025MemGen} & Textual & Latent Token & Latent memory trigger and weaver \\
ESR \citep{shenEncodeStoreRetrieveAugmentingHuman2024} & Multimodal & Latent Vector & Video-to-Language-to-Vector encoding \\
CoMEM \citep{Wu2025CoMEM} & Multimodal & Continuous Embedding & Vision-language compression via Q-Former \\
Mem2Ego \citep{DBLP:journals/corr/abs-2502-14254} & Multimodal & Multimodal Embedding & Embedding landmark semantics as latent memory \\
KARMA \citep{wangKARMAAugmentingEmbodied2025} & Multimodal & Multimodal Embedding & Hybrid long/short-term memory encoding \\
\midrule
\multicolumn{4}{c}{\textit{V. Parametric Internalization}} \\
\midrule
MEND \citep{mend} & Knowledge & Gradient Decomposition & Auxiliary network for fast edits \\
ROME \citep{DBLP:conf/nips/ROME} & Knowledge & Model Parameters & Causal tracing and rank-one update \\
MEMIT \citep{DBLP:conf/iclr/MEMIT} & Knowledge & Model Parameters & Mass-editing via residual distribution \\
CoLoR \citep{DBLP:journals/corr/color} & Knowledge & LoRA Parameters & Low-rank adapter training \\
ToolFormer \citep{schick2023toolformer} & Capability & Model Parameters & Supervised fine-tuning on API calls \\
\bottomrule
\end{tabular}
}
\end{table*}

\subsubsection{Semantic Summarization}
\label{ssec:summarization}
Semantic summarization transforms raw observational data into compact and semantically rich summaries. The resulting summaries capture the global, high-level information of the original data, rather than specific factual or experiential details~\citep{DBLP:conf/aaai/Zhao0XLLH24,anokhin2024arigraph}. Typical examples of such summaries include the overarching narrative of a document~\citep{kim2025nexussum,yu2025memagent}, the procedural flow of a task~\citep{ye2025agentfold,DBLP:journals/corr/abs-2510-12635}, or a user’s historical profile~\citep{zhang2024guided,DBLP:journals/corr/abs-2510-07925}. By filtering out redundant content while preserving task-relevant global semantics, semantic summarization provides a high-level guiding blueprint for subsequent reasoning without introducing excessive contextual overhead. To achieve these effects, the compression process can be implemented in two primary ways: incremental and partitioned semantic summarization. 

\paragraph{Incremental Semantic Summarization} This paradigm employs a temporal integration mechanism that continuously fuses newly observed information with the existing summary, producing an evolving representation of global semantics. This chunk-by-chunk paradigm supports incremental learning~\citep{MCCLOSKEY1989109}, circumvents the $O(n^2)$ computational burden of full-sequence processing~\citep{yu2025memagent}, and promotes progressive convergence toward global semantics~\citep{chen2024write}. Early implementations such as MemGPT~\citep{packerMemGPTLLMsOperating2023} and Mem0~\citep{Chhikara2025mem0} directly merged new chunks with existing summaries at appropriate moments, relying solely on the LLM’s inherent summarization ability. However, this approach was constrained by the model’s limited capacity, often resulting in inconsistency or semantic drift. To alleviate these issues, \citet{chen2024write} and \citet{wu2025incremental} incorporated external evaluators to filter redundant or incoherent content, using a convolutional-based discriminator for consistency verification and DeBERTa~\citep{he2020deberta} for filtering trivial content, respectively. Instead of relying on auxiliary networks, subsequent methods, such as Mem1~\citep{zhou2025mem1learningsynergizememory} and MemAgent~\citep{yu2025memagent}, enhanced the LLM’s own summarization capability through reinforcement learning with PPO~\citep{schulman2017proximal} and GRPO~\citep{shao2024deepseekmath}. 

As incremental summarization advanced from heuristic fusion to filtered integration and ultimately to learning-based optimization, the summarization competence became increasingly internalized within the model, thereby reducing cumulative errors across iterations. Nevertheless, the serial update nature still poses computational bottlenecks~\citep{fang2025lightmem} and potential information forgetting, motivating the development of Partitioned semantic summarization approaches. 

\paragraph{Partitioned Semantic Summarization} This paradigm adopts a spatial decomposition mechanism, dividing information into distinct semantic partitions and generating separate summaries for each. Early studies typically adopted heuristic partitioning strategies for handling long contexts. MemoryBank~\citep{zhong2023memorybankenhancinglargelanguage} and COMEDY~\citep{chen2025compress} summarize and aggregate long-term dialogues by treating each day or session as a basic unit. Along the structural dimension, \citet{wu2021recursively} and \citet{bailly2025divide} generate summaries of summaries by segmenting long documents into chapters or paragraphs. While intuitive, such approaches often suffer from semantic discontinuity across partitions. To address this issue, methods such as ReadAgent~\citep{lee2024humaninspiredreadingagentgist} and LightMem~\citep{fang2025lightmem} introduce semantic or topical clustering before summarization, thereby enhancing inter-chunk coherence. Extending beyond textual compression, DeepSeek-OCR~\citep{wei2025deepseekocr} pioneers the idea of compressing long contexts via optical 2D mapping, achieving higher compression ratios in multimodal scenarios. In the video memory domain, FDVS~\citep{you2024towards}  and LangRepo~\citep{kahatapitiya2025language} segment long videos into clips and generate textual summaries by integrating multi-source signals such as subtitles, object detection, and scene descriptions, which are then hierarchically aggregated into a global long video story. 

Compared with incremental summarization, the partitioned approach offers superior efficiency and captures finer-grained semantics. However, its independent processing of each sub-chunk can lead to the loss of cross-partition semantic dependencies.

\paragraph{Summary} Semantic summarization operates as a lossy compression mechanism, aiming to distill the gist from lengthy interaction logs. Unlike verbatim storage, it prioritizes global semantic coherence over local factual precision, transforming linear streams into compact narrative blocks. The primary strength of semantic summarization is efficiency: it drastically reduces context length, making it ideal for long-term dialogue. However, the trade-off is resolution loss: specific details or subtle cues may be smoothed out, limiting their utility in evidence-critical tasks.

\subsubsection{Knowledge Distillation} 
\label{ssec:knowledgedistill}
While semantic summarization captures the global semantics of raw data at a macro level, knowledge distillation operates at a finer granularity, extracting reusable knowledge from interaction trajectories or documents. In a broad sense, knowledge refers to the various forms of factual and experiential memory described in \Cref{sec:why-memory}, depending on the task’s underlying functions.

\paragraph{Distilling Factual Memory} This process focuses on transforming raw interactions and documents into explicit, declarative knowledge regarding users and environmental states. This process ensures that the agent maintains consistency and adaptability by retaining verifiable facts rather than transient context. In the domain of user modeling, systems such as TiM~\citep{DBLP:journals/corr/abs-2311-08719}, RMM~\citep{DBLP:conf/acl/rmm2025}, and EMem~\citep{zhou2025simplestrongbaselinelongterm} employ abstraction mechanisms to convert dialogue turns into high-level thoughts or enriched elementary discourse units, thereby preserving long-term persona coherence. For users' objective modeling, approaches like  MemGuide~\citep{duMemGuideIntentDrivenMemory2025} extract user intent descriptions from dialogues. During reasoning, it captures and updates goal states, separating confirmed constraints from unresolved intents to mitigate goal drift. Furthermore, this distillation extends to multimodal environments, where agents like ESR~\citep{shenEncodeStoreRetrieveAugmentingHuman2024} and M3-Agent~\citep{long2025seeing} compress egocentric visual observations into text-addressable facts about object locations and user routines. Furthermore, by equipping agents with multimodal understanding capabilities, Video-RAG~\citep{luo2025videoragvisuallyalignedretrievalaugmentedlong} converts audio, subtitles, and objects in videos into textual notes, i.e., factual memories, to enhance long-video understanding.

\paragraph{Distilling Experiential Memory} This process focuses on extracting the strategies underlying task execution from historical trajectories. By deriving planning principles from successful rollouts and corrective signals from failures, this paradigm enhances the agent’s problem-solving ability on specific tasks. Through abstraction and generalization, it further supports cross-task knowledge transfer. As a result, experiential generalization enables the agent to continually refine its competence and move toward lifelong learning.

This line of research aims to derive high-level planning strategies and key insights from both successful and failed trajectories. Some approaches focus on success-based distillation, where systems such as AgentRR~\citep{DBLP:journals/corr/agentrr} and AWM~\citep{wang2024agentworkflowmemory} summarize overall task plans from successful cases. Mem$^p$~\citep{fang2025mempexploringagentprocedural} analyzes and summarizes the gold trajectories from the training set, distilling them into abstract procedural knowledge. Others adopt failure-driven reflection, exemplified by Matrix~\citep{DBLP:journals/corr/matrix}, SAGE~\citep{DBLP:journals/ijon/SAGE}, and R2D2~\citep{huang_r2d2_2025}, which compare reasoning traces against ground-truth answers to identify error sources and extract reflective insights. Combining both, ExpeL~\citep{DBLP:conf/aaai/Zhao0XLLH24}, From Experience to Strategy~\citep{xia2025experience}, and ReMe~\citep{cao2025remembermerefineme} contrast successful and failed experiences to uncover holistic planning insights. 

However, prior work primarily focuses on summarizing task-level planning knowledge, lacking fine-grained, step-level insights. To address this gap, H$^2$R~\citep{DBLP:journals/corr/abs-2509-12810} introduces a two-tier reflection mechanism: it follows ExpeL to construct a pool of high-level planning insights, while further segmenting trajectories by subgoal sequences to derive step-wise execution insights.

Earlier methods relied on fixed prompts for insight extraction, making their performance sensitive to prompt design and the underlying LLM’s capacity. Recently, trainable distillation methods have become prevalent. Learn-to-Memorize~\citep{DBLP:journals/corr/learntomemorize} optimizes task-specific prompts for different agents. On the other hand, Memory-R1~\citep{yan2025memory} uses an LLMExtract module to obtain experiential and factual knowledge, while only the subsequent fusion component is trained to integrate these outputs into the memory bank. Although these approaches adopt an end-to-end framework, they still fall short in enhancing the LLM’s intrinsic ability to distill insights. To overcome this limitation, Mem-$\alpha$~\citep{DBLP:journals/corr/abs-2509-25911} explicitly trains the LLM on what insights to extract and how to preserve them.

\paragraph{Summary} This part focuses on extracting function-specific knowledge from the raw context, without addressing the structure of memory storage. Each piece of knowledge can be viewed as a flat memory unit. Simply storing multiple units in an unstructured table ignores the semantic and hierarchical relations among them. To address this, the memory formation process can apply structured rules to derive insights and store them within a hierarchical architecture. Simple but essential, the single knowledge distillation method introduced here serves as a foundational component for more complex and structured memory formation mechanisms.

\subsubsection{Structured Construction}
\label{ssec:structured}
While Semantic Summarization (\Cref{ssec:summarization}) and Knowledge Distillation (\Cref{ssec:knowledgedistill}) effectively compress summaries and knowledge at different levels of granularity, they often treat memory as isolated units. In contrast, Structured Construction transforms amorphous data into organized topological representations. This process is not merely a change in storage format, but an active structural operation that determines how information is linked and layered. Unlike unstructured plaintext summarization, structured extraction significantly enhances both interpretability and retrieval efficiency. Crucially, such structural prior excels at capturing complex logic and dependencies in multi-hop reasoning tasks, offering substantial advantages over traditional retrieval-augmented methods. 

Based on the operational granularity of how the underlying structure is derived, we categorize existing methods into two paradigms: Entity-Level Construction, which builds underlying topology by atomizing text into entities and relations, and Chunk-Level Construction, which builds structure by organizing intact text segments or memory items. 

\paragraph{Entity-Level Construction}  
The foundational structure of this paradigm is derived from relational triples extraction, which decomposes the raw context into its finest-grained semantic atomic entities and relations. Traditional approaches model memory as a planar knowledge graph. For instance, KGT~\citep{sun2024knowledgegraphtuningrealtime} introduces a real-time personalization mechanism where user preferences and feedback are directly encoded as nodes and edges within a user-specific knowledge graph. Similarly, $\texttt{Mem0}^{g}$~\citep{Chhikara2025mem0} utilizes LLMs to convert conversation messages directly into entities and relation triplets in the extraction phase. 

However, these direct extraction methods are often limited by the inherent capabilities of the LLM, leading to potential noise or structural errors. To improve the quality of the constructed graph, D-SMART~\citep{Lei2025DSMART} adopts a refined approach: it first employs an LLM to distill core semantic content into concise, assertion-like natural language statements, and subsequently extracts an OWL-compliant knowledge graph fragment through a neuro-symbolic pipeline. Additionally, Ret-LLM~\citep{modarressi2023ret} applies supervised fine-tuning to the LLM, enabling more robust read-write interactions with the relational graph.

While the aforementioned methods focus on planar structures, recent advancements have progressed towards constructing hierarchical memory to capture high-level abstractions. For example, GraphRAG~\citep{edge2025localglobalgraphrag} derives an entity knowledge graph from source documents and applies community detection algorithms to extract graph communities and generate community summaries iteratively. This hierarchical approach identifies higher-level cluster associations between entities, enabling the extraction of generalized insights and facilitating flexible retrieval at varying granularities. 

To better reflect the internal coherence and temporal information of the original data, some works extend the semantic knowledge graph by incorporating an episodic graph. AriGraph~\citep{anokhin2024arigraph} and HippoRAG~\citep{gutiérrez2025hipporagneurobiologicallyinspiredlongterm} establish a dual-layer structure comprising the semantic and episodic graph. They extract semantic triplets from dialogues while connecting nodes that occur simultaneously or establishing node–paragraph indices. Zep~\citep{Rasmussen2025Zep} further formalizes this into a three-layer temporal graph architecture: an episodic subgraph ($\mathcal{G}_{e}$) that logs the occurrence and processing times of raw messages via a bi-temporal model, a semantic subgraph ($\mathcal{G}_{s}$) for entities and time-bounded facts, and a community subgraph ($\mathcal{G}_{c}$) for high-level clustering and summarization of entities.

\paragraph{Chunk-Level Construction}
This paradigm treats continuous text spans or discrete memory items as nodes, preserving local semantic integrity while organizing them into topological structures. The evolution of this field progresses from static, planar (2D) extraction from fixed corpora to dynamic adaptation with incoming trajectories, and ultimately to hierarchical (3D) architectures.

Early approaches focused on organizing fixed text libraries into static planar structures. HAT~\citep{a2024enhancinglongtermmemoryusing} processes long texts by segmenting them and progressively aggregating summaries to construct a hierarchical tree. Similarly, RAPTOR~\citep{Sarthi2024RAPTOR} recursively clusters text chunks using UMAP for dimensionality reduction and Gaussian Mixture Models for soft clustering, iteratively summarizing these clusters to form a tree. However, these static methods lack the flexibility to handle streaming data without costly reconstruction.

To address this, dynamic planar approaches incrementally build memory structures as new trajectories arrive, differing based on their foundational elements. Methods based on raw text include MemTree~\citep{rezazadehIsolatedConversationsHierarchical2025} and H-MEM~\citep{Sun2025HMEM}. MemTree adopts a bottom-up approach where new text fragments retrieve the most similar nodes and are inserted as children or iteratively into a subtree, triggering bottom-up summary updates for all parent nodes. Conversely, H-MEM utilizes a top-down strategy, prompting the LLM to organize data into a four-level JSON hierarchy comprising the domain, category, memory trace, and episode layer. Alternatively, A-MEM~\citep{xuAMEMAgenticMemory2025} and PREMem~\citep{DBLP:journals/corr/PREMem} focus on reorganizing the extracted memory items. A-MEM summarizes knowledge into discrete notes and links relevant ones to construct a networked memory. PREMem clusters extracted factual, experiential, and subjective memories to identify and store higher-dimensional cross-session reasoning patterns.

Recent advancements move beyond planar layouts to construct hierarchical structures, offering richer semantic depth. SGMem~\citep{Wu2025SGMemSG} constructs a hierarchy by using NLTK to split text into sentences, forming a KNN graph across all sentence nodes, and subsequently calling an LLM to extract summaries, facts, and insights corresponding to each dialogue. To support the incremental construction of hierarchical structures as streaming data arrives, CAM~\citep{liCAMConstructivistView2025} establishes edges between text blocks based on semantic relevance and narrative coherence. It iteratively summarizes the ego graph and handles new memory insertions by explicitly disentangling overlapping clusters through node replication. In multi-agent scenarios, G-memory~\citep{Zhang2025GMemory} extends this dynamic 3D approach by maintaining three distinct graphs: an interaction graph for raw chat history, a query graph for specific tasks, and an insight graph. This structure enables each agent to receive customized memory at varying levels of granularity.

\paragraph{Summary} The main advantage of structured construction is explainability and the ability to handle complex relational queries. Such methods capture intricate semantic and hierarchical relationships between memory elements, support reasoning over multi-step dependencies, and facilitate integration with symbolic or graph-based reasoning frameworks. However, the downside is schema rigidity: pre-defined structures may fail to represent nuanced or ambiguous information, and the extraction and maintenance costs are typically high.

\subsubsection{Latent Representation}
\label{ssec:latentrepre}
The previous chapters focused on how to build token-level memory; this part focuses on encoding memory into the machine's native latent representation. Latent representation encodes raw experiences into embeddings that reside in a latent space. Unlike semantic compression and structured extraction, which summarize experiences before embedding them into vectors, latent encoding inherently stores experiences in latent space, thereby reducing information loss during summarization and text embedding. Furthermore, latent encoding is more conducive to machine cognition, enabling a unified representation across different modalities and ensuring both density and semantic richness in memory representation.

\paragraph{Textual Latent Representation} 
Although originally designed to accelerate inference, the KV cache can also be viewed as a form of latent representation within the context of memory~\citep{DBLP:journals/tmlr/kvcachemanagement,Jiang_towards_2025}. It utilizes additional memory to store past information, thereby avoiding redundant computation. MEMORYLLM~\citep{Wang2024MEMORYLLM} and M+~\citep{DBLP:journals/corr/abs-2502-00592} represent memory as self-updatable latent embeddings, which are injected into transformer layers during inference. Moreover, MemGen~\citep{Zhang2025MemGen} introduces a memory trigger to monitor the agent's reasoning state and determine when to explicitly invoke memory, as well as a memory waiver that leverages the agent’s current state to construct a latent token sequence. This sequence acts as machine-native memory, enriching the agent's reasoning capabilities.

\paragraph{Multimodal Latent Representation}
In multimodal memory research, CoMEM~\citep{Wu2025CoMEM} compresses vision-language inputs into fixed-length tokens via a Q-Former, enabling dense, continuous memory and supporting plug-and-play usage for infinite context lengths. Encode-Store-Retrieve~\citep{shenEncodeStoreRetrieveAugmentingHuman2024} converts egocentric video frames into language encodings using Ego-LLaVA, which are subsequently transformed into vector representations through an embedding model. Although embedding models are employed to ensure semantic alignment, these methods often face a trade-off between compression loss and computational overhead, particularly in handling gradient flow in long-context sequences. 

When integrated with Embodied AI, multimodal latent memory can fuse data from multiple sensors. For example, Mem2Ego~\citep{DBLP:journals/corr/abs-2502-14254} dynamically aligns global contextual information with local perception, embedding landmark semantics as latent memory to enhance spatial reasoning and decision-making in long-horizon tasks. KARMA~\citep{wangKARMAAugmentingEmbodied2025} adopts a hybrid long- and short-term memory form that encodes object information into multimodal embeddings, achieving a balance between immediate responsiveness and consistent representation. These explorations underscore the advantages of latent encoding in providing unified and semantically rich representations across modalities.

\paragraph{Summary} Latent representation bypasses human-readable formats, encoding experiences directly into machine-native vectors or KV-caches. This high-density format preserves rich semantic signals that might be lost in text decoding, enabling smoother integration with the model's internal computations. And it supports multimodal alignment seamlessly. However, it suffers from opaqueness. The latent memory is a black box, making it difficult for humans to debug, edit, or verify the knowledge it stores.

\subsubsection{Parametric Internalization}
\label{ssec:parametricinter}
As LLMs increasingly incorporate memory systems to support long-term adaptation, a central research question is how these external memories should be consolidated into parametric form. While the latent representation methods discussed above parameterize memory externally to the model, parametric internalization directly adjusts the model’s internal parameters. It leverages the model’s capacity to encode and generalize information through its learned parameter space. This paradigm fundamentally enhances the model’s intrinsic competence, eliminating the overhead of external storage and retrieval while seamlessly supporting continual updates. As we discussed in~\Cref{sec:why-memory}, not all memory content serves the same function: some entries provide declarative knowledge, while others encode procedural strategies that shape an agent’s reasoning and behavior. This distinction motivates a finer-grained view of memory internalization, separating it into knowledge internalization and capability internalization.

\paragraph{Knowledge Internalization} This strategy entails converting externally stored factual memories, such as conceptual definitions or domain knowledge, into the model’s parameter space. Through this process, the model can directly recall and utilize these facts without relying on explicit retrieval or external memory modules. In practice, knowledge internalization is typically achieved through model editing~\citep{DBLP:conf/iclr/Editable,de-cao-etal-2021-editing}. Early work, such as MEND~\citep{mend}, introduced an auxiliary network that enables rapid, single-step edits by decomposing fine-tuning gradients, thereby minimizing interference with unrelated knowledge. Building on this line of work, ROME~\citep{DBLP:conf/nips/ROME} refined the editing process by using causal tracing to precisely locate the MLP layers that store specific facts and applying rank-one updates to inject new information with higher precision and better generalization. MEMIT~\citep{DBLP:conf/iclr/MEMIT} further advanced this line by supporting batch edits, enabling thousands of facts to be updated simultaneously through multi-layer residual distributions and batch formulas, which substantially improves scalability. With the rise of parameter-efficient paradigms like LoRA~\citep{hu2022lora}, knowledge internalization can be performed through lightweight adapters rather than direct parameter modification. For instance, CoLoR~\citep{DBLP:journals/corr/color} freezes the pretrained Transformer parameters and trains only small LoRA adapters to internalize new knowledge, avoiding the high cost of full-parameter fine-tuning. Despite these advances, these approaches can still incur off-target effects~\citep{de-cao-etal-2021-editing} and remain vulnerable to catastrophic forgetting in continual learning scenarios.

\paragraph{Capability Internalization} This strategy seeks to embed experiential knowledge, such as procedural expertise or strategic heutistics, into the model's parameter space. The paradigm represents a memory formation operation in a broad sense, shifting from the acquisition of factual knowledge to the internalization of experiential capabilities. Specifically, these capabilities include domain-specific solution schemas, strategic planning, and the effective deployment of agentic skills, among others. Technically, capability internalization is achieved by learning from reasoning traces, through supervised fine-tuning~\citep{wei2022finetunedlanguagemodelszeroshot,zelikman2022,schick2023toolformer,mukherjee2023orcaprogressivelearningcomplex} or preference-guided optimization methods such as DPO~\citep{rafailov2023direct,tunstall2023zephyrdirectdistillationlm,DBLP:conf/icml/selfrewarding,dubey2024llama} and GRPO~\citep{shao2024deepseekmath,deepseekai2025deepseekr1incentivizingreasoningcapability}. As an attempt to integrate external RAG with parameterized training, Memory Decoder~\citep{cao2025memorydecoder} is a plug-and-play method that does not modify the base model, like external RAG, while achieving parameter-internalized inference speed by eliminating external retrieval overhead. Such plug-and-play parameterized memory may have broad potential.

\paragraph{Summary} Parametric internalization represents the ultimate consolidation of memory, where external knowledge is fused into the model's weights via gradients. This shifts the paradigm from retrieving information to possessing capability, mimicking biological long-term potentiation. As knowledge becomes effectively instinctive, access is zero-latency, enabling the model to respond immediately without querying external memory. However, this approach faces several challenges, including catastrophic forgetting and high update costs. Unlike external memory, parameterized internalization is difficult to modify or remove precisely without unintended side effects, limiting flexibility and adaptability. 

\subsection{Memory Evolution}
\label{ssec:evolution}
Memory Formation introduced in \Cref{ssec:formation} extracts memory from raw data. The next important step is to integrate the newly extracted memories with the existing memory repository, enabling the dynamic evolution of the memory system. A naive strategy is simply appending new entries to the existing memory bank. However, it overlooks the semantic dependencies and potential contradictions between memory entries and neglects the temporal validity of information. To address these limitations, we introduce Memory Evolution. This mechanism consolidates new and existing memories to synthesize high-level insights, resolve logical conflicts, and prune obsolete data. By ensuring the compactness, consistency, and relevance of long-term knowledge, this approach enables the memory system to adapt its cognitive processes and contextual understanding as environments and tasks evolve.

Based on the objectives of memory evolution, we categorize it into the following mechanisms:

\begin{tcolorbox}[
  colback=selfevolagent_light!20,
  colframe=selfevolagent_light!80,
  colbacktitle=selfevolagent_light!80,
  coltitle=black,
  title={\bfseries\fontfamily{ppl}\selectfont{Three Mechanisms of Memory Evolution}},
  boxrule=2pt,
  arc=5pt,
  drop shadow,
  parbox=false,
  before skip=5pt,
  after skip=5pt,
  left=5pt,   
  right=5pt,
]

\begin{itemize}
    \item \textbf{Memory Consolidation} (\Cref{ssec:consolidation}) merges new and existing memories and performs reflective integration, forming more generalized insights. This ensures that learning is cumulative rather than isolated.
    \item \textbf{Memory Updating} (\Cref{ssec:updating}) resolves conflicts between new and existing memories, correcting and supplementing the repository to maintain accuracy and relevance. It allows the agent to adapt to changes in the environment or task requirements.
    \item \textbf{Memory Forgetting} (\Cref{ssec:forgetting}) removes outdated or redundant information, freeing capacity and improving efficiency. This prevents performance degradation due to knowledge overload and ensures that the memory repository remains focused on actionable and current knowledge.
\end{itemize}
\end{tcolorbox}

\begin{figure}[!ht]
    \centering
    \includegraphics[width=0.8\linewidth]{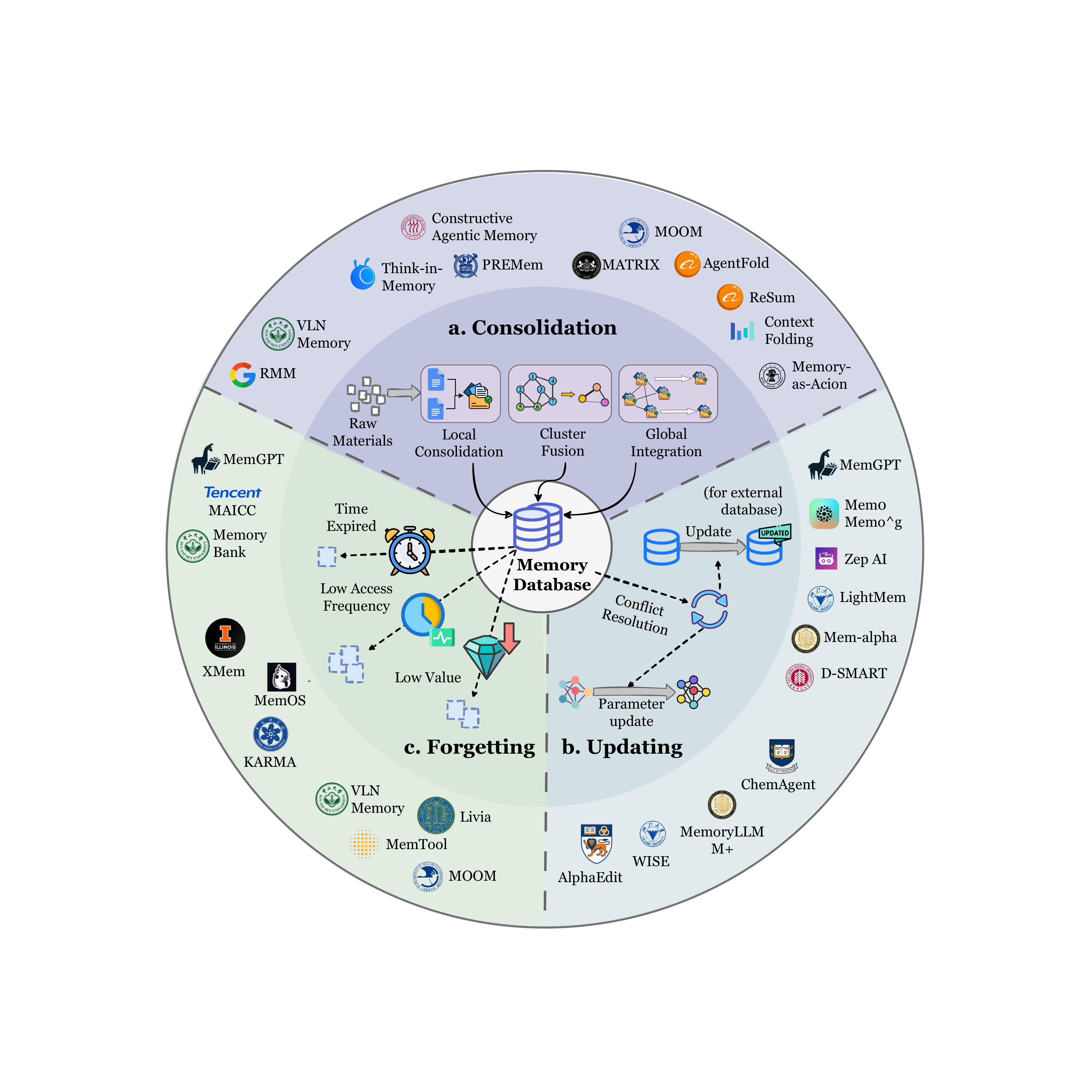}
    \caption{
    The landscape of Memory Evolution mechanisms. We categorize the evolution process into three distinct branches that maintain the central \textbf{Memory Database}:
    (a) \textbf{Consolidation} synthesizes insights by processing raw materials through local consolidation, cluster fusion, and global integration;
    (b) \textbf{Updating} ensures accuracy and consistency by performing conflict resolution on external databases and applying parameter updates to the internal model; and
    (c) \textbf{Forgetting} optimizes efficiency by pruning data based on specific criteria: time expiration, low access frequency, and low informational value.
    The outer ring displays representative frameworks and agents associated with each evolutionary mechanism.
    }
    \label{fig:sec5-mem-evolution}
\end{figure}

These mechanisms collectively maintain the generalization, accuracy, and timeliness of the memory repository. By actively managing memory evolution, these mechanisms underscore the agentic capabilities of the memory system, facilitating continuous learning and autonomous self-improvement.
\autoref{fig:sec5-mem-evolution} provides a unified view of these memory evolution mechanisms, illustrating their operational roles and representative frameworks within a shared memory database.

\subsubsection{Consolidation}
\label{ssec:consolidation}
Memory consolidation aims to transform newly acquired short-term traces into structured and generalizable long-term knowledge. Its core mechanism is to identify semantic relationships between new and existing memories and to integrate them into higher-level abstractions or insights. This process serves two main purposes. First, it reorganizes fragmented pieces of information into coherent structures, preventing the loss of critical details during short-term retention and enabling the formation of stable knowledge schemas. Second, by abstracting, compressing, and generalizing experiential data, consolidation extracts reusable patterns from specific events, yielding insights that support cross-task generalization.

A central challenge is determining the granularity at which new memories should be matched and merged with existing ones. Prior work spans a spectrum of consolidation strategies, from local content merging to cluster-level fusion and global integration.

\paragraph{Local Consolidation} This operation focuses on fine-grained updates involving highly similar memory fragments. In RMM~\citep{DBLP:conf/acl/rmm2025}, each new topic memory retrieves its top-K most similar candidates, and an LLM decides whether merging is appropriate, thereby reducing the risk of incorrect generalization. In multimodal settings, VLN~\citep{DBLP:conf/cvpr/VLN} triggers a pooling mechanism when capacity is saturated. It identifies the most similar or redundant memory pairs and compresses them into higher-level abstractions. These approaches refine detailed knowledge while preserving the global structure of the memory store, improving precision and storage efficiency. However, they cannot fully capture cluster-level relations or the higher-order dependencies that emerge across semantically related memories. 

\paragraph{Cluster-level Fusion} Adopting cluster-level fusion is essential for capturing cross-instance regularities as memory grows. Across clusters, PREMem~\citep{DBLP:journals/corr/PREMem} aligns new memory clusters with similar existing ones and applies fusion modes such as generalization and refinement to form higher-order reasoning units, substantially improving interpretability and reasoning depth. EverMemOS~\citep{hu2026evermemosselforganizingmemoryoperating} computes the similarity between a newly generated MemCell and the centroids of all MemScenes, and merges it into the MemScene that is sufficiently similar. Within a cluster, TiM~\citep{DBLP:journals/corr/abs-2311-08719} periodically invokes an LLM to examine memories that share the same hashing bucket and merges semantically redundant entries. CAM~\citep{liCAMConstructivistView2025} merges all nodes within the target cluster into a representative summary, yielding higher-level and consistent cross-sample representations. These methods reorganize the memory structure at a broader scale and mark an important step toward structured knowledge.
\paragraph{Global Integration} This operation performs holistic consolidation to maintain global coherence and to distill system-level insights from accumulated experience. Compared with \Cref{ssec:summarization}, semantic summarization focuses on deriving a global summary from the existing context and can be viewed as the initial construction of the summary. In contrast, this paragraph emphasizes how new information is integrated into an existing summary as additional data arrives. For user factual memory, MOOM~\citep{chen2025moommaintenanceorganizationoptimization} constructs stable role profiles by integrating temporary role snapshots with historical traces using rule-based processing, embedding methods, and LLM-driven abstraction. For experiential memory, Matrix~\citep{DBLP:journals/corr/matrix} performs iterative optimization to combine execution trajectories and reflective insights with global memory, distilling task-agnostic principles that support reuse across scenarios. As single-step reasoning contexts and environmental feedback lengthen, methods like AgentFold~\citep{ye2025agentfold} and Context Folding~\citep{DBLP:journals/corr/abs-2510-12635} internalize the ability to compress working memory. In multi-step interactions, including web navigation, these methods automatically summarize and condense the global context after each step, supporting efficient and effective reasoning. Global integration consolidates high-level, structured knowledge from the complete history of experience, providing a reliable contextual foundation while improving generalization, reasoning accuracy, and personalized decision-making.

\paragraph{Summary} Consolidation is the cognitive process of reorganizing fragmented short-term traces into coherent long-term schemas. It moves beyond simple storage to synthesize connections between isolated entries, forming a structured worldview. It enhances generalization and reduces storage redundancy. However, it risks information smoothing, where outlier events or unique exceptions are lost during the abstraction process, potentially reducing the agent's sensitivity to anomalies and specific events.

\subsubsection{Updating}
\label{ssec:updating}
Memory Update refers to the process by which an agent revises or replaces its existing memory when conflicts arise or new information is acquired. The goal is to maintain factual consistency and continual adaptation without full model retraining. Unlike memory consolidation described in \Cref{ssec:consolidation}, which focuses on abstraction and generalization, memory update emphasizes localized correction and synchronization, enabling the agent to remain aligned with an evolving environment.

Through continuous updating, agentic memory systems preserve the accuracy and timeliness of knowledge, preventing outdated information from biasing reasoning. It is thus a core mechanism for achieving lifelong learning and self-evolution. Depending on where the memory resides, updates fall into two categories: (1) External Memory Update: updates to external memory stores and (2) Model Editing: model-internal editing within the parameter space.

\paragraph{External Memory Update} Entries in vector databases or knowledge graphs are revised whenever contradictions or new facts emerge. Instead of altering model weights, this approach maintains factual alignment through dynamic modifications of external storage. Static memories inevitably accumulate outdated or conflicting entries, leading to logical inconsistencies and reasoning errors. Updating external memories enables lightweight corrections while avoiding the cost of full retraining or re-indexing.

The development of external memory update mechanisms has progressed along a trajectory, moving from rule-based corrections to temporally aware soft deletion, then to delayed-consistency strategies, and ultimately to fully learned update policies. 
Early systems such as MemGPT~\citep{packerMemGPTLLMsOperating2023}, D-SMART~\citep{Lei2025DSMART}, and Mem0$^g$~\citep{Chhikara2025mem0} followed a straightforward pipeline in which the LLM detects conflicts between new information and then invokes replace or delete operations to update the memory.
Although effective for basic factual repair, these systems relied on destructive replacement, erasing valuable historical context and breaking temporal continuity. To address this issue, Zep~\citep{Rasmussen2025Zep} introduced temporal annotations, marking conflicting facts with invalid timestamps rather than deleting them, thereby preserving both semantic consistency and temporal integrity. This marked a shift from hard replacement to soft, time-aware updating. However, real-time updates impose significant computational and I/O burdens under high-frequency interaction. MOOM~\citep{chen2025moommaintenanceorganizationoptimization} and LightMem~\citep{fang2025lightmem}, therefore, introduced dual-phase updating: a soft online update for real-time responsiveness, followed by an offline reflective consolidation phase where similar entries are merged and conflicts resolved via LLM reasoning. This eventual consistency paradigm balances latency and coherence. As agentic reinforcement learning matured, it became possible to enhance the LLM’s intrinsic memory update decision-making through reinforcement learning. Mem-$\alpha$~\citep{DBLP:journals/corr/abs-2509-25911} formulated memory updating as a policy-learning problem, enabling the LLM to learn when, how, and whether to update, thereby achieving dynamic trade-offs between stability and freshness.

Overall, external memory updates have transitioned from manually triggered corrections to self-regulated, temporally aware learning processes, maintaining factual consistency and structural stability through LLM-driven retrieval, conflict detection, and revision.

\paragraph{Model Editing}
Model editing performs direct modifications within the model’s parameter space to correct or inject knowledge without full retraining, representing implicit knowledge updates. Retraining is costly and prone to catastrophic forgetting. Model editing enables precise, low-cost corrections that enhance adaptability and internal knowledge retention.

Approaches of model editing fall into two main categories. (1) Explicit localization and modification: ROME~\citep{Tan2025MemoTime} identifies the parameter region encoding specific knowledge via gradient tracing and performs targeted weight updates; Model Editor Networks~\citep{tang2025chemagentselfupdatinglibrarylarge} trains an auxiliary meta-editor network to predict optimal parameter adjustments.
(2) Latent-space self-updating: MEMORYLLM~\citep{xuAMEMAgenticMemory2025} embeds a memory pool within Transformer layers, periodically replacing memory tokens to integrate new knowledge; M+~\citep{DBLP:journals/corr/abs-2502-00592} maintains dual-layer memories, discarding obsolete short-term entries and compressing key information into long-term storage.

Hybrid approaches such as ChemAgent~\citep{tang2025chemagentselfupdatinglibrarylarge} further combine external memory updates with internal model editing, synchronizing factual and representational changes for rapid cross-domain adaptation.

\paragraph{Summary} From an implementation standpoint, memory updating focuses on resolving conflicts and revising knowledge triggered by the arrival of new memories, whereas memory consolidation emphasizes the integration and abstraction of new and existing knowledge. The two memory updating strategies discussed above establish a dual-pathway mechanism involving conflict resolution in external databases and parameter editing within the model, enabling agents to perform continuous self-correction and support long-term evolution. The key challenge is the stability–plasticity dilemma: determining when to overwrite existing knowledge versus when to treat new information as noise. Incorrect updates can overwrite critical information, leading to knowledge degradation and faulty reasoning.

\subsubsection{Forgetting}
\label{ssec:forgetting}
Memory forgetting refers to the deliberate removal of outdated, redundant, or low-value information to free capacity and maintain focus on salient knowledge. Unlike update mechanisms, which resolve conflicts between memories, forgetting prioritizes eliminating outdated information to ensure efficiency and relevance. Over time, unbounded memory accumulation leads to increased noise, retrieval delays, and interference from outdated knowledge. Controlled forgetting helps mitigate overload and maintain cognitive focus. Yet, overly aggressive pruning risks erasing rare but essential knowledge, harming reasoning continuity in long-term contexts. 

Forgetting mechanisms can be categorized into Time-based Forgetting, Frequency-based Forgetting, and Importance-driven Forgetting, corresponding respectively to creation time, retrieval activity, and integrated semantic valuation.

\paragraph{Time-based Forgetting}
Time-driven forgetting considers only the creation time of memories, gradually decaying their strength over time to emulate human memory fading. MemGPT~\citep{packerMemGPTLLMsOperating2023} evicts the earliest messages upon context overflow. \citet{xuAMEMAgenticMemory2025} and \citet{DBLP:journals/corr/abs-2502-00592} employ stochastic token replacement, with a replacement ratio of K/N, to simulate exponential forgetting in human cognition, discarding the oldest entries once the pool exceeds capacity. Unlike explicit deletion of old memories, MAICC~\citep{jiang2025maicc} implements soft forgetting by gradually decaying the weights of memories over time. This process mirrors natural forgetting, ensuring continuous adaptation without historical overload. 

\paragraph{Frequency-based Forgetting}
Frequency-driven forgetting prioritizes memory based on retrieval behavior, retaining frequently accessed entries while discarding inactive ones. XMem~\citep{cheng2022xmemlongtermvideoobject} employs an LFU policy to remove low-frequency entries; KARMA~\citep{wangKARMAAugmentingEmbodied2025} uses counting Bloom filters to track access frequency; MemOS~\citep{li2025memosoperatingmemoryaugmentedgeneration} applies an LRU strategy, removing long-unused items while archiving highly active ones. This ensures efficient retrieval and storage equilibrium.
By distinguishing between creation time and retrieval frequency, these two axes form a more orthogonal taxonomy: time-based decay captures natural temporal aging, while frequency-based forgetting reflects usage dynamics, together maintaining system efficiency and recency.

\paragraph{Importance-driven Forgetting}
Importance-driven forgetting integrates temporal, frequency, and semantic signals to retain high-value knowledge while pruning redundancy. Early works such as \citet{zhong2023memorybankenhancinglargelanguage} and \citet{chen2025moommaintenanceorganizationoptimization} quantified importance via composite scores combining temporal decay and access frequency, achieving numeric-based selective forgetting. Later methods evolved toward semantic-level evaluation: VLN~\citep{DBLP:conf/cvpr/VLN} pools semantically redundant memories via similarity clustering, while Livia~\citep{DBLP:journals/corr/abs-2509-05298} incorporates emotional salience and contextual relevance to model emotion-driven selective forgetting. As LLMs develop increasingly powerful judgment capabilities, TiM~\citep{DBLP:journals/corr/abs-2311-08719} and MemTool~\citep{lumer2025memtool} leverage LLMs to assess memory importance and explicitly prune or forget less important memories. 
This shift reflects a transition from static numeric scoring to semantic intelligence. Agents can now perform conscious forgetting and selectively retain memories most pertinent to the task context, semantics, and affective cues. 

\paragraph{Summary} Time-based decay reflects the natural temporal fading of memory, frequency-based forgetting ensures efficient access to frequently used memories, and importance-driven forgetting introduces semantic discernment. These three forgetting mechanisms jointly govern how agentic memory remains timely, efficiently accessible, and semantically relevant. However, heuristic forgetting mechanisms like LRU may eliminate long-tail knowledge, which is seldom accessed but essential for correct decision-making. Therefore, when storage cost is not a critical constraint, many memory systems avoid directly deleting certain memories.

The above discussion describes how prior work manually designs memory architectures at different stages, enabling agents’ memory contents to evolve online. More recently, MemEvolve~\citep{zhang2025memevolvemetaevolutionagentmemory} proposes a meta-evolutionary framework that jointly evolves both agents’ experiential knowledge and their underlying memory architecture, allowing the memory framework itself to continuously learn and adapt.

\subsection{Memory Retrieval}
\label{ssec:retrieval}
\begin{figure}[!t]
    \centering
    \resizebox{\linewidth}{!}{
    \begin{forest}
      for tree={
        grow'=east,                  
        forked edges,                
        draw=black!35,               
        rounded corners=2pt,        
        align=left,                  
        font=\sffamily\small,      
        inner xsep=4pt,
        inner ysep=3pt,
        l sep=10pt,                  
        s sep=6pt,                  
        edge={thick, gray!60},      
        parent anchor=east,
        child anchor=west,
        anchor=west,
        base=left,
        edge path={
          \noexpand\path [\forestoption{edge}]
            (!u.parent anchor) -- +(5pt,0) |- (.child anchor)\forestoption{edge label};
        },
      },
      root style/.style={
        fill=gray!15,
        draw=gray!60,
        text=black,
        font=\sffamily\bfseries\normalsize,
        minimum width=3.3cm,
        align=center,
        inner xsep=6pt,
        inner ysep=5pt,
      },
      level 1/.style={
        font=\sffamily\bfseries\small,
        minimum width=3.4cm,
        inner xsep=5pt,
        inner ysep=4pt,
      },
      level 2/.style={
        fill=white,
        draw=black!25,
        text=black,
        line width=0.9pt,
        font=\sffamily\bfseries\footnotesize,
        minimum width=2.7cm,
        inner xsep=4pt,
        inner ysep=3pt,
      },
      level 3/.style={
        fill=white,
        draw=black!25,
        text=black,
        font=\scriptsize,  
        line width=0.6pt,
        edge={black!35},
        align=left,
        text width=8.5cm,           
        inner xsep=3pt,
        inner ysep=3pt,
        anchor=west,
      }
      [Memory Retrieval \\ (\Cref{ssec:retrieval}), root style
        [Timing \& Intent \\ (\Cref{ssec:retrievaltiming}),
          level 1,
          fill=BlueGreen!15,
          draw=BlueGreen!70,
          edge={thick, BlueGreen!70}
            [Automated Timing, level 2
                [{MemGPT \citep{packerMemGPTLLMsOperating2023}, MemTool \citep{lumer2025memtool} \\ComoRAG \citep{DBLP:journals/corr/comorag}, PRIME \citep{DBLP:journals/corr/PRIME} \\ MemGen \citep{Zhang2025MemGen}}, level 3]
            ]
            [Automated Intent, level 2
                [{AgentRR \citep{DBLP:journals/corr/agentrr}, MemOS \citep{li2025memosoperatingmemoryaugmentedgeneration} \\ H-MEM \citep{Sun2025HMEM}}, level 3]
            ]
        ]
        [Query Construction \\ (\Cref{ssec:retrievalquery}),
          level 1,
          fill=Periwinkle!20,
          draw=Periwinkle!70,
          edge={thick, Periwinkle!70}
            [Decomposition, level 2
                [{Visconde \citep{DBLP:conf/ecir/visconde}, ChemAgent \citep{tang2025chemagentselfupdatinglibrarylarge} \\ PRIME \citep{DBLP:journals/corr/PRIME}, MA-RAG \citep{DBLP:journals/corr/MA-RAG} \\ Agent KB \citep{tang2025agentkbleveragingcrossdomain}, Auto-RAG \citep{DBLP:journals/corr/autorag}}, level 3]
            ]
            [Rewriting, level 2
                [{HyDE \citep{DBLP:conf/acl/HyDE23}, MemoRAG \citep{DBLP:conf/www/Qian0ZMLD025} \\ MemGuide \citep{duMemGuideIntentDrivenMemory2025}, ToC \citep{kim2023treeclarificationsansweringambiguous} \\ Rewrite-Retrieve-Read \citep{ma2023queryrewrite}\\
                Auto-RAG \citep{DBLP:journals/corr/autorag}}, level 3]
            ]
        ]
        [Retrieval Strategies \\ (\Cref{ssec:retrievalstrategy}),
          level 1,
          fill=Goldenrod!20,
          draw=Goldenrod!70,
          edge={thick, Goldenrod!70}
            [Lexical Retrieval, level 2
                [{Agent KB \citep{tang2025agentkbleveragingcrossdomain}, MemAlpha \citep{DBLP:journals/corr/abs-2509-25911} \\ SeCom \citep{panSeCom2025}}, level 3]
            ]
            [Semantic Retrieval, level 2
                [{MemGPT \citep{packer2023memgpt}, MemoryBank \citep{zhong2023memorybankenhancinglargelanguage}\\
                Voyager \citep{wang_voyager_2024}, $\text{Memory}^3$ \citep{DBLP:journals/corr/abs-2407-01178}\\A-MEM \citep{xuAMEMAgenticMemory2025}, RMM \citep{DBLP:conf/acl/rmm2025} \\TIM \citep{du2025rethinkingmemoryaitaxonomy}, MA-RAG \citep{DBLP:journals/corr/MA-RAG} \\MemoRAG \citep{DBLP:conf/www/Qian0ZMLD025}, R2D2 \citep{huang_r2d2_2025}}, level 3]
            ]
            [Graph Retrieval, level 2
                [{AriGraph \citep{anokhin2024arigraph}, EMG-RAG \citep{wang2024craftingpersonalizedagentsretrievalaugmented} \\ Mem0$^g$ \citep{Chhikara2025mem0}, SGMem \citep{Wu2025SGMemSG} \\ HippoRAG \citep{gutiérrez2025hipporagneurobiologicallyinspiredlongterm}, CAM \citep{liCAMConstructivistView2025}\\ D-SMART \citep{Lei2025DSMART}, Zep \citep{Rasmussen2025Zep}\\ MemoTime \citep{Tan2025MemoTime}}, level 3]
            ]
            [Hybrid Retrieval, level 2
                [{Agent KB \citep{tang2025agentkbleveragingcrossdomain}, MIRIX \citep{wang2025mirixmultiagentmemoryllmbased} \\ Semantic Anchoring \citep{DBLP:journals/corr/SemanticAnchor}\\ Generative Agents \citep{kaiya2023lyfeagentsgenerativeagents}, MAICC \citep{jiang2025maicc}\\ MemoriesDB \citep{ward2025memoriesdb}}, level 3]
            ]
        ]
        [Post-Retrieval \\ (\Cref{ssec:retrievalpost}),
          level 1,
          fill=Melon!20,
          draw=Melon!70,
          edge={thick, Melon!70}
            [Re-ranking \& \\ Filtering, level 2
                [{Semantic Anchoring \citep{DBLP:journals/corr/SemanticAnchor} \\ RCR-Router \citep{DBLP:journals/corr/rcrrouter}, MemoTime \citep{Tan2025MemoTime}  \\  Zep \citep{Rasmussen2025Zep}, learn-to-memorize \citep{DBLP:journals/corr/learntomemorize} \\ Memory-R1 \citep{yan2025memory} \\ Personalized Long term Interaction \citep{DBLP:journals/corr/abs-2510-07925}\\
                RMM \citep{DBLP:conf/acl/rmm2025}, Memento \citep{DBLP:journals/corr/abs-2508-16153}}, level 3]
            ]
            [Aggregation \& \\ Compression, level 2
                [{ComoRAG \citep{DBLP:journals/corr/comorag}, MA-RAG \citep{DBLP:journals/corr/MA-RAG} \\ G-Memory \citep{Zhang2025GMemory}}, level 3]
            ]
        ]
      ]
    \end{forest}
    }
    \caption{Taxonomy of memory retrieval methodologies in agentic systems. The mindmap organizes existing literature into four distinct phases of the retrieval pipeline: \textbf{Timing and Intent}, which governs the initiation of the process; \textbf{Query Construction}, covering techniques for query decomposition and rewriting; \textbf{Retrieval Strategies}, categorizing search paradigms into lexical, semantic, graph-based, and hybrid approaches; and \textbf{Post-Retrieval Processing}, which focuses on refining outputs through re-ranking, filtering, and aggregation.
    }
    \label{fig:retrievalmindmap}
\end{figure}
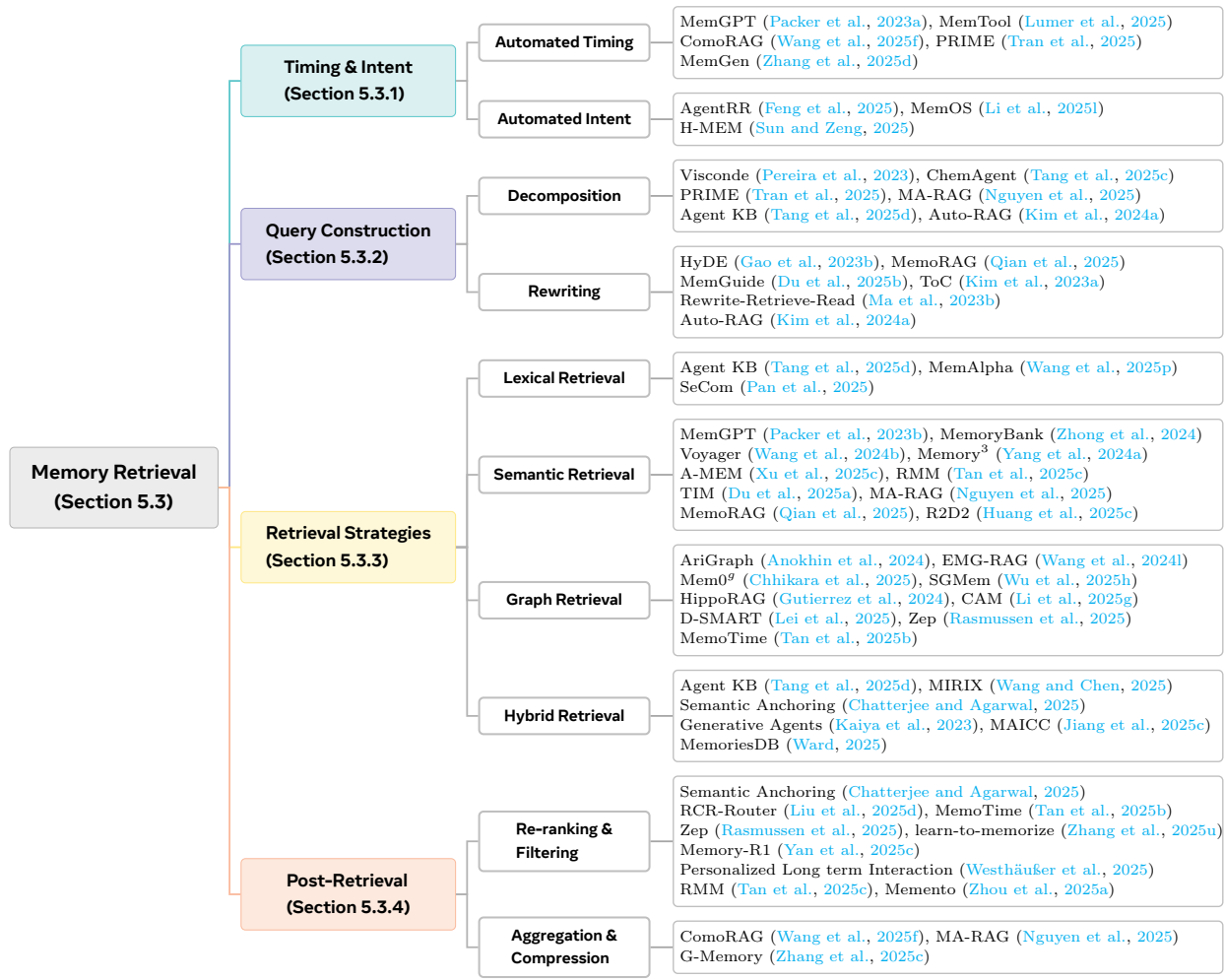

Building on the memory bank established in \Cref{ssec:formation} and \Cref{ssec:evolution}, the next critical step is how to retrieve and utilize memories during reasoning. We define memory retrieval as the process of retrieving relevant and concise knowledge fragments from a certain memory repository to support current reasoning tasks at the right moment. The key challenge lies in efficiently and accurately locating the required knowledge fragments within a large-scale memory store. To address this, many algorithms employ heuristic strategies or learnable models to optimize various stages of the retrieval process. Based on the execution order of retrieval, this process can be decomposed into four aspects.
\autoref{fig:retrievalmindmap} provides a structured overview of this retrieval pipeline, organizing existing methods according to their roles across retrieval stages.

\begin{tcolorbox}[
  colback=selfevolagent_light!20,
  colframe=selfevolagent_light!80,
  colbacktitle=selfevolagent_light!80,
  coltitle=black,
  title={\bfseries\fontfamily{ppl}\selectfont{Four Steps of Memory Retrieval}},
  boxrule=2pt,
  arc=5pt,
  drop shadow,
  parbox=false,
  before skip=5pt,
  after skip=5pt,
  left=5pt,   
  right=5pt,
]

\begin{itemize} 
\item \textbf{Retrieval Timing and Intent}(\Cref{ssec:retrievaltiming}) determines the specific moments and objectives for memory retrieval, shifting from passive, instruction-driven triggers to autonomous, self-regulated decisions. 
\item \textbf{Query Construction}(\Cref{ssec:retrievalquery}) bridges the semantic gap between the user's raw input and the stored memory index by decomposing or rewriting queries into effective retrieval signals. 
\item \textbf{Retrieval Strategies}(\Cref{ssec:retrievalstrategy}) executes the search over the memory repository, employing paradigms ranging from sparse lexical matching to dense semantic embedding and structure-aware graph traversal. 
\item \textbf{Post-Retrieval Processing}(\Cref{ssec:retrievalpost}) refines the retrieved raw fragments through re-ranking, filtering, and aggregation, ensuring that the final context provided to the model is concise and coherent. 
\end{itemize}
\end{tcolorbox}

Collectively, these mechanisms transform memory retrieval from a static search operation into a dynamic cognitive process. Retrieval timing and intent determine when and where to retrieve. Next, query construction specifies what to retrieve, and retrieval strategies focus on how to execute the retrieval. Finally, post-retrieval processing decides how the retrieved information is integrated and used. A robust agentic system typically orchestrates these components within a unified pipeline, enabling agents to approximate human-like associative memory activation for efficient knowledge access.

\subsubsection{Retrieval Timing and Intent}
\label{ssec:retrievaltiming}
The retrieval intent and timing determine when to trigger the retrieval mechanism and which memory store to query. Existing memory systems adopt different design choices in this regard, ranging from always-on retrieval to retrieval triggered by explicit instructions or internal signals~\citep{DBLP:conf/aaai/Zhao0XLLH24,DBLP:journals/corr/abs-2509-25911,fang2025lightmem}. For example, MIRIX~\citep{wang2025mirixmultiagentmemoryllmbased} performs retrieval from all six memory databases for each query and concatenates the retrieved contents, reflecting a design that prioritizes comprehensive memory access. Other approaches instead aim to trigger retrieval more selectively, allowing the model to decide both the timing and scope of memory access, which can lead to more targeted and efficient use of memory resources. In this subsection, we review the literature from two complementary perspectives: automated retrieval timing and automated retrieval intent.

\paragraph{Automated Retrieval Timing} This term refers to the model’s ability to autonomously determine when to trigger a memory retrieval operation during reasoning. The simplest strategy is to delegate the decision to either the LLM or an external controller, allowing it to determine solely from the query whether retrieval is necessary. For example, MemGPT~\citep{packerMemGPTLLMsOperating2023} and MemTool~\citep{lumer2025memtool} allow the LLM itself to invoke retrieval functions, enabling efficient access to external memory within an operating-system-like framework. However, these methods rely on static judgments from the query alone, neglecting the model’s dynamically evolving cognitive state during reasoning.

To address this limitation, recent work integrates fast–slow thinking mechanisms into retrieval timing. ComoRAG~\citep{DBLP:journals/corr/comorag} and PRIME~\citep{DBLP:journals/corr/PRIME}, for instance, first produce a fast response and then let the agent evaluate its adequacy. If the initial reasoning is deemed insufficient, the system triggers deeper retrieval and reasoning based on failure feedback. MemGen~\citep{Zhang2025MemGen} further refines the triggering mechanism by converting the explicit agent-level decision into a latent, trainable process. It introduces memory triggers that detect critical retrieval moments from latent rollout states, thereby improving the precision of retrieval timing while preserving end-to-end differentiability.

\paragraph{Automated Retrieval Intent} This aspect concerns the model’s ability to autonomously decide which memory source to access within a hierarchical storage form. AgentRR~\citep{DBLP:journals/corr/agentrr}, for example, dynamically switches between low-level procedural templates and high-level experiential abstractions based on environmental feedback. However, its reliance on explicit feedback limits applicability in open-ended reasoning settings.

To overcome this constraint, MemOS~\citep{li2025memosoperatingmemoryaugmentedgeneration} employs a MemScheduler that dynamically selects among parametric, activation, and plaintext memory based on user-, task-, or organization-level context. Yet, this flat selection scheme overlooks the hierarchical structure of the memory system. H-MEM~\citep{Sun2025HMEM} addresses this by introducing an index-based routing mechanism, which performs coarse-to-fine retrieval, moving from the domain layer to the episode layer and gradually narrowing the search space to the most relevant sub-memories. This hierarchical routing not only improves retrieval precision but also mitigates information overload.

\paragraph{Summary} Autonomous timing and intent help reduce computational overhead and suppress unnecessary noise, but they also create a potential vulnerability. When an agent overestimates its internal knowledge and fails to initiate retrieval when needed, the system can fall into a silent failure mode in which knowledge gaps may lead to hallucinated outputs. Therefore, a balance needs to be achieved: providing the agent with essential information at the right moments while avoiding excessive retrieval that introduces noise.

\subsubsection{Query Construction}
\label{ssec:retrievalquery}
After initiating the retrieval process, the next challenge lies in transforming the raw query into an effective retrieval signal aligned with the memory index. Query construction acts as the translation layer between the user's surface utterance and the memory's latent storage. Traditional approaches typically perform retrieval directly based on the user query, which is simple but fails to align the query semantics with those of the memory index. To bridge this gap, agentic memory systems proactively perform query decomposition or query rewriting, generating intermediate retrieval signals that better match the latent structure of the memory.

\paragraph{Query Decomposition}
This approach breaks down a complex query into simpler sub-queries, allowing the system to retrieve more fine-grained and relevant information. Such decomposition alleviates the one-shot retrieval bottleneck by enabling modular retrieval and reasoning over intermediate results. For instance, Visconde~\citep{DBLP:conf/ecir/visconde} and ChemAgent~\citep{tang2025chemagentselfupdatinglibrarylarge} employ LLMs to decompose the original question into sub-problems, retrieve candidate results for each from the memory, and finally aggregate them into a coherent answer. However, these methods lack global planning. To address this issue, PRIME~\citep{DBLP:journals/corr/PRIME} and MA-RAG~\citep{DBLP:journals/corr/MA-RAG} introduce a Planner Agent, inspired by the ReAct~\citep{yao2023react} paradigm, that first formulates a global retrieval plan before decomposing it into sub-queries. Yet, these approaches mainly rely on problem-driven decomposition and thus cannot explicitly identify what specific knowledge the model is missing. To make sub-queries more targeted, Agent KB~\citep{tang2025agentkbleveragingcrossdomain} adopts a two-stage retrieval process in which a teacher model observes the student model’s failures and generates fine-grained sub-queries accordingly. This targeted decomposition improves retrieval precision and reduces irrelevant results, particularly in knowledge-intensive tasks.

\paragraph{Query Rewriting}
Instead of decomposing, this strategy rewrites the original query or generates a hypothetical document to refine its semantics before retrieval. Such rewriting mitigates the mismatch between user intent and the memory index. HyDE~\citep{DBLP:conf/acl/HyDE23}, for example, instructs the LLM to generate a hypothetical document in a zero-shot manner and performs retrieval using its semantic embedding. The generated document encapsulates the desired semantics, effectively bridging the gap between the user query and the target memory. MemoRAG~\citep{DBLP:conf/www/Qian0ZMLD025} extends this idea by incorporating global memory into hypothetical document generation. It first compresses the global memory and then generates a draft answer conditioned on both the query and the compressed memory; this draft is then used as a rewritten query. Since the draft has access to the global memory context, it captures user intent more faithfully and uncovers implicit information needs. Similarly, MemGuide~\citep{duMemGuideIntentDrivenMemory2025} leverages the dialogue context to prompt an LLM to produce a concise, command-like phrase that serves as a high-level intent description for retrieval. Beyond directly prompting an LLM to rewrite the query, Rewrite-Retrieve-Read~\citep{ma2023queryrewrite} trains a small language model as a dedicated rewriter through reinforcement learning, while ToC~\citep{kim2023treeclarificationsansweringambiguous} employs a Tree of Clarifications to progressively refine and specify the user’s retrieval objective. 

\paragraph{Summary}
These two paradigms, decomposition and rewriting, are not mutually exclusive. Auto-RAG~\citep{DBLP:journals/corr/autorag} integrates both by evaluating HyDE and Visconde under identical retrieval conditions and then selecting the strategy that performs best for the given task. The findings of this work demonstrate that the quality of the memory-retrieval query has a substantial impact on reasoning performance. In contrast to earlier research, which primarily focused on designing sophisticated memory architectures, recent studies~\citep{yanGeneralAgenticMemory2025} place increasing emphasis on the retrieval construction process, shifting the role of memory toward serving retrieval. The choice of what to retrieve with is, unsurprisingly, a critical component of this process.

\subsubsection{Retrieval Strategies}
\label{ssec:retrievalstrategy}
After clarifying the retrieval objective, we obtain a query with a well-defined intent. The next core challenge lies in leveraging this query to efficiently and accurately retrieve truly relevant knowledge from a large and complex memory repository. Retrieval strategies serve as the bridge between queries and the memory base, and their design directly determines both retrieval efficiency and result quality. In this section, we systematically review various retrieval paradigms and analyze their strengths, limitations, and application scenarios—from traditional sparse retrieval based on keyword matching, to modern dense retrieval using semantic embeddings, to graph-based retrieval for structured knowledge, to the emerging class of generative retrieval methods, and finally to hybrid retrieval techniques that integrate multiple paradigms.

\paragraph{Lexical Retrieval} This strategy relies on keyword matching to locate relevant documents, with representative methods including TF-IDF~\citep{TFIDF1972} and BM25~\citep{DBLP:journals/ftir/RobertsonZ09BM25}. TF-IDF measures the importance of keywords based on term frequency and inverse document frequency, enabling fast and interpretable retrieval. BM25 further refines this approach by incorporating term frequency saturation and document length normalization. Such methods are often employed in precision-oriented retrieval scenarios, where accuracy and relevance of results take precedence over recall~\citep{tang2025agentkbleveragingcrossdomain,DBLP:journals/corr/abs-2509-25911,panSeCom2025}. However, purely lexical matching struggles to capture semantic variations and contextual relationships, making it highly sensitive to linguistic expression differences and thus less effective in open-domain knowledge or multimodal memory settings.

\paragraph{Semantic Retrieval} This strategy encodes queries and memory entries into a shared embedding space and matches them based on semantic similarity rather than lexical overlap. Representative approaches utilize semantic encoders, including Sentence-BERT~\citep{DBLP:conf/emnlp/SentenceBert19} and CLIP~\citep{DBLP:conf/icml/CLIP21}. Within memory systems, this approach better captures task context and supports semantic generalization and fuzzy matching, making it the default choice in most agentic memory frameworks~\citep{DBLP:conf/nips/RAG2020,wang_voyager_2024,DBLP:journals/corr/abs-2407-01178,xuAMEMAgenticMemory2025,DBLP:conf/acl/rmm2025,DBLP:journals/corr/MA-RAG,DBLP:conf/www/Qian0ZMLD025,hassell2025learning,huang_r2d2_2025}. However, semantic drift and forced top-K retrieval often introduce retrieval noise and spurious recall. To address these issues, recent systems incorporate dynamic retrieval policies, reranking modules, and hybrid retrieval schemes.

\paragraph{Graph Retrieval} This strategy leverages not only semantic signals but also the explicit topological structure of graphs, enabling inherently more precise and structure-aware retrieval. By directly accessing structural paths, these methods exhibit stronger multi-hop reasoning capabilities and can more effectively explore long-range dependencies. Moreover, treating relational structure as a constraint on inference paths naturally supports retrieval governed by exact rules and symbolic constraints. Representative approaches such as AriGraph~\citep{anokhin2024arigraph}, EMG-RAG~\citep{wang2024craftingpersonalizedagentsretrievalaugmented}, Mem0$^g$~\citep{Chhikara2025mem0}, and SGMem~\citep{Wu2025SGMemSG} first identify the most relevant nodes or triples and then expand to their semantically related K-hop neighbors to construct an ego-graph. HippoRAG~\citep{gutiérrez2025hipporagneurobiologicallyinspiredlongterm} performs personalized PageRank~\citep{Page1999ThePC} seeded on the retrieved nodes and ranks the rest of the graph by their proximity to these seeds, enabling effective multi-hop retrieval. Going beyond fixed expansion rules, CAM~\citep{liCAMConstructivistView2025} and D-SMART~\citep{Lei2025DSMART} employ LLMs to steer subgraph exploration: CAM uses an LLM to select informative neighbors and children of a central node for associative exploration, while D-SMART treats the LLM as a planner that performs beam search over a KG memory to retrieve one-hop neighbors of target entities and the relations connecting a given entity pair. For temporal graphs, Zep~\citep{Rasmussen2025Zep} and MemoTime~\citep{Tan2025MemoTime} further enable entity-subgraph construction and relation retrieval under explicit temporal constraints, ensuring that the returned results satisfy the required time rules.

\paragraph{Generative Retrieval} This strategy replaces lexical or semantic retrieval with a model that directly generates the identifiers of relevant documents~\citep{DBLP:conf/nips/Taytransformer22,DBLP:conf/nips/Wangneuralcorpus0022}. By framing retrieval as a conditional generation task, the model implicitly stores candidate documents in its parameters and performs deep query–document interaction during decoding\citep{DBLP:journals/tois/LiJZZZZD25}. Leveraging the semantic capabilities of pretrained language models, this paradigm often outperforms traditional retrieval methods, particularly in small-scale settings\citep{DBLP:conf/www/Zeng0JSWZ24}.
However, generative retrieval requires additional training to internalize the semantics of all candidate documents, resulting in limited scalability when the corpus evolves~\citep{DBLP:conf/eacl/YuanWFPLWML24}. For these reasons, agentic memory systems have paid relatively little attention to this paradigm, although its tight integration of generation and retrieval suggests untapped potential.

\paragraph{Hybrid Retrieval} This strategy integrates the strengths of multiple retrieval paradigms. Systems such as Agent KB~\citep{tang2025agentkbleveragingcrossdomain} and MIRIX~\citep{wang2025mirixmultiagentmemoryllmbased} combine lexical and semantic retrieval to balance precise term or tool matching with broader semantic alignment. Similarly, Semantic Anchoring~\citep{DBLP:journals/corr/SemanticAnchor} performs parallel searches over semantic embeddings and symbolic inverted indices to achieve complementary coverage. Some other methods combine multiple evaluation signals to guide retrieval. Generative Agents~\citep{kaiya2023lyfeagentsgenerativeagents}, for example, illustrate this multi-factor approach through a scoring scheme that accumulates recency, importance, and relevance. MAICC~\citep{jiang2025maicc} adopts a mixed-utility scoring function that integrates similarity with both global and predicted individual returns. In graph-based settings, retrieval typically proceeds in two stages: semantic retrieval first identifies relevant nodes or triples, and graph topology is subsequently leveraged to expand the search space~\citep{anokhin2024arigraph,wang2024craftingpersonalizedagentsretrievalaugmented,gutiérrez2025hipporagneurobiologicallyinspiredlongterm,liCAMConstructivistView2025}. 

At the database infrastructure level, MemoriesDB~\citep{ward2025memoriesdb} introduces a temporal–semantic–relational database designed for long-term agent memory, providing a hybrid retrieval architecture that integrates these dimensions into a unified storage and access framework. 

By fusing heterogeneous retrieval signals, hybrid approaches preserve the precision of keyword matching while incorporating the contextual understanding of semantic methods, ultimately yielding more comprehensive and relevant results.

\subsubsection{Post-Retrieval Processing}
\label{ssec:retrievalpost}
Initial retrieval often returns fragments that are redundant, noisy, or semantically inconsistent. Directly injecting these results into the prompt can lead to excessively long contexts, conflicting information, and reasoning distracted by irrelevant content. Post-retrieval processing, therefore, becomes essential for ensuring prompt quality. Its goal is to distill the retrieved results into a concise, accurate, and semantically coherent context. In practice, two components are central:
(1) \textbf{Re-ranking and Filtering:} performing fine-grained relevance estimation to remove irrelevant or outdated memories and reorder the remaining fragments, thereby reducing noise and redundancy.
(2) \textbf{Aggregation and Compression:} integrating the retrieved memories with the original query, eliminating duplication, merging semantically similar information, and reconstructing a compact and coherent final context.

\paragraph{Re-ranking and Filtering} To maintain a concise and coherent context, initial retrieval results are re-ranked and filtered to ensure a concise and coherent context by removing low-relevance items. Early approaches rely on heuristic criteria for evaluating semantic consistency. For example, Semantic Anchoring~\citep{DBLP:journals/corr/SemanticAnchor} integrates vector similarity with entity- and discourse-level alignment, whereas RCR-Router~\citep{DBLP:journals/corr/rcrrouter} combines multiple handcrafted signals, including role relevance, task-stage priority, and recency. These methods, however, often require extensive hyperparameter tuning to balance heterogeneous importance scores. To alleviate this burden, learn-to-memorize~\citep{DBLP:journals/corr/learntomemorize} formulates score aggregation as a reinforcement-learning problem, enabling the model to learn optimal weights over retrieval signals. While these techniques primarily optimize semantic coherence, scenarios demanding strict temporal reasoning require additional constraints: \citet{Rasmussen2025Zep} and \citet{Tan2025MemoTime} filter memories based on their timestamps and validity windows to satisfy complex temporal dependencies.

With the increasing capability of LLMs, recent methods leverage their intrinsic language understanding to assess memory quality directly. Memory-R1~\citep{yan2025memory} and \citet{DBLP:journals/corr/abs-2510-07925} both introduce LLM-based evaluators (Answer Agents or Self-Validator Agents) that filter retrieved content before producing the final response. However, prompt-based filtering remains limited by the LLM’s inherent capacity and by mismatches between prompt semantics and downstream usage. Consequently, many systems train auxiliary models to estimate memory importance more robustly~\citep{DBLP:conf/acl/rmm2025}. Memento~\citep{DBLP:journals/corr/abs-2508-16153} uses Q-learning~\citep{Watkins1992Qlearning} to predict the probability that a retrieved item contributes to a correct answer, and MemGuide~\citep{duMemGuideIntentDrivenMemory2025} fine-tunes LLaMA-8B~\citep{dubey2024llama} to re-rank candidates using marginal slot-completion gain. Together, these re-ranking and filtering strategies refine retrieval results without modifying the underlying retriever, enabling compatibility with any pre-trained retrieval model while supporting task-specific optimization.

\paragraph{Aggregation and Compression} Another approach to improving both the quality and efficiency of downstream reasoning through post-retrieval processing is the aggregation and compression. This process integrates the retrieved evidence with the query to form a coherent and compact context. Unlike filtering and re-ranking, which mainly address noise and prioritization, this stage focuses on merging multiple fragmented memory items into higher-level and distilled knowledge representations, and on refining these representations when task-specific adaptations are required. ComoRAG~\citep{DBLP:journals/corr/comorag} illustrates this idea through its Integration Agent, which identifies historical signals that are semantically aligned with the query and combines them into an abstract global summary that provides broad contextual grounding. The Extractor Agent in MA-RAG~\citep{DBLP:journals/corr/MA-RAG} performs fine-grained content selection over the retrieved documents, retaining only the key information that is strongly relevant to the current subquery and producing concise snippets tailored to local reasoning needs.

Furthermore, G-Memory~\citep{Zhang2025GMemory} extends aggregation and compression into the personalization for multi-agent systems. It consolidates retrieved high-level insights and sparsified trajectories, and then uses an LLM to customize these condensed experiences according to the agent’s role. This process refines general knowledge into role-specific prompts that populate the agent’s personalized memory.

\paragraph{Summary} 
In conclusion, post-retrieval processing acts as a crucial intermediate step that transforms noisy, fragmented retrieval results into a precise and coherent context for reasoning. Through the above mechanisms, the post-retrieval processing not only enhances the density and fidelity of the memories supplied to the model but also aligns the information with task requirements and agent characteristics.

\begin{table*}[!t]
\caption{
Overview of benchmarks relevant to LLM agent memory, long-term, lifelong learning, and self-evolving evaluation.
The table covers two categories of benchmarks: (i) benchmarks explicitly designed for memory-, lifelong learning-, or self-evolving agent evaluation, and (ii) other agent-oriented benchmarks that implicitly stress long-horizon memory through sequential, multi-step, or multi-task interactions.
\textbf{Fac.} and \textbf{Exp.} indicate whether a benchmark evaluates factual memory or experiential (interaction-derived) memory, respectively. \textbf{MM.} denotes the presence of multimodal inputs, while \textbf{Env.} indicates whether the benchmark is conducted in a simulated or real environment. \textbf{Feature} summarizes the primary capability under evaluation, and \textbf{Scale} reports the approximate benchmark size in terms of \textit{samples} (s.) or \textit{tasks} (t.). PDDL denotes commonly used PDDL-based planning subsets.
}
\centering
\label{tab:benchmark}
\small
\begin{tabular}{l l c c c c l l}
\hline
\textbf{Name} & \textbf{Link} & \textbf{Fac.} & \textbf{Exp.} & \textbf{MM.} & \textbf{Env.} & \textbf{Feature} & \textbf{Scale} \\
\hline
\multicolumn{8}{c}{\textbf{Memory/Lifelong-learning/Self-evolving-oriented Benchmarks}}\\

MemBench &
\ghlink{https://github.com/import-myself/Membench} &
\cmark & \cmark & \xmark & simulated &
interactive scenarios &
53{,}000 s.\\%

MemoryAgentBench &
\ghlink{https://github.com/HUST-AI-HYZ/MemoryAgentBench} &
\cmark & \cmark & \xmark & simulated &
multi-turn interactions & 
4 t. \\%

LoCoMo &
\weblink{https://snap-research.github.io/locomo/} &
\cmark & \xmark & \cmark & real &
conversational memory &
300 s. \\%

WebChoreArena &
\ghlink{github.com/web-arena/WebChoreArena} &
\cmark & \cmark & \cmark & real &
tedious web browsing &
4 t./532 s. \\%

MT-Mind2Web &
\ghlink{https://github.com/magicgh/self-map} &
\cmark & \cmark & \xmark & real &
conversational web navigation &
720 s. \\

PersonaMem &
\weblink{https://arxiv.org/abs/2504.14225} &
\cmark & \xmark & \xmark & simulated &
dynamic user profiling & 
15 t./180 s. \\%

LongMemEval &
\ghlink{https://github.com/xiaowu0162/LongMemEval} &
\cmark & \xmark & \xmark & simulated &
interactive memory &
5 t./500 s. \\%

PerLTQA &
\weblink{https://arxiv.org/abs/2402.16288} &
\cmark & \xmark & \xmark & simulated &
social personalized interactions &
8{,}593 s. \\%

MemoryBank &
\weblink{https://arxiv.org/abs/2305.10250} &
\cmark & \xmark & \xmark & simulated &
user memory updating &
194 s. \\%

MPR &
\ghlink{https://github.com/nuster1128/MPR} &
\cmark & \xmark & \xmark & simulated &
user personalization &
108{,}000 s. \\%

PrefEval & 
\weblink{https://prefeval.github.io} &
\cmark & \xmark & \xmark & simulated &
personal preferences & 
3{,}000 s.\\%

LOCCO &
\weblink{https://aclanthology.org/2025.findings-acl.1014/} &
\cmark & \xmark & \xmark & simulated &
 chronological conversations & 
 3{,}080 s. \\%

StoryBench &
\weblink{https://arxiv.org/abs/2506.13356} &
\cmark & \cmark & \xmark & mixed &
interactive fiction games &
3 t. \\%

MemoryBench &
\weblink{https://arxiv.org/abs/2510.17281} &
\cmark & \cmark & \xmark & simulated &
continual learning &
4 t./$\sim$ 20{,}000 s. \\%

Madial-Bench &
\ghlink{https://github.com/hejunqing/MADial-Bench} &
\cmark & \xmark & \xmark & simulated &
memory recalling &
331 s. \\%

Evo-Memory &
\weblink{https://arxiv.org/abs/2511.20857} &
\cmark & \cmark & \xmark & simulated &
test-time learning & 10 t./$\sim$ 3{,}700 s.
 \\%

LifelongAgentBench &
\weblink{https://arxiv.org/abs/2505.11942} &
\cmark & \cmark & \xmark & simulated &
lifelong learning &
1{,}396 s. \\%

StreamBench &
\weblink{https://stream-bench.github.io} &
\cmark & \cmark & \xmark & simulated &
continuous online learning &
9{,}702 s. \\%

DialSim &
\weblink{https://arxiv.org/abs/2406.13144} &
\cmark & \cmark & \xmark & real &
multi-dialogue understanding &
$\sim$ 1{,}300 s. \\%

LongBench &
\weblink{https://arxiv.org/abs/2308.14508} &
\cmark & \xmark & \xmark & mixed &
long-context understanding &
21 t./4{,}750 s. \\%

LongBench v2 &
\weblink{https://longbench2.github.io/} &
\cmark & \xmark & \xmark & mixed &
long-context multitasks &
20 t./503 s. \\%

RULER &
\ghlink{https://github.com/NVIDIA/RULER} &
\cmark & \xmark & \xmark & simulated &
long-context retrieval &
13 t. \\%

BABILong &
\ghlink{https://github.com/booydar/babilong} &
\cmark & \xmark & \xmark & simulated &
long-context reasoning & 20 t.
 \\%

MM-Needle &
\weblink{https://mmneedle.github.io/} &
\cmark & \xmark & \cmark & simulated &
multimodal long-context retrieval & $\sim$ 280{,}000 s.
 \\%

HaluMem & 
\ghlink{https://github.com/MemTensor/HaluMem} &
\cmark & \xmark & \xmark & simulated &
memory hallucinations & 
3{,}467 s. \\%

HotpotQA &  \weblink{https://hotpotqa.github.io/} & \cmark & \xmark & \xmark & simulated & long-context QA & 113k s. \\

\hline
\multicolumn{8}{c}{\textbf{Other Related Benchmarks}}\\

ALFWorld & 
\weblink{https://alfworld.github.io/} & 
\cmark & \cmark & \xmark & simulated & 
text-based embodied environment & 
3{,}353 t. \\%

ScienceWorld & 
\ghlink{http://github.com/allenai/ScienceWorld} & 
\cmark & \cmark & \xmark & simulated & 
interactive embodied environment & 
10 t./30 t. \\%

AgentGym &
\weblink{https://agentgym.github.io} &
\xmark & \cmark & \xmark & mixed &
multiple environments  &
89 t./20{,}509 s. \\%

AgentBoard &
\ghlink{https://github.com/hkust-nlp/AgentBoard} &
\xmark & \cmark & \xmark & mixed &
multi-round interaction &
9 t./1013 s.\\%

PDDL$^*$ & \weblink{https://hkust-nlp.github.io/agentboard/static/leaderboard.html} & \xmark & \cmark & \xmark & simulated & strategy game & -  \\%

BabyAI &
\weblink{https://arxiv.org/abs/1810.08272} &
\xmark & \cmark & \xmark & simulated &
language learning &
19 t. \\%

WebShop &
\weblink{https://arxiv.org/pdf/2207.01206} &
\xmark & \cmark & \cmark & simulated &
e-commerce web interaction  &
12{,}087 s. \\%

WebArena &
\weblink{https://webarena.dev/} &
\xmark & \cmark & \cmark & real &
web interaction &
812 s. \\%

MMInA &
\weblink{https://mmina.cliangyu.com/} &
\cmark & \cmark & \cmark & real &
multihop web interaction  &
1{,}050 s. \\%

SWE-Bench Verified &
\weblink{https://www.swebench.com/} &
\xmark & \cmark & \xmark & real &
code repair &
500 s. \\%

GAIA &
\weblink{https://arxiv.org/abs/2311.12983} &
\xmark & \cmark & \cmark & real &
human-level deep research &
466 s. \\%

xBench-DS & \weblink{https://xbench.org/} & \xmark & \cmark & \cmark & real & deep-search evaluation & 100 s. 
\\%

ToolBench &
\ghlink{https://github.com/OpenBMB/ToolBench} &
\xmark & \cmark & \xmark & real &
API tool use &
126{,}486 s. \\%

GenAI-Bench &
\weblink{https://arxiv.org/abs/2406.13743} &
\xmark & \cmark & \cmark & real &
visual generation evaluation &
$\sim$ 40{,}000 s. \\%
\hline
\end{tabular}
\end{table*}

\section{Resources and Frameworks}
\label{sec:resource}
\subsection{Benchmarks and Datasets}
\label{sec:benchmarks}

In this section, we survey representative benchmarks and datasets that have been used (or could be used) to evaluate the memory, long-term, continual-learning, or long-context capabilities of LLM-based agents. We classify these benchmarks into two broad categories: (1) those explicitly designed for memory / lifelong learning / self-evolving agents, and (2) those originally developed for other purposes (e.g., tool-use capacity, web search, embodied action) but nevertheless relevant for memory evaluation due to their long-horizon, multi-task, or sequential nature.

\subsubsection{Benchmarks for Memory / Lifelong / Self-Evolving Agents}
Memory-oriented benchmarks focus primarily on how well an agent can construct, maintain, and exploit an explicit memory of past interactions or world facts. These tasks typically probe the retention and retrieval of information across multi-turn dialogues, user-specific sessions, or long synthetic narratives, sometimes including multimodal signals. 

A consolidated overview of these benchmarks, including their memory focus, environment type, modality, and evaluation scale, is provided in \autoref{tab:benchmark}, which serves as a structured reference for comparing their design objectives and evaluation settings.
Representative examples such as MemBench~\citep{tan2025membench}, LoCoMo~\citep{maharana2024evaluating}, WebChoreArena~\citep{miyai2025webchorearena}, MT-Mind2Web~\citep{deng2024multi}, PersonaMem~\citep{jiang2025know}, PerLTQA~\citep{du2024perltqa}, MPR~\citep{zhang2025explicit}, PrefEval~\citep{DBLP:conf/iclr/Zhao00HL25}, LOCCO~\citep{jia2025evaluating}, StoryBench~\citep{wan2025storybench}, Madial-Bench~\citep{he2025madial}, DialSim~\citep{zheng2025lifelongagentbench}, LongBench~\citep{bai2024longbench}, LongBench v2~\citep{bai2025longbench}, RULER~\citep{hsieh2024ruler}, BALILong~\citep{DBLP:conf/nips/KuratovBARSS024} MM-Needle~\citep{wang2025multimodal}, and HaluMem~\citep{chen2025halumem} stress user modeling, preference tracking, and conversation-level consistency, often under simulated settings where ground-truth memories can be precisely controlled.

Lifelong-learning benchmarks extend beyond isolated memory retrieval to examine how agents continually acquire, consolidate, and update knowledge over long horizons and evolving task distributions. Benchmarks such as LongMemEval~\citep{DBLP:conf/iclr/WuWYZCY25}, MemoryBank~\citep{zhong2023memorybankenhancinglargelanguage}, MemoryBench~\citep{ai2025memorybench}, LifelongAgentBench~\citep{zheng2025lifelongagentbench}, and StreamBench~\citep{DBLP:conf/nips/WuTLCL24},  are designed around sequences of tasks or episodes in which new information gradually arrives and earlier information may become obsolete or conflicting. These setups emphasize phenomena like catastrophic forgetting, forward and backward transfer, and test-time adaptation, making them suitable for studying how memory mechanisms  interact with continual-learning objectives. In many cases, performance is tracked not only on the current task but also on previously seen tasks or conversations, thereby quantifying how well the agent preserves useful knowledge while adapting to new users, domains, or interaction patterns.

Self-evolving-agent benchmarks go a step further by treating the agent as an open-ended system that can iteratively refine its own memory, skills, and strategies through interaction. Here, the focus is not only on storing and recalling information, but also on meta-level behaviors such as self-reflection, memory editing, tool-augmented storage, and policy improvement over multiple episodes or games. Benchmarks like   MemoryAgentBench~\citep{hu2025evaluating}, Evo-Memory~\citep{wei2025evo}, and other multi-episode or mission-style environments can be instantiated in a self-evolving setting by allowing the agent to accumulate trajectories, synthesize higher-level abstractions, and adjust its behavior in future runs based on its own past performance. When viewed through this lens, these benchmarks provide a testbed for evaluating whether an agent can autonomously bootstrap more capable behaviors over time-turning static tasks into arenas for long-term adaptation, strategy refinement, and genuinely self-improving memory use.

\subsubsection{Other Related Benchmarks} 
\label{sec:other_benchmarks}

Beyond benchmarks explicitly designed for memory or lifelong learning, a wide range of agent-oriented and long-horizon evaluation suites are also relevant for studying memory-related capabilities in LLM-based agents. Although these benchmarks were originally introduced to assess other aspects such as tool use, embodied interaction, or knowledge-intensive reasoning, their sequential, multi-step, and multi-task nature implicitly places strong demands on long-term information retention, context management, and state tracking.

Embodied and interactive environments constitute a major class of such benchmarks. Frameworks like ALFWorld~\citep{shridhar2021alfworld} and ScienceWorld~\citep{wang2022scienceworldagentsmarter5th} evaluate agents in simulated text-based or partially grounded environments where success requires remembering past observations, intermediate goals, and environment dynamics across extended action sequences. Similarly, BabyAI~\citep{chevalierboisvert2019babyaiplatformstudysample} focuses on language-conditioned instruction following over temporally extended episodes, implicitly testing an agent’s ability to maintain task-relevant state throughout interaction. While these benchmarks do not explicitly model external memory modules, effective performance often depends on the agent’s capacity to preserve and reuse information over long horizons.

Another prominent category includes web-based and tool-augmented interaction benchmarks. WebShop~\citep{yao2023webshopscalablerealworldweb}, WebArena~\citep{zhou2024webarenarealisticwebenvironment}, and MMInA~\citep{tian2025mminabenchmarkingmultihopmultimodal} assess agents operating in realistic or semi-realistic web environments involving multi-step navigation, information gathering, and decision making. These settings naturally induce long-context trajectories in which earlier actions, retrieved information, or user constraints must be recalled and integrated at later stages. ToolBench~\citep{qin2024toolllm} further extends this paradigm by evaluating an agent’s ability to select and invoke APIs across complex workflows, where memory of prior tool outputs and tool-use experience is critical for coherent execution.

Multi-task and general agent evaluation platforms also provide indirect but valuable signals about memory usage. AgentGym~\citep{xi2024agentgymevolvinglargelanguage} and AgentBoard~\citep{xi2024agentgymevolvinglargelanguage} aggregate diverse environments or tasks into unified evaluation suites, requiring agents to adapt across tasks while retaining task-specific knowledge and strategies. PDDL-based planning environments, commonly used in agent benchmarks, evaluate strategic reasoning over structured action spaces, where agents benefit from accumulating and reusing experience across episodes to improve long-horizon planning performance.

Finally, several recent benchmarks target demanding real-world or near-real-world reasoning scenarios that inherently stress long-context and cross-step consistency. SWE-Bench Verified~\citep{jimenez2023swe} evaluates code repair over realistic software repositories, where agents must reason over long files and evolving code states. GAIA~\citep{mialon2023gaia} and xBench~\citep{chen2025xbenchtrackingagentsproductivity} assess deep research and search-intensive tasks that require synthesizing information gathered across multiple steps and sources. GenAI-Bench~\citep{li2024genaibenchevaluatingimprovingcompositional}, while focusing on multimodal generation quality, similarly involves complex workflows in which memory of prior prompts, intermediate outputs, or visual constraints plays a nontrivial role.

Taken together, these benchmarks complement memory-oriented evaluations explicitly by situating LLM-based agents in rich, interactive, and long-horizon settings. Although memory is not always an explicit target of measurement, sustained performance in these environments implicitly depends on an agent’s ability to manage long contexts, preserve relevant information, and integrate past experience into ongoing decision making, making them valuable testbeds for studying memory-related behaviors in practice.

\begin{table}[!t]
\caption{
Overview of representative open-source memory frameworks for LLM-based agents.
The table compares widely used frameworks in terms of the types of memory they support (factual vs. experiential), multimodality, internal memory structure, and reported evaluation benchmarks. \textbf{Fac.} and \textbf{Exp.} denote factual and experiential memory, respectively, \textbf{MM.} indicates multimodal memory support, and \textbf{Structure} summarizes the core memory abstraction or organization mechanism adopted by each framework. \textbf{Evaluation} lists publicly reported benchmarks used to assess memory-related capabilities, when available.
}
\centering
\label{tab:open_source_framework}
\small
\begin{tabular}{p{2.6cm} p{3.cm} c c c p{4.cm} p{3.cm}}
\hline
\textbf{Framework} & \textbf{Links} & \textbf{Fac.} & \textbf{Exp.} & \textbf{MM.} & \textbf{Structure} & \textbf{Evaluation} \\
\hline

MemGPT &
\ghlink{https://github.com/cpacker/MemGPT} \weblink{https://docs.letta.com/} &
\cmark & \cmark & \xmark &
hierachical (S/LTM) &
LoCoMo \\%

Mem0 &
\ghlink{https://github.com/mem0ai/mem0} \weblink{https://mem0.ai/} &
\cmark & \cmark & \xmark &
graph + vector &
LoCoMo \\%

Memobase &
\ghlink{https://github.com/memodb-io/memobase} \weblink{https://www.memobase.io/} &
\cmark & \cmark & \xmark &
structured profiles &
LoCoMo \\%

MIRIX &
\ghlink{https://github.com/Mirix-AI/MIRIX} \weblink{https://mirix.io/} &  \cmark & \cmark & \cmark &
structured memory & 
LoCoMo, MemoryAgentBench \\%

MemoryOS &
\ghlink{https://github.com/BAI-LAB/MemoryOS} \weblink{https://baijia.online/memoryos/} &
\cmark & \cmark & \xmark &
hierarchical (S/M/LTM) & 
LoCoMo, MemoryBank \\%

MemOS &
\ghlink{https://github.com/MemTensor/MemOS} \weblink{https://memos.openmem.net/} &
\cmark & \cmark & \xmark &
tree memory + memcube &
LoCoMo, PreFEval, LongMemEval, PersonaMem \\%

Zep &
\ghlink{https://github.com/getzep/zep} \weblink{https://www.getzep.com/} &
\cmark & \cmark & \xmark &
temporal knowledge graph &
LongMemEval \\%

LangMem &
\ghlink{https://github.com/langchain-ai/langmem} \weblink{https://langchain-ai.github.io/langmem/} &
\cmark & \cmark & \xmark &
core API + manager & 
- \\%

SuperMemory &
\ghlink{https://github.com/supermemoryai/supermemory} \weblink{https://supermemory.ai/} &
\cmark & \cmark & \cmark &
vector + semantic &
- \\%

Cognee &
\ghlink{https://github.com/topoteretes/cognee} \weblink{https://www.cognee.ai/} &
\cmark & \cmark & \cmark &
knowledge graph & 
- \\%

Memary &
\ghlink{https://github.com/kingjulio8238/Memary} \weblink{https://kingjulio8238.github.io/memarydocs/} &
\cmark & \cmark & \xmark &
stream + entity store &
- \\%

Pinecone &
\ghlink{https://github.com/pinecone-io/assistant-mcp} \weblink{https://docs.pinecone.io/guides/assistant/overview} &
\cmark & \xmark & \xmark &
vector database &
- \\%

Chroma &
\ghlink{https://github.com/chroma-core/chroma} \weblink{https://www.trychroma.com/} &
\cmark & \xmark & \cmark &
vector database &
- \\%

Weaviate &
\ghlink{https://github.com/weaviate/weaviate} \weblink{https://weaviate.io/} &
\cmark & \xmark & \cmark &
vector + graph &
- \\%

Second Me & 
\ghlink{https://github.com/mindverse/Second-Me} \weblink{https://home.second.me/}& 
\cmark & \xmark & \xmark &
agent ego & 
- \\%

MemU & 
\ghlink{https://github.com/NevaMind-AI/memU} \weblink{hhttps://memu.pro/} & 
\cmark & \cmark & \cmark & 
hierachical layers &
- \\%

MemEngine & 
\ghlink{https://github.com/nuster1128/MemEngine} & 
\cmark & \cmark & \cmark & 
modular space & 
- \\%

Memori & 
\ghlink{https://github.com/MemoriLabs/Memori} \weblink{https://memori.ai/}
& \cmark & \cmark & \xmark & 
memory database & 
- \\%
 
ReMe & 
\ghlink{https://github.com/agentscope-ai/ReMe} \weblink{https://www.memoryscope.ai/} & 
\cmark & \cmark & \xmark &
memory management &
- \\%

AgentMemory & 
\ghlink{https://github.com/elizaOS/agentmemory} \weblink{https://www.getzep.com/product/agent-memory/} &
\cmark & \cmark & \xmark &
memory management &
- \\%

MineContext &
\ghlink{https://github.com/volcengine/MineContext} \weblink{https://app.daily.dev/posts/minecontext-is-your-proactive-context-aware-ai-partner-b0fvvtds3} &
\cmark & \cmark & \cmark &
context engineering &
- \\%

Acontext & \ghlink{https://github.com/memodb-io/Acontext} & \cmark & \cmark & \cmark & context engineering + skill learning & - \\

PowerMem & \ghlink{https://github.com/oceanbase/powermem} & \cmark & \xmark & \cmark & oceanbase & -\\

ReMe & \ghlink{https://github.com/agentscope-ai/ReMe} & \cmark & \cmark & \xmark & agentscope & BFCL, AppWorld\\

HindSight & \ghlink{https://github.com/vectorize-io/hindsight} & \cmark & \cmark & \xmark & parallel retrieval + reflection & -\\

\hline
\end{tabular}
\end{table}

\subsection{Open-Source Frameworks}

A rapidly growing ecosystem of open-source memory frameworks aims to provide reusable infrastructure for building memory-augmented LLM agents. 
A structured comparison of representative open-source memory frameworks, including their supported memory types, architectural abstractions, and evaluation coverage, is summarized in \autoref{tab:open_source_framework}.
Most of these frameworks support factual memory via vector or structured stores, and an increasing subset also models experiential traces, such as dialogue histories, user actions, and episodic summaries, with multimodal memory emerging more recently. Open-source memory frameworks for LLM agents span a spectrum from agent-centric systems with rich, hierarchical memory abstractions to more general-purpose retrieval or memory-as-a-service backends, e.g., MemGPT~\citep{packer2023memgpt}, Mem0~\citep{Chhikara2025mem0}, Memobase, MemoryOS~\citep{kang2025memoryosaiagent}, MemOS~\citep{li2025memosoperatingmemoryaugmentedgeneration}, Zep~\citep{Rasmussen2025Zep}, LangMem~\citep{githubGitHubLangchainailangmem}, SuperMemory~\citep{supermemorySupermemoryUniversal}, Cognee~\citep{githubGitHubTopoteretescognee}, Memary~\citep{githubGitHubKingjulio8238Memary}, Pinecone, Chroma, Weaviate, Second Me, MemU, MemEngine~\citep{zhang2025memengineunifiedmodularlibrary}, Memori, ReMe~\citep{githubGitHubAgentscopeaiReMe}, AgentMemory, and MineContext~\citep{githubGitHubVolcengineMineContext}. Many of them explicitly separate short- and long-term stores and offer graph-based, profile-based, or modular memory spaces, and some have begun to report results on memory-based benchmarks. The others typically provide scalable vector or graph databases, APIs, and semantic or streaming entity layers that help organize context but often leave agent behavior and evaluation protocols to the application. Overall, these frameworks are rapidly maturing in their representational flexibility and system design.

\section{Positions and Frontiers}
\label{sec:frontier}

This section articulates key positions and emerging frontiers in the design of memory systems for LLM-based agents. Moving beyond descriptive surveys of existing methods, we focus on paradigm-level shifts that redefine how memory is constructed, managed, and optimized in long-horizon agentic settings. Specifically, we examine the transition from retrieval-centric to generative memory, from manually engineered to autonomously managed memory systems, and from heuristic pipelines to reinforcement learning–driven memory control. We further discuss how these shifts intersect with multimodal reasoning, multi-agent collaboration, and trustworthiness, outlining open challenges and research directions that are likely to shape the next generation of agent memory architectures.

\subsection{Memory Retrieval vs.\ Memory Generation}

\subsubsection{Look Back: From Memory Retrieval to Memory Generation}

Historically, the dominant paradigm in agent memory research has centered on \textbf{memory retrieval}. Under this paradigm, the primary objective is to identify, filter, and select the most relevant memory entries from an existing memory store given the current context. A large body of prior work focuses on improving retrieval accuracy through better indexing strategies, similarity metrics, reranking models, or structured representations such as knowledge graphs~\citep{DBLP:conf/acl/rmm2025,githubGitHubMemodbiomemobase}. In practice, this includes techniques such as vector similarity search with dense embeddings, hybrid retrieval combining lexical and semantic signals, hierarchical filtering, and graph-based traversal. These methods emphasize precision and recall in accessing stored information, implicitly assuming that the memory base itself is already well formed.

Recently, however, increasing attention has shifted toward \textbf{memory generation}. Rather than treating memory as a static repository to be queried, memory generation emphasizes the agent’s ability to actively synthesize new memory representations on demand. The goal is not merely to retrieve and concatenate existing fragments, but to integrate, compress, and reorganize information in a manner that is tailored to the current context and future utility. This shift reflects a growing recognition that effective memory usage often requires abstraction and recomposition, especially when raw stored information is noisy, redundant, or misaligned with the immediate task.

Existing approaches to memory generation can be broadly grouped into two directions. One line of work adopts a \textbf{retrieve then generate} strategy, where retrieved memory items serve as raw material for reconstruction. In this setting, the agent first accesses a subset of relevant memories and then generates a refined memory representation that is more concise, coherent, and context specific, as implemented in ComoRAG~\citep{DBLP:journals/corr/comorag}, G-Memory~\citep{Zhang2025GMemory} and CoMEM~\citep{Wu2025CoMEM}. This approach preserves grounding in historical information while enabling adaptive summarization and restructuring. A second line of work explores \textbf{direct memory generation}, in which memory is produced without any explicit retrieval step. Instead, the agent generates memory representations directly from the current context, interaction history, or latent internal states. Systems such as MemGen~\citep{Zhang2025MemGen} and VisMem~\citep{yu2025vismemlatentvisionmemory} exemplify this direction by constructing latent memory tokens that are customized to the task at hand, bypassing explicit memory lookup altogether.

\subsubsection{Future Perspective}

Looking ahead, we anticipate that generative approaches will play an increasingly central role in agent memory systems. We highlight three properties that future generative memory mechanisms should ideally exhibit.

First, generative memory should be \textbf{context adaptive}. Rather than storing generic summaries, the memory system should generate representations that are explicitly optimized for the agent’s anticipated future needs. This includes adapting the granularity, abstraction level, and semantic focus of memory to different tasks, stages of problem solving, or interaction regimes.

Second, generative memory should support \textbf{integration across heterogeneous signals}. Agents increasingly operate over diverse modalities and information sources, including text, code, tool outputs, and environmental feedback. Memory generation provides a natural mechanism for fusing these fragmented signals into unified representations that are more useful for downstream reasoning than raw concatenation or retrieval alone. We hypothesize that latent memory (as discussed in \Cref{ssec:latent}) might be a promising technical path for this gaol.

Third, generative memory should be \textbf{learned and self optimizing}. Rather than relying on manually specified generation rules, future systems should learn when and how to generate memory through optimization signals, such as reinforcement learning or long horizon task performance. In this view, memory generation becomes an integral component of the agent’s policy, co evolving with reasoning and decision making.

\subsection{Automated Memory Management}

\subsubsection{Look-Back: From Hand-crafted to Automatically Constructed Memory Systems.} 

Existing agent memory systems ~\citep{xuAMEMAgenticMemory2025,packerMemGPTLLMsOperating2023} typically rely on manually designed strategies to determine what information to store, when to use it, and how to update or retrieve it. By guiding fixed LLMs with detailed instructions ~\citep{Chhikara2025mem0}, predefined thresholds ~\citep{kang2025memoryosaiagent}, or explicit human-crafted rules drafted by human experts ~\citep{xuAMEMAgenticMemory2025}, system designers can integrate memory modules into current agent frameworks with relatively low computational and engineering cost, enabling rapid prototyping and deployment. Besides, they also offer \textbf{interpretability, reproducibility, and controlled}, allowing the developers to precisely specify the state and behavior of memory. However, similar to expert systems in other areas, such manually curated approaches suffer from significant limitations: they are inherently inflexible and often fail to generalize across diverse, dynamic environments. Consequently, these systems tend to underperform in long-term or open-ended interactions.

Recent developments in agent memory research begin to address these limitations by enabling the agents themselves to autonomously manage the memory evolution and retrieval. For example, CAM ~\citep{liCAMConstructivistView2025} empowers LLM agents to automatically cluster fine-grained memory entries into high-level abstract units. Memory-R1 ~\citep{yan2025memory} introduces an auxiliary agent equipped with a dedicated ``memory mcanager'' tool to handle memory updates. Despite these advances, current solutions remain constrained: many are still driven by manually engineered rules or are optimized for narrow, task-specific learning objectives, making them difficult to generalize to open-ended settings. 

\subsubsection{Future Perspective}

To support truly automated memory management, a promising direction is to \textbf{integrate memory construction, evolution, and retrieval directly into the agent's decision loop via explicit tool calls}, making the agent itself reason about memory operations instead of depending on external modules or hand-crafted workflows. Compared with existing designs that separate an agent's internal reasoning process from its memory management actions, an LLM agent can know precisely what memory actions it performs (e.g., add/update/delete/retrieval) in this tool-based strategy, leading to more coherent, transparent, and contextually grounded memory behavior.

Another key frontier lies in developing \textbf{self-optimizing memory structures} adopting hierarchical and adaptive architectures inspired by cognitive systems. First, hierarchical memory structure has been shown to improve the efficiency and performance ~\citep{kang2025memoryosaiagent}. Beyond hierarchy, self-evolving memory systems that dynamically link, index, and reconstruct memory entries enable the memory storage itself to self-organize over time, supporting richer reasoning and reducing dependence on hand-designed rules. Ultimately, such adaptive, self-organizing memory architectures pave the way toward agents capable of maintaining robust, scalable, and truly autonomous memory management. 

\subsection{Reinforcement Learning Meets Agent Memory}

\begin{figure}[!ht]
    \centering
    \includegraphics[width=\linewidth]{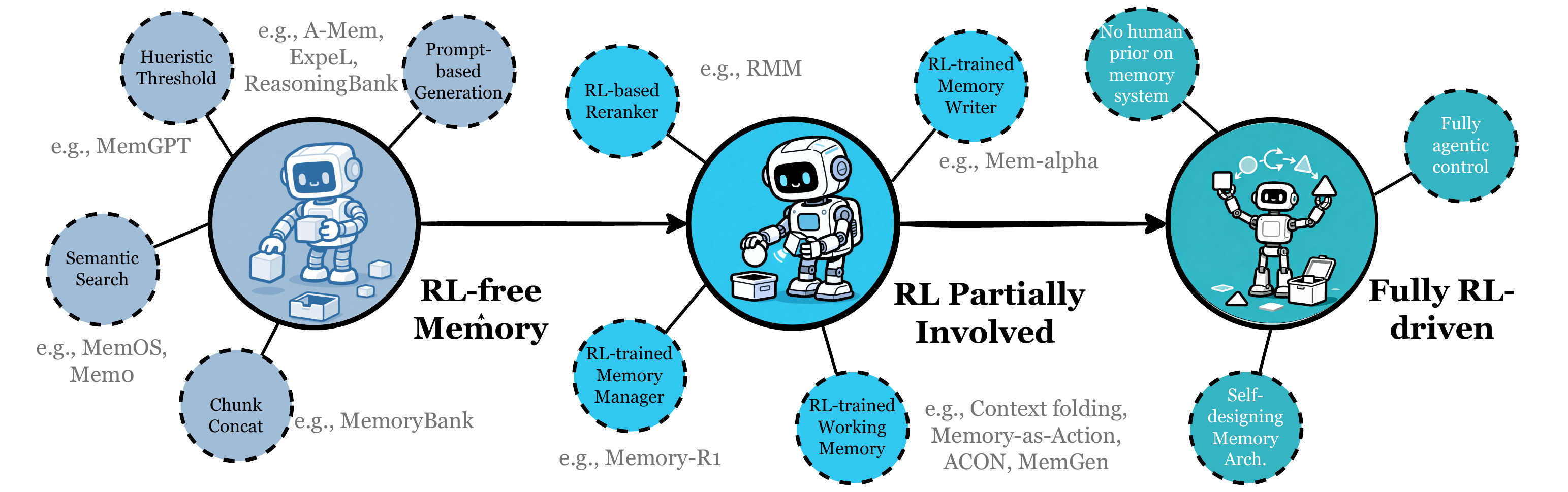}
    \caption{
    The evolution of RL-enabled agent memory systems.
A conceptual progression from \textbf{RL-free} memory systems based on heuristic or prompt-driven pipelines, to \textbf{partially RL-involved} designs where reinforcement learning governs selected memory operations, and finally to fully \textbf{RL-driven} memory systems in which memory architectures and control policies are learned end-to-end. This evolution reflects a broader paradigm shift from \textit{manually} engineered memory pipelines toward \textit{model-native}, \textit{self-optimizing} memory management in LLM-based agents.
    }
    \label{fig:sec6-rl-memory}
\end{figure}

\subsubsection{Look-Back: RL is Internalizing Memory Management Abilities for Agents.} 
Reinforcement learning is rapidly reshaping the development paradigm of modern LLM-based agents. Across a wide spectrum of agentic capabilities, including planning, reasoning, tool use, as well as across diverse task domains such as mathematical reasoning, deep research, and software engineering, RL has begun to play a central role in driving agent performance~\citep{zhang2025landscapeagenticreinforcementlearning,zhang2025surveyreinforcementlearninglarge,feng2024naturallanguagereinforcementlearning}. Memory, as one of the foundational components of agentic capability, follows a similar trend from pipeline-based to model-native paradigm~\citep{sang2025pipelinessurveyparadigmshift}. The agent memory research community is collectively transitioning from \textbf{early heuristic and manually engineered designs} to approaches in which \textbf{RL increasingly governs key decisions}. Looking ahead, it is reasonable to expect that \textbf{fully RL-based memory systems} may eventually become the dominant direction. Before discussing this trajectory in detail, we briefly outline the first stage of development.
This transition, in which memory management is progressively internalized and optimized through reinforcement learning, is schematically illustrated in \autoref{fig:sec6-rl-memory}.

\paragraph{RL-free Memory Systems} 
A substantial portion of the agent memory literature surveyed earlier can be categorized as RL-free memory systems. These approaches typically rely on heuristic or manually specified mechanisms, such as fixed thresholding rules inspired by curves of forgetting, rigid semantic search pipelines found in frameworks such as MemOS~\citep{li2025memosoperatingmemoryaugmentedgeneration}, Mem0~\citep{Chhikara2025mem0}, and MemoBase~\citep{githubGitHubMemodbiomemobase}, or simple concatenation-based strategies for storing memory chunks. In some systems, an LLM participates in memory management in a way that appears \textit{agentic}, yet the underlying behavior is entirely prompt-driven. The LLM is asked to generate memory entries but has not received any dedicated training for effective memory control, as seen in systems such as Dynamic Cheatsheet~\citep{suzgun2025dynamiccheatsheettesttimelearning}, ExpeL~\citep{DBLP:conf/aaai/Zhao0XLLH24}, EvolveR~\citep{wu2025evolverselfevolvingllmagents}, and G-Memory~\citep{Zhang2025GMemory}. This class of methods has dominated early work in the field and is likely to remain influential for some time due to its simplicity and practical accessibility.

\paragraph{RL-assisted Memory Systems}  
As the field progressed, many works began to incorporate RL-based methods into selected components of the memory pipeline. An early attempt in this direction is RMM~\citep{DBLP:conf/acl/rmm2025}, which employed a lightweight policy gradient learner to rank memory chunks after an initial retrieval stage based on BM25 or other semantic similarity metrics. Later systems explored substantially more ambitious designs. For example, Mem-$\alpha$~\citep{DBLP:journals/corr/abs-2509-25911} delegates the entire process of memory construction to an agent trained with RL, and Memory-R1~\citep{yan2025memory} employs a similar philosophy. A rapidly expanding line of research investigates how an agent can autonomously fold, compress, and manage context in ultra-long multi-turn tasks. This setting corresponds to the management of working memory~\citep{kangACONOptimizingContext2025,ye2025agentfold}. Many of the leading systems in this area are trained with RL, including but not limited to Context Folding~\citep{DBLP:journals/corr/abs-2510-11967}, Memory-as-Action~\citep{DBLP:journals/corr/abs-2510-12635}, MemSearcher~\citep{yuanMemSearcherTrainingLLMs2025}, and IterResearch~\citep{chen2025iterresearchrethinkinglonghorizonagents}. These RL-assisted approaches have already demonstrated strong capabilities and point toward the increasing role of RL in future memory system design.

\subsubsection{Future Perspective}

Looking forward, we anticipate that \textbf{fully RL-driven memory systems will constitute the next major stage} in the evolution of agent memory. We highlight two properties that such systems should ideally embody.

\begin{itemize}
\item First, \textit{memory architectures managed by agents should minimize reliance on human-engineered priors}. Many existing frameworks inherit design patterns inspired by human cognition, such as cortical or hippocampal analogies~\citep{gutiérrez2025hipporagneurobiologicallyinspiredlongterm}, or predefined hierarchical taxonomies that partition memory into episodic, semantic, and core categories~\citep{wang2025mirixmultiagentmemoryllmbased}. Although these abstractions have been useful for grounding early work, they may not represent the most effective or natural structures for artificial agents operating in complex environments. A fully RL-driven setting offers the possibility for agents to invent novel and potentially more suitable memory organizations that emerge directly from optimization dynamics rather than human intuition. In this view, the agent is encouraged to design new memory formats, storage schemas, or update rules through RL incentives, enabling memory architectures that are adaptive and creative rather than handcrafted. 

\item Second, \textit{future memory systems should provide agents with complete control over all stages of memory management}. Current RL-assisted approaches typically intervene in only a subset of the memory lifecycle. For instance, Mem-$\alpha$~ automates certain aspects of memory writing yet still relies on manually defined retrieval pipelines, whereas systems such as MemSearcher~\citep{yuanMemSearcherTrainingLLMs2025} focus primarily on short-term working memory without addressing long-term consolidation or evolution. A fully agentic memory system would require the agent to autonomously handle multi-granular memory formation, memory evolution, and memory retrieval in an integrated manner. Achieving this level of control will almost certainly require end-to-end RL training, since heuristic or prompt-based methods are insufficient for coordinating the complex interactions among these components across long-time horizons.

\end{itemize}

Together, these two directions suggest a future in which memory is not merely an auxiliary mechanism bolted onto an LLM agent, but rather a fully learnable and self-organizing subsystem that coevolves with the agent through RL. Such systems hold the potential to enable genuinely continual learning and long-term competence in artificial agents.

\subsection{Multimodal Memory}

\subsubsection{Look-Back}

As research on text-based memory becomes increasingly mature and extensively explored, and as multimodal large language models and unified models that jointly support multimodal understanding and generation continue to advance, attention has naturally expanded toward \textbf{multimodal memory}. This shift reflects a broader recognition that many real-world agentic settings are inherently multimodal, and that memory systems limited to text alone are insufficient to support long-horizon reasoning and interaction in complex environments.

Existing efforts on multimodal memory can be broadly grouped into two complementary directions. The first focuses on enabling \textbf{multimodal agents} to store, retrieve, and utilize memories derived from diverse sensory inputs~\citep{long2025seeing,zuo2025videolucydeepmemorybacktracking}. This direction is a natural extension of agent memory, since agents operating in realistic environments inevitably encounter heterogeneous data sources, including images, audio, video, and other non-textual signals~\citep{xie2024largemultimodalagentssurvey}. The degree of progress in multimodal memory closely follows the maturity of corresponding modalities. Visual modalities such as images and videos have received the most attention, leading to a growing body of work on visual and video memory mechanisms that support tasks such as visual grounding, temporal tracking, and long-term scene consistency~\citep{long2025seeing,wangVideoAgentLongFormVideo2024,gurukar2025longvmnetacceleratinglongformvideo,yu2025vismemlatentvisionmemory,bo2025agenticlearnergrowandrefinemultimodal,DBLP:journals/pami/WangCLJHZLHZYML25,li2024optimus1hybridmultimodalmemory}. In contrast, memory systems for audio and other modalities remain relatively underexplored~\citep{li2025omnivideobenchaudiovisualunderstandingevaluation}.

The second direction treats memory as an enabling component for \textbf{unified models}. In this setting, memory is leveraged not primarily to support agent decision making, but to enhance multimodal generation and consistency. For example, in image and video generation systems, memory mechanisms are often used to preserve entity consistency, maintain world state across frames, or ensure coherence across long generation horizons~\citep{DBLP:journals/corr/abs-2506-03141}. Here, memory serves as a stabilizing structure that anchors generation to previously produced content, rather than as a record of agent experience per se.

\subsubsection{Future Perspective}

Looking forward, multimodal memory is likely to become an indispensable component of agentic systems. As agents increasingly move toward embodied and interactive settings, their information sources will be inherently multimodal, spanning perception, action, and environmental feedback. Effective memory systems must therefore support the storage, integration, and retrieval of heterogeneous signals in a unified manner.

Despite recent progress, there is currently no memory system that provides truly \textbf{omnimodal support}. Most existing approaches remain specialized to individual modalities or loosely coupled across modalities. A key future challenge lies in designing memory representations and operations that can flexibly accommodate diverse modalities while preserving semantic alignment and temporal coherence. Moreover, multimodal memory must evolve beyond passive storage to support abstraction, cross-modal reasoning, and long-term adaptation. Addressing these challenges will be essential for enabling agents that can operate robustly and coherently in rich, multimodal environments.

\subsection{Shared Memory in Multi-Agent Systems}

\subsubsection{Look-Back: From Isolated Memories to Shared Cognitive Substrates}

As LLM-based multi-agent systems (MAS) have gained prominence, \textbf{shared memory} has emerged as a key mechanism for enabling coordination, consistency, and collective intelligence. Early multi-agent frameworks primarily relied on \textbf{isolated local memories} coupled with explicit message passing, where agents exchanged information through dialogue histories or task-specific communication protocols~\citep{qian2024chatdevcommunicativeagentssoftware,wu2024autogen,hu2025automated,zhang2025aflow}. While this design avoided direct interference between agents, it often suffered from redundancy, fragmented context, and high communication overhead, especially as team size and task horizon increased.

Subsequent work introduced \textbf{centralized shared memory structures}, such as global vector stores, blackboard systems, or shared documents~\citep{hong2023metagpt}, accessible to all agents. These designs enabled a form of team-level memory that supported joint attention, reduced duplication, and facilitated long-horizon coordination. Representative systems demonstrated that shared memory could serve as a persistent common ground for planning, role handoff, and consensus building~\citep{rezazadeh2025collaborativememorymultiusermemory,xu2025sedmscalableselfevolvingdistributed}. However, naive global sharing also exposed new challenges, including memory clutter, write contention, and the lack of role- or permission-aware access control.

\subsubsection{Future Perspective}

Looking forward, shared memory is likely to evolve from a passive repository into an \textbf{actively managed and adaptive collective representation}. One important direction is the development of \textbf{agent-aware shared memory}, where read and write behaviors are conditioned on agent roles, expertise, and trust, enabling more structured and reliable knowledge aggregation. 

Another promising avenue lies in \textbf{learning-driven shared memory management}. Rather than relying on hand-designed policies for synchronization, summarization, or conflict resolution, future systems may train agents to decide when, what, and how to contribute to shared memory based on long-horizon team performance. Finally, as MAS increasingly operate in open-ended and multimodal environments, shared memory must support abstraction across heterogeneous signals while maintaining temporal and semantic coherence, for which we believe latent memory exhibits a promising path~\citep{Wu2025CoMEM}. Advancing in these directions will be critical for scaling shared memory from a coordination aid into a foundation for robust collective intelligence.

\subsection{Memory for World Model}
\subsubsection{Look-Back}
The core objective of a World Model is to construct an internal environment capable of high-fidelity simulation of the physical world. These systems serve as the critical infrastructure for next-generation artificial intelligence. The core attribute of world model is to generate content that is both infinitely extensible and interactive in real time. Unlike traditional video generation that creates fixed-length clips, world models operate in an iterative manner by receiving actions at each step and predicting the next state to provide continuous feedback.
In this iterative framework, the memory mechanism becomes the cornerstone of the system. Memory stores and maintains the spatial and semantic information or hidden states from the previous time step. It ensures that the generation of the next chunk maintains long-term consistency with the preceding context regarding scene layout, object attributes, and motion logic. Essentially, the memory mechanism enables world models to handle long-term temporal dependencies and realize trustworthy simulation interactions.

Previously, memory modeling relied on simplistic buffering approaches. Frame Sampling conditioned generation on a few historical frames~\citep{bruce2024genie}. While intuitive, this led to context fragmentation and perceptual drift as early details were lost. Sliding Window methods adapted LLM techniques like attention sinks and local KV caches~\citep{liu2025rolling}. Although this resolved computational bottlenecks, it restricted memory to a fixed window. Once an object left this view, the model effectively forgot it, preventing complex tasks like loop closure.
By late 2025, the field shifted from finite context windows to structured state representations. Current architectures follow three main paths:

\begin{itemize}
    \item State-Space Models (SSMs) architectures like Long-Context SSMs utilize Mamba-style backbones~\citep{po2025long,yu2025videossm}. These compress infinite history into a fixed-size recursive state, enabling theoretically infinite memory capacity with constant inference costs.  
\item Explicit Memory Banks. Unlike compressed states, these systems maintain an external storage of historical representations to support precise recall. Approaches differ in their structuring logic: UniWM employs a \textit{hierarchical design}, separating short-term perception from long-term history via feature-based similarity gating~\citep{dong2025unified}. Conversely, \textbf{retrieval-based approaches} like WorldMem and Context-as-Memory (CaM) maintain a flat bank of past contexts, utilizing \textit{geometric retrieval} (e.g., FOV overlap) to dynamically select relevant frames for maintaining 3D scene consistency~\citep{xiao2025worldmem, yu2025context}.

\item Sparse Memory and Retrieval To balance long-term adherence with efficiency, Genie Envisioner and Ctrl-World utilize sparse memory mechanisms~\citep{liao2025genie, guo2025ctrl}. These models augment current observations by injecting sparsely sampled historical frames or retrieving pose-conditioned context to anchor predictions and prevent drift during manipulation tasks.  

\end{itemize}

\subsubsection{Future Perspective}

From an architectural perspective, the field is undergoing a fundamental transition from Data Caching which focuses on passive retention to State Simulation which focuses on active maintenance. This evolution is currently crystallizing into two distinct paradigms that aim to solve the conflict between real-time responsiveness and long-term logical consistency.
\begin{itemize}
    \item The Dual-System Architecture. Inspired by cognitive science, world models could be bifurcated into fast and slow components. System 1 represents the fast and instinctive layer that handles immediate physics and fluid interaction using efficient backbones like SSMs. System 2 represents the slow and deliberative layer that handles complex reasoning, planning, and world consistency using large-scale VLMs or explicit memory databases. 
    \item Active Memory Management. Passive mechanisms are being superseded by Active Memory Policies. Instead of treating memory as a fixed buffer that blindly stores recent history, new models are designed as Cognitive Workspaces that actively curate, summarize, and discard information based on task relevance. Recent empirical studies demonstrate that such active memory management significantly outperforms static retrieval methods in handling functional infinite context. This shift marks the move from simply remembering the last N tokens to maintaining a coherent and queryable world state.
\end{itemize}

\subsection{Trustworthy Memory}

\subsubsection{Look-Back: From Trustworthy RAG to Trustworthy Memory} 
As shown throughout this survey, memory plays a foundational role in enabling agentic behavior, which supports persistence, personalization, and continual learning. However, as memory systems become more deeply embedded into LLM-based agents, the question of \textit{trustworthiness} has become paramount. 

Earlier concerns around hallucination and factuality in retrieval-augmented generation (RAG) systems~\citep{niu2024ragtruth,sunredeep,lu2025spad} have now evolved into a broader trust discourse for memory-augmented agents. Similar to RAG, one major motivation for using external or long-term memory is to reduce hallucinations by grounding model outputs in retrievable, factual content~\citep{ru2024ragchecker,wang2025astute}. However, unlike RAG, agent memory often stores user-specific, persistent, and potentially sensitive content, ranging from factual knowledge to past interactions, preferences, or behavioral traces. This introduces additional challenges in privacy, interpretability, and safety.

Recent work by~\citet{wang2025unveilingprivacyrisksllm} demonstrates that memory modules can leak private data through indirect prompt-based attacks, highlighting the risk of memorization and over-retention. Concurrently,~\citet{wu2025humanmemoryaimemory}  argues that agent memory systems must support explicit mechanisms for \textit{access control}, \textit{verifiable forgetting}, and \textit{auditable updates} to remain trustworthy. Notably, such threats are magnified in agent scenarios where memory persists across long time horizons.

Explainability also remains a critical bottleneck. While explicit memory, such as text logs or key-value stores, offers some transparency, users and developers still lack tools to trace which memory items were retrieved, how they influenced generation, or whether they were misused. In this regard, diagnostic tools like RAGChecker~\citep{ru2024ragchecker} and conflict-resolution frameworks such as RAMDocs with MADAM-RAG~\citep{wang2025retrieval} provide inspiration for tracing memory usage and reasoning under uncertainty. 

Moreover, beyond individual memory,~\citet{shi2025privacy} and~\citet{rezazadeh2025collaborative} highlight the emerging importance of \textit{collective privacy} in shared or federated memory systems, which may operate across multi-agent deployments or organizations. All these developments collectively signal a need to elevate trust as a first-class principle in memory design.

\subsubsection{Future Perspective}
Looking ahead, we argue that \textbf{trustworthy memory} must be built around three interlinked pillars: \textit{privacy preservation}, \textit{explainability}, and \textit{hallucination robustness}—each demanding architectural and algorithmic innovations.

For privacy, future systems should support granular permissioned memory, user-governed retention policies, encrypted or on-device storage, and federated access where needed~\citep{wu2025humanmemoryaimemory,shi2025privacy,rezazadeh2025collaborative}. Techniques like differential privacy, memory redaction, and adaptive forgetting (e.g., decay-based models or user-erasure interfaces) can serve as safeguards against both memorization and leakage~\citep{Chhikara2025mem0}.

Explainability requires moving beyond visible content to include \textit{traceable access paths}, \textit{self-rationalizing retrievals}, and possibly counterfactual reasoning (e.g., what would have changed without this memory?)~\citep{openai2024memorycontrol,zhang2025explicit}. Visualizations of memory attention, causal graphs of memory influence, and user-facing debugging tools may become standard components.

Hallucination mitigation will benefit from continued advances in conflict detection, multi-document reasoning, and uncertainty-aware generation. Strategies such as abstention under low-confidence retrieval, fallback to model priors~\citep{wang2025astute}, or multi-agent cross-checking~\citep{hu2024refchecker} are promising. Beyond behavioral safeguards, emerging \textit{mechanistic interpretability} techniques offer a complementary direction by analyzing how internal representations and reasoning circuits contribute to hallucinated outputs. 
Methods such as representation-level probing and reasoning-path decomposition enable finer-grained diagnosis of where hallucinations originate, and provide principled tools for intervention and control~\citep{sunredeep,sunlargepig}.

In the long term, we envision memory systems governed by OS-like abstractions: segmented~\citep{lscs}, version-controlled, auditable, and jointly managed by agent and user~\citep{packer2023memgpt}. Building such systems will require coordinated efforts across representation learning, system design, and policy control. As LLM agents begin to operate in persistent, open-ended environments, trustworthy memory will not just be a desirable feature—but a foundational requirement for real-world deployment.

\subsection{Human-Cognitive Connections} \label{ssec:human_cognitive_connections}
\subsubsection{Look Back}
The architecture of contemporary agent memory systems has converged with foundational models of human cognition established over the last century. The prevailing design, which couples a capacity-limited context window with massive external vector databases, mirrors the Atkinson-Shiffrin multi-store model~\citep{atkinsonHumanMemoryProposed1968}, effectively instantiating an artificial counterpart to the distinction between working memory and long-term memory~\citep{baddeleyWorkingMemoryTheories2012}. 
Furthermore, the partitioning of agent memory into interaction logs, world knowledge, and code-based skills exhibits a striking structural alignment with \textit{Tulving's} classification of \textit{episodic}, \textit{semantic}, and \textit{procedural} memory~\citep{tulvingEpisodicSemanticMemory1972,squireMemorySystemsBrain2004}.
Current frameworks~\citep{zhong2023memorybankenhancinglargelanguage,park2023generativeagentsinteractivesimulacra,gutiérrez2025hipporagneurobiologicallyinspiredlongterm,li2025memosoperatingmemoryaugmentedgeneration} operationalize these biological categories into engineering artifacts, where episodic memory provides autobiographical continuity and semantic memory offers generalized world knowledge.

Despite these structural parallels, a fundamental divergence remains in the \textit{dynamics} of retrieval and maintenance. Human memory operates as a \textit{constructive process}, where the brain actively reconstructs past events based on current cognitive states rather than replaying exact recordings~\citep{schacterConstructiveMemoryGhosts2007}. In contrast, the majority of existing agent memory systems rely on verbatim retrieval mechanisms like RAG, treating memory as a repository of \textit{immutable} tokens to be queried via semantic similarity~\citep{packer2023memgpt,Chhikara2025mem0}. Consequently, while agents possess a veridical record of the past, they lack the biological capacity for memory distortion, abstraction, and the dynamic remodeling of history that characterizes human intelligence.

\subsubsection{Future Perspective}
To bridge the gap between static storage and dynamic cognition, the next generation of agents must evolve beyond exclusive \textbf{online updating} by incorporating \textbf{offline consolidation} mechanisms analogous to biological sleep~\citep{linSleeptimeComputeInference2025}. Drawing from the Complementary Learning Systems (CLS) theory~\citep{kumaranWhatLearningSystems2016,mcclellandWhyThereAre1995}, future architectures will likely introduce dedicated consolidation intervals where agents decouple from environmental interaction to engage in memory reorganization and generative replay~\citep{mattarPrioritizedMemoryAccess2018}. During these offline periods, agents can autonomously distill generalizable schemas from raw episodic traces, perform \textbf{active forgetting} to prune redundant noise~\citep{andersonActiveForgettingAdaptation2021}, and optimize their internal indices without the latency constraints of real-time processing.

Ultimately, this evolution suggests a paradigm shift in memory forms and functions: moving from explicit text retrieval to \textbf{generative reconstruction}. Future systems may utilize generative memory~\citep{Zhang2025MemGen} where the agent synthesizes latent memory tokens on demand, mirroring the brain's reconstructive nature. By integrating sleep-like consolidation cycles, agents will evolve from entities that merely archive data to those that internalize experience, resolving the stability-plasticity dilemma by periodically compacting vast episodic streams into efficient, parametric intuition.



\section{Conclusion}\label{sec:conclusion}

This survey has examined agent memory as a foundational component of modern LLM-based agentic systems. By framing existing research through the unified lenses of \textit{forms, functions, and dynamics}, we have clarified the conceptual landscape of agent memory and situated it within the broader evolution of agentic intelligence. On the level of \textbf{forms}, we identify three principal realizations: token-level, parametric, and latent memory, each of which has undergone distinct and rapid advances in recent years, reflecting fundamentally different trade-offs in representation, adaptability, and integration with agent policies. On the level of \textbf{functions}, we move beyond the coarse long-term versus short-term dichotomy prevalent in prior surveys, and instead propose a more fine-grained and encompassing taxonomy that distinguishes \textit{factual, experiential, and working memory} according to their roles in knowledge retention, capability accumulation, and task-level reasoning. Together, these perspectives reveal that memory is not merely an auxiliary storage mechanism, but an essential substrate through which agents achieve temporal coherence, continual adaptation, and long-horizon competence.

Beyond organizing prior work, we have identified key challenges and emerging directions that point toward the next stage of agent memory research. In particular, the increasing integration of reinforcement learning, the rise of multimodal and multi-agent settings, and the shift from retrieval-centric to generative memory paradigms suggest a future in which memory systems become fully learnable, adaptive, and self-organizing. Such systems hold the potential to transform large language models from powerful but static generators into agents capable of sustained interaction, self-improvement, and principled reasoning over time.

We hope this survey provides a coherent foundation for future research and serves as a reference for both researchers and practitioners. As agentic systems continue to mature, the design of memory will remain a central and open problem, one that is likely to play a decisive role in the development of robust, general, and enduring artificial intelligence.

\bibliographystyle{assets/plainnat}
\bibliography{citation}
\clearpage
\end{document}